\documentclass[accepted]{uai2022} 

\usepackage[dvipsnames]{xcolor, colortbl}

\usepackage[american]{babel}

\usepackage{natbib} 
\usepackage{mathtools} 
\usepackage{booktabs} 
\usepackage{tikz} 

\usepackage{graphicx}
\usepackage{subfig}

\usepackage{amsmath,amssymb,amsfonts}
\usepackage{bbm}
\usepackage{hhline}
\usepackage[normalem]{ulem}
\usepackage{xr}
\makeatletter
\newcommand*{\addFileDependency}[1]{
  \typeout{(#1)}
  \@addtofilelist{#1}
  \IfFileExists{#1}{}{\typeout{No file #1.}}
}
\makeatother

\newcommand*{\myexternaldocument}[1]{%
    \externaldocument{#1}%
    \addFileDependency{#1.tex}%
    \addFileDependency{#1.aux}%
}
\myexternaldocument{uhlemeyer_491-supp}

\makeatletter
\DeclareMathOperator*{\argmax}{arg\,max}

\definecolor{Gray}{gray}{0.9}
\definecolor{DarkGray}{gray}{0.8}
\definecolor{maroon}{cmyk}{0,0.87,0.68,0.32}
\newcolumntype{a}{>{\columncolor{DarkGray}}c}
\newcolumntype{g}{>{\columncolor{Gray}}c}

\usepackage[capitalize]{cleveref}
\crefname{section}{Sec.}{Secs.}
\Crefname{section}{Section}{Sections}
\Crefname{table}{Table}{Tables}
\crefname{table}{Tab.}{Tabs.}

\title{Towards Unsupervised Open World Semantic Segmentation}

\author[1]{\href{mailto:<uhlemeyer@math.uni-wuppertal.de>?Subject=Your UAI 2022 paper}{Svenja Uhlemeyer}{}}
\author[1]{\href{mailto:<rottmann@uni-wuppertal.de>?Subject=Your UAI 2022 paper}{Matthias Rottmann}{}}
\author[1]{\href{mailto:<hgottsch@uni-wuppertal.de>?Subject=Your UAI 2022 paper}{Hanno Gottschalk}{}}
\affil[1]{%
    IZMD and Faculty of Mathematics and Natural Sciences\\
    University of Wuppertal, Germany\\
}

\begin{document}
  
\newcommand{\MR}[1]{\textcolor{green!50!blue}{#1}}
\newcommand{\HG}[1]{\textcolor{orange}{#1}}
\newcommand{\SU}[1]{\textcolor{magenta}{#1}}
\newcommand{\try}[1]{\textcolor{violet}{#1}}
\newcommand{\outMR}[1]{\textcolor{green!50!blue}{\sout{#1}}}
\newcommand{\outHG}[1]{\textcolor{orange}{\sout{#1}}}
\newcommand{\outcom}[1]{\textcolor{violet}{\sout{#1}}}  
\newcommand{\ia}{\textit{i.a., }} 
\newcommand{\ie}{\textit{i.e., }} 
\newcommand{\eg}{\textit{e.g., }} 

\maketitle

\begin{abstract}
   For the semantic segmentation of images, state-of-the-art deep neural networks (DNNs) achieve high segmentation accuracy if that task is restricted to a closed set of classes. However, as of now DNNs have limited ability to operate in an open world, where they are tasked to identify pixels belonging to unknown objects and eventually to learn novel classes, incrementally. Humans have the capability to say: ``I don't know what that is, but I've already seen something like that''. Therefore, it is desirable to perform such an incremental learning task in an unsupervised fashion. We introduce a method where unknown objects are clustered based on visual similarity. Those clusters are utilized to define new classes and serve as training data for unsupervised incremental learning. More precisely, the connected components of a predicted semantic segmentation are assessed by a segmentation quality estimate. Connected components with a low estimated prediction quality are candidates for a subsequent clustering. Additionally, the component-wise quality assessment allows for obtaining predicted segmentation masks for the image regions potentially containing unknown objects. The respective pixels of such masks are pseudo-labeled and afterwards used for re-training the DNN, \ie without the use of ground truth generated by humans. In our experiments we demonstrate that, without access to ground truth and even with few data, a DNN's class space can be extended by a novel class, achieving considerable segmentation accuracy.
\end{abstract}

\section{Introduction}

\begin{figure}[t]
    \captionsetup[subfigure]{labelformat=empty, position=top}
    \centering
    \subfloat[image \& novelty annotation ]{\includegraphics[width=0.235\textwidth]{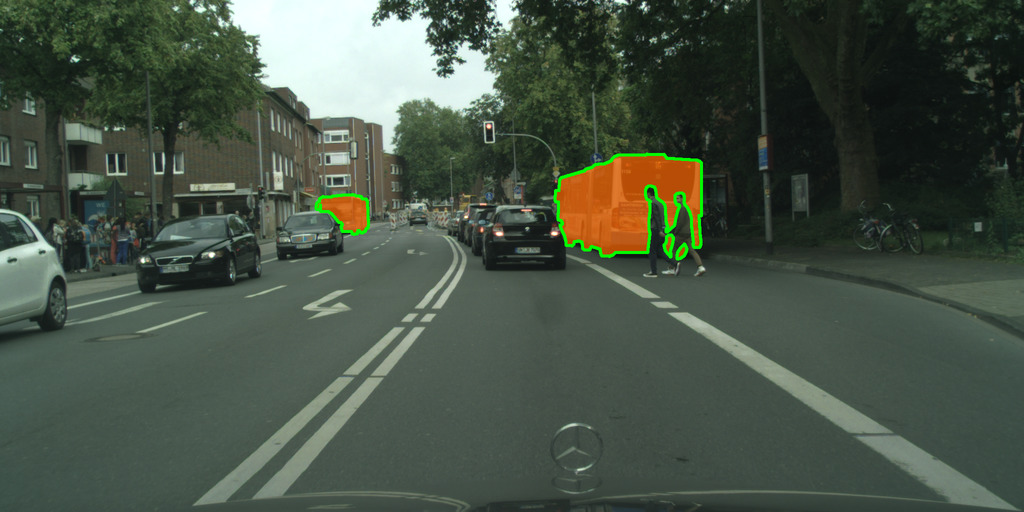}}~
    \subfloat[prediction quality estimation]{\includegraphics[width=0.235\textwidth]{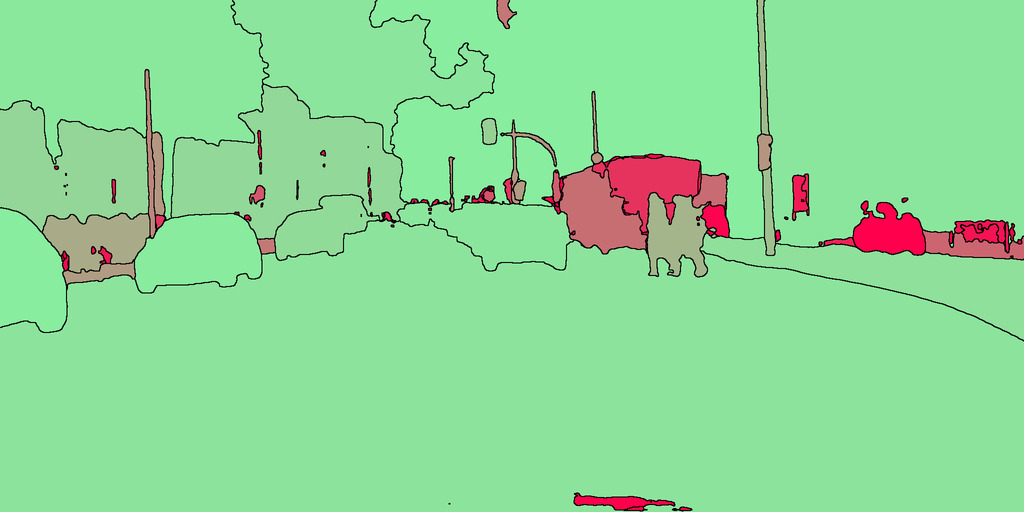}}\\
    \captionsetup[subfigure]{labelformat=empty, position=bottom}
    \subfloat[prediction of the initial DNN]{\includegraphics[width=0.235\textwidth]{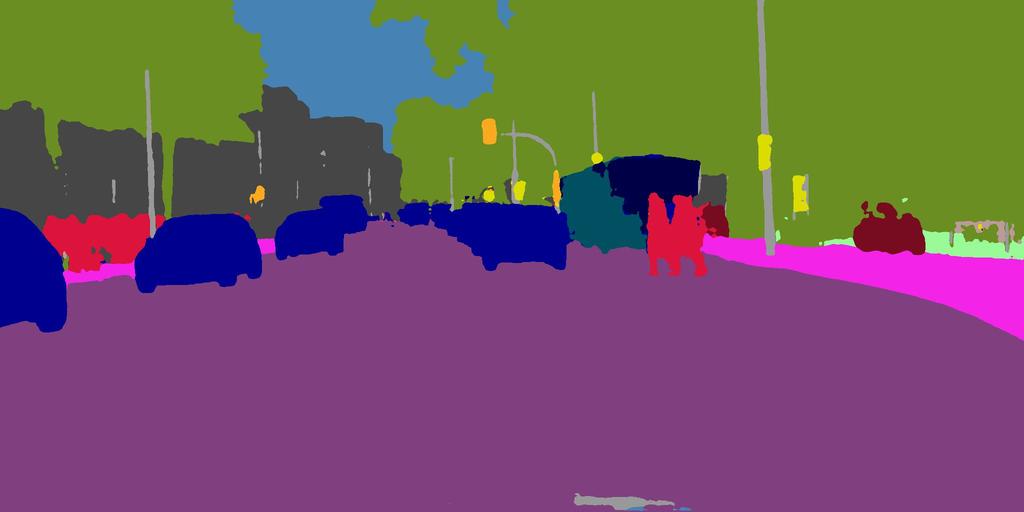}}~
    \subfloat[prediction of our extended DNN]{\includegraphics[width=0.235\textwidth]{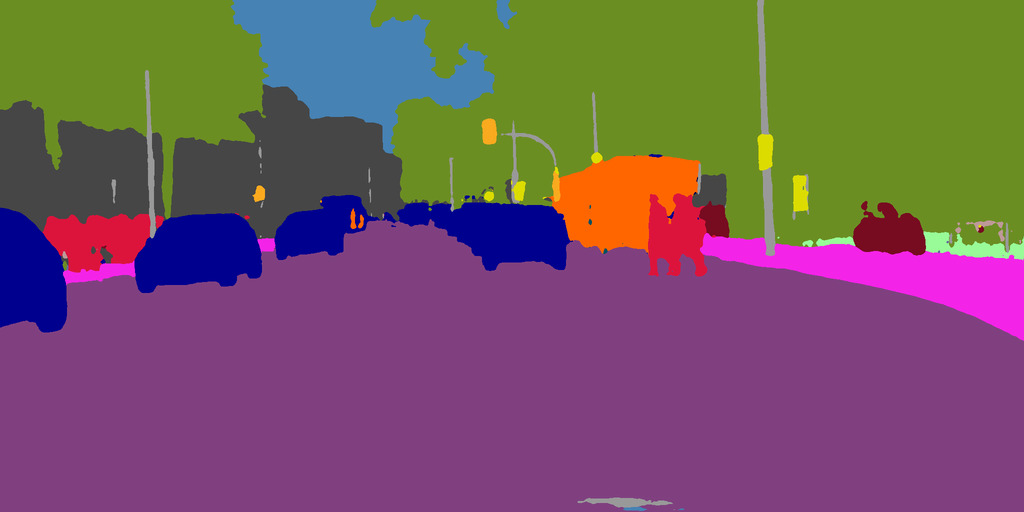}}
    \caption{Comparison of the semantic segmentation predictions of an initial DNN (bottom left) whose semantic space does not include the category \emph{bus} and a DNN which is incrementally extended by this novel class (bottom right, novel class in orange) for an image from the Cityscapes dataset. The novel class is highlighted in orange (top left). Further, the initial prediction exhibits a low prediction quality (top right) on pixels belonging to the novel objects, which is indicated by red color.
    }
    \label{fig:page1}
\end{figure}

Semantic segmentation is a computer vision task that terms the classification of image data on pixel level. State-of-the-art approaches are based on deep convolutional neural networks (DNNs) \citep{chen2018encoderdecoder,wang2020deep,Zhao2017PyramidSP}, benefiting from finely annotated datasets, \eg for automated driving \citep{Cordts2016TheCD,geyer2020a2d2,8237796,yu2020bdd100k}. However, DNNs for semantic segmentation are usually trained on a predefined, closed set of classes. This closed world setting assumes, that all classes present during testing were already included in the training set. In an open world setting, this assumption does not hold. In particular for safety-critical open-world applications like perception systems for automated driving, it is indispensable that neural networks recognize previously unseen objects instead of wrongly assigning them to \emph{one-of-the-known} classes. In addition, they must constantly adapt to evolving environments.

Some terms often used interchangeably for anomaly are \emph{outlier}, \emph{out-of-distribution} (OoD) object and \emph{novelty}. As there is no clear convention on how to distinguish these terms, we define them as subcategories of anomalies: outliers and OoD objects denote noise or samples drawn from another distribution than the model was trained on, respectively. In this work, we are seeking novelties, which we define as previously-unseen objects that constitute a new concept, \ie objects of the same category appear frequently. In automated driving, detecting and learning those novel classes becomes necessary, \eg due to new appearances like e-scooters or due to local specialities like boat trailers near the sea. The concept of detecting and learning novelties was first introduced in \citet{Bendale2015TowardsOW} as \emph{open world recognition}.
Open world recognition for different computer vision tasks is an emerging research area \citep{Bendale2015TowardsOW,Joseph_2021_CVPR,Cen_2021_ICCV,shu2018unseen}, still only little explored for unsupervised methods \citep{He2021UnsupervisedCL,Nakajima2019IncrementalCD}, yet.

We propose a new and modular procedure for learning new classes of novel objects without any handcrafted annotation:
\begin{enumerate}
    \setlength\itemsep{0mm} 
    \item Anomaly segmentation to detect suspicious objects,
    \item clustering of potentially novel objects,
    \item creation of so-called \emph{pseudo labels}, and
    \item incremental learning of novel classes.
\end{enumerate}
In the following, we will outline each of these four steps in more detail.

For the first step, we post-process the predictions of an underlying semantic segmentation DNN via a \emph{meta regressor}, that estimates the quality of the predicted segments, similar as proposed in \citet{rottmann2019uncertainty,rottmann2019prediction,Maag2020TimeDynamicEO}. In the following, the term \textbf{segment} will always refer to connected components of pixels in the semantic segmentation prediction.
The segment-wise quality score is obtained on the basis of aggregated dispersion measures and geometrical information, \ie without requiring ground truth. The output of the semantic segmentation DNN on anomalous objects is often split into several segments. To this end, we first aggregate neighboring segments, \ie segments that have at least one adjacent pixel each, with quality estimates below some threshold, into (potentially) anomalous objects, termed \textbf{suspicious objects}.

For the second step, we adapt the idea introduced in \citet{oberdiek2020detection} to gather segments with poor prediction quality and to cluster them into visually related neighborhoods.  
Therefore, all suspicious objects (of sufficient size) are cropped out in the RGB images and the resulting image patches are fed into a convolutional neural network (CNN), \eg for image classification. Whether an image patch is sufficiently large depends on the minimum input size required by this CNN.
To obtain comparable information about the suspicious objects, we then extract the features provided by the penultimate layer of the CNN, \ie right before the final classification layer. By reducing the dimensionality of these features up to two, we enable the use of low-dimensional, unsupervised clustering techniques, such as \citet{dbscan,kmeans}.

As third, we obtain pseudo labels for novel classes in an automated manner: each (large / dense enough) cluster constitutes a novel category, and each pixel belonging to a clustered object is assigned to the appropriate (not necessarily named) class. More precisely, the prediction of the segmentation model is updated at those pixel positions to the next ``free'' label ID.

Finally, the segmentation network is incrementally extended by these novel classes (see \cref{fig:page1} for an example). To this end, we apply established incremental learning methods \citep{hinton2015distilling,rehearsal}. However, these are mainly examined for supervised learning tasks, while we do not include any hand-labeled new data. This last two steps were never done in literature so far.

We perform five experiments, following a hierarchical structure of complexity. For the first three experiments, the initial segmentation network is trained on the Cityscapes dataset, but on different subsets of the available training classes. Here, we do not change the data itself, but the training IDs of the Cityscapes classes. For the other experiments, we start with an initial segmentation network that is trained on Cityscapes and test our method on the A2D2 dataset. For those, we have a mapping between the Cityscapes and the A2D2 classes. For most Cityscapes classes, there is a matching class in A2D2. In some cases, A2D2 has coarser classes, \eg we map the Cityscapes classes \emph{vegetation} and \emph{terrain} to the A2D2 class \emph{nature}.

To outline our contributions, we demonstrate in our experiments that our method is able to incrementally extend a neural network by novel classes without collecting or annotating novelties manually. To the best of our knowledge, we are the first to introduce an unsupervised approach for open world semantic segmentation with DNNs. Fine-tuning neural networks on automatically created pseudo-labels instead of human-made annotations is economically valuable. We observe in all experiments, that even a poor labeling quality is sufficient to learn novel classes, achieving IoU values around $40\%$. Further, the amount of new data was less mostly than 100 images, respectively. Unsupervised open world semantic segmentation therefore is a powerful tool for open world applications, that provides an enormous potential for future improvement.

\section{Related Work}

\begin{figure*}[t]
    \center
    \includegraphics[width = 0.95\textwidth]{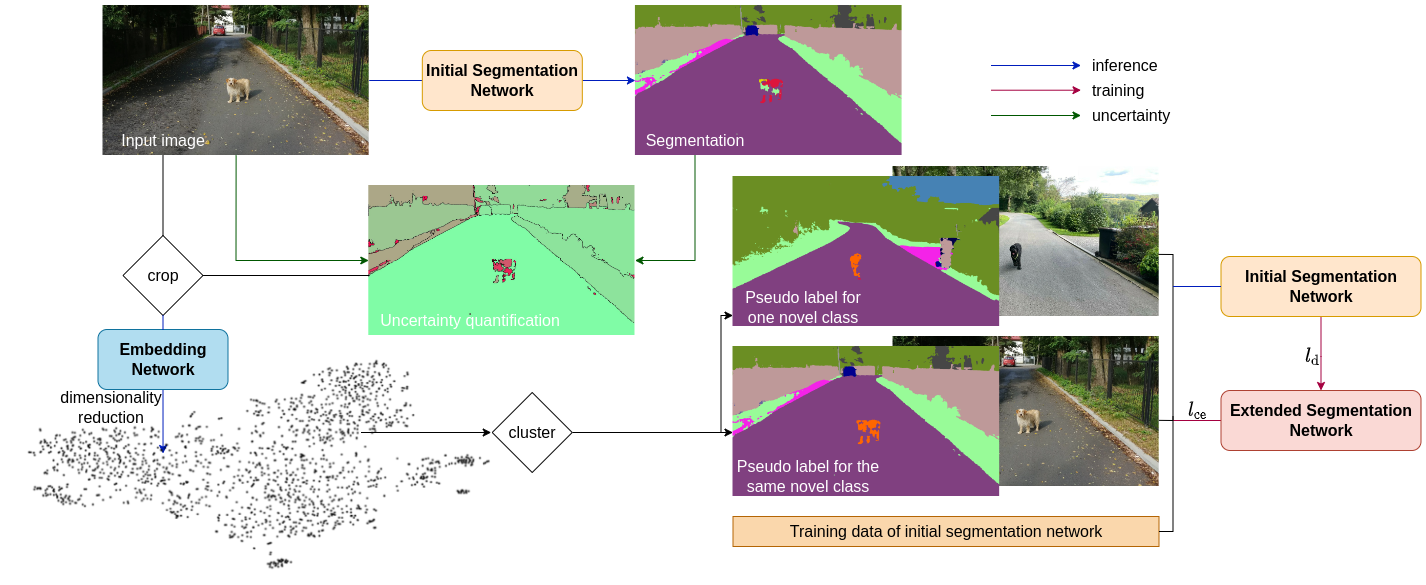}
    \caption{Illustration of the overall framework.}
    \label{fig:overview}
\end{figure*}

In this section, we first review anomaly detection methods and briefly go into class discovery approaches. Then we describe different strategies for class-incremental learning. Finally, we give an overview of existing work on open world computer vision tasks.

\paragraph{Novelty Detection.}
The detection of anomalous objects in general is a key task in many machine learning applications. Early works estimate the prediction uncertainty, \eg by uncertainty measures derived from the softmax probability \citep{Hendrycks2017msp,liang18odin}. Uncertainty-based approaches can be further improved by integrating anomalous data into the training procedure \citep{Devries2018LearningCF, Chan_2021_ICCV}. Another line of works employs generative models such as autoencoders (AEs) or generative adversarial models (GANs) to reconstruct or synthesise images and measure the reconstruction quality. Various of those novelty detection methods are described in \citet{Vasilev2018qSpaceND}, not only reconstruction-, but also density- or distance-based. A benchmark for anomaly segmentation, \ie anomaly detection methods for semantic segmentation, was recently published in \citet{chan2021segmentmeifyoucan}, providing a cleaner comparison of proposed methods. Given a set of anomalies, the prevailing approach for class discovery is to form clusters based on some similarity measure or intrinsic features with traditional clustering methods. A detailed survey of image clustering has been published in \citet{9517087}.

\paragraph{Class-Incremental Learning.} 
Class-incremental learning refers to the extension of a neural network's semantic space by further, previously unknown, classes. This extension is achieved by fine-tuning a model on additional, usually human-annotated data \citep{Jung2018LessforgetfulLF,Li2018LearningWF,Klingner2020ClassIncrementalLF,michieli2019incremental}, whereas in this work we only provide pseudo labels for these new images. The primary issue to tackle when re-training a neural network is to mitigate the performance loss on previously learned classes, commonly known as catastrophic forgetting \citep{McCloskey1989CatastrophicII}. To this end, we employ two different strategies: first, we penalize large variations of the softmax output (compared to the one of the original network) \citep{hinton2015distilling}, second we utilize a subset of the previously-seen training data \citep{rehearsal}. 

The first strategy belongs to the category of regularization based approaches, or more specifically to knowledge distillation methods. These were originally developed to distill knowledge from sophisticated into simpler models \citep{hinton2015distilling}, \ie for model compression. Thereupon, distillation methods have evolved for incremental learning in image classification \citep{Li2018LearningWF,Yao2019AdversarialFA,Kim2019IncrementalLW,Jung2018LessforgetfulLF,Lee2019OvercomingCF}, some of which were later adapted to semantic segmentation \citep{Klingner2020ClassIncrementalLF,michieli2019incremental,Tasar2019IncrementalLF}. 

The second approach belongs to so-called rehearsal methods \citep{rehearsal}, where old training data is included in the re-training process \citep{Rebuffi2017iCaRLIC,Castro2018EndtoEndIL}.

\paragraph{Open World.} 
The open world setting was first introduced in \citet{Bendale2015TowardsOW} for image classification. The authors formally define the solution of open world recognition problems as a tuple, consisting of a recognition function, a novelty detector, a labeling process and an incremental learning function. Ideally, these steps should be automated, however, most approaches presume a supervised setting, \ie they require ground truth for detected novelties. In summary, open world recognition covers the entire process from discovering up to learning novel classes.

A supervised solution for open world object detection is presented in \citet{Joseph_2021_CVPR}, based on contrastive clustering, an unknown-aware proposal network and energy based unknown identification. A similar approach was proposed in \citet{Cen_2021_ICCV} for open world semantic segmentation, where novel classes are learned via few-shot learning. In \citet{He2021UnsupervisedCL}, an unsupervised method to obtain pseudo labels for image classification based on cluster assignments is introduced. There exists also some prior work for unsupervised open world semantic segmentation \citep{Nakajima2019IncrementalCD}, however, the segmentation mask is obtained via agglomerative clustering of superpixels and there is no update of the neural network at all. While it is capable of creating ad hoc novel classes unsupervisedly on given images, it does not create a consistent semantic category over multiple images. 

Our work introduces an open world semantic segmentation framework, where a neural network is incrementally extended by novel classes. These classes are discovered \textbf{and} labeled without any human effort. Therefore, our work goes beyond all existing approaches in this research area.

\section{Discovery of Unknown Semantic Classes}\label{sec:discovery}

\begin{figure*}[t]
    \captionsetup[subfigure]{labelformat=empty}
    \centering
    \subfloat[image from A2D2]{\includegraphics[trim = {0 0 0 124px}, clip, width=0.23\textwidth]{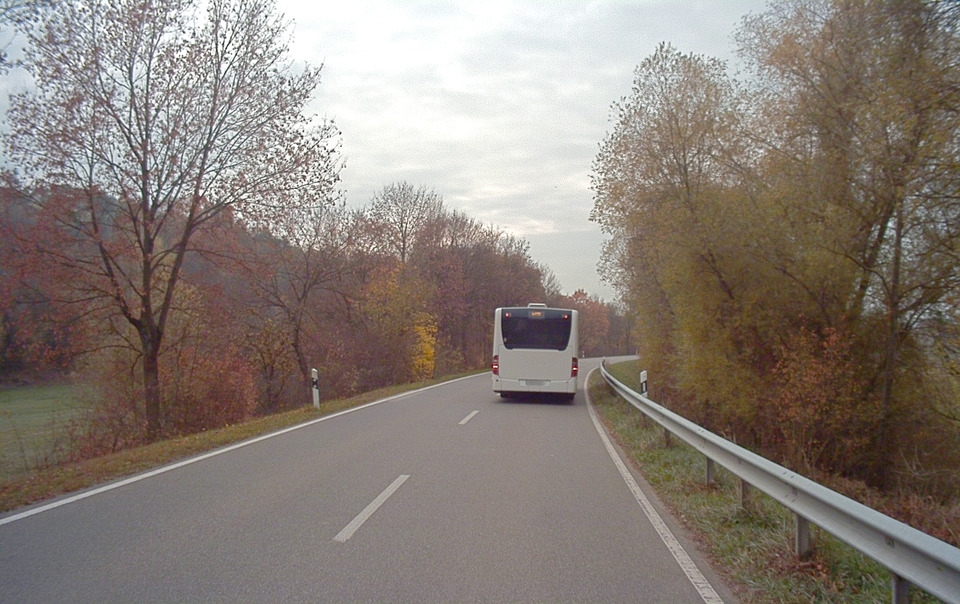}}~
    \subfloat[\centering{semantic segmentation prediction}]{\includegraphics[trim = {0 0 0 124px}, clip,width=0.23\textwidth]{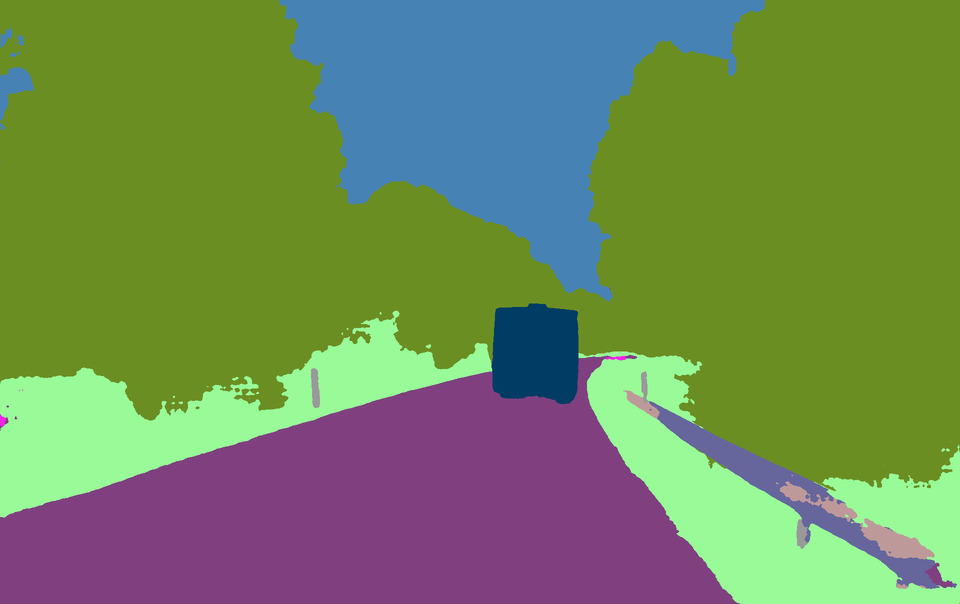}}~
    \subfloat[\centering{prediction quality estimation from 0 (red) to 1 (green)}]{\includegraphics[trim = {0 0 0 124px}, clip,width=0.23\textwidth]{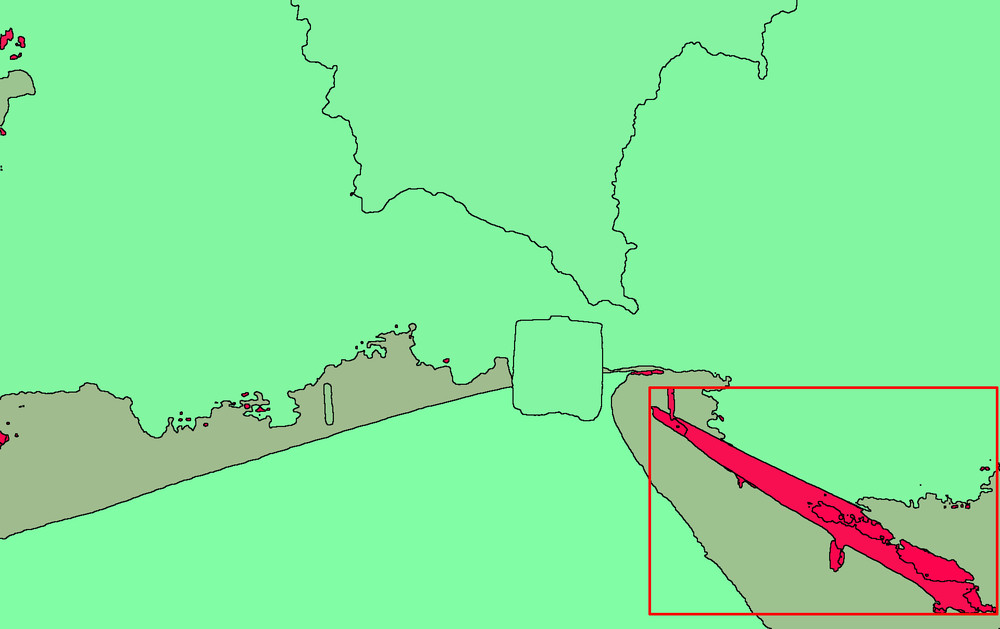}}~
    \subfloat[]{\includegraphics[width=0.0135\textwidth]{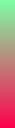}}~
    \subfloat[pseudo ground truth]{\includegraphics[trim = {0 0 0 124px}, clip,width=0.23\textwidth]{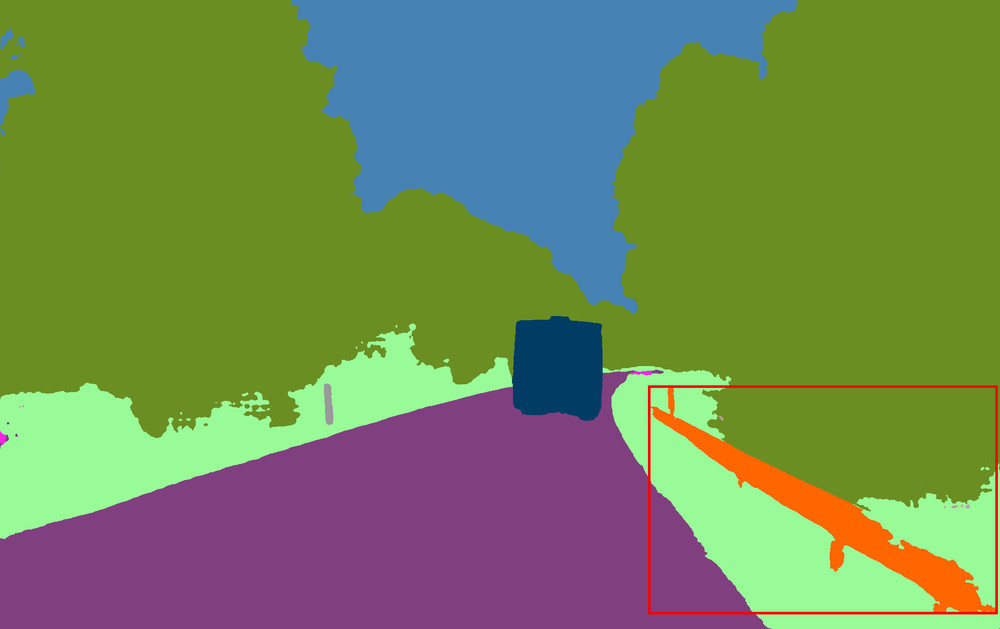}}
    \caption{Novelty segmentation: example for obtaining pseudo ground truth with regard to some image patch (outlined in red) of image $x$. If segments inside the red box exhibit quality estimates below some predefined threshold, they are ``re-labeled'' in the segmentation mask $m(x)$.
    }
    \label{fig:pseudolabel}
\end{figure*}

Whether a class is novel or not depends on the neural network's underlying set of known classes $\mathcal{C}=\{1,\ldots,C\}$. Let $f:\mathcal{X}\to(0,1)^{|\mathcal{H}|\times|\mathcal{W}|\times|\mathcal{C}|}$ be a semantic segmentation DNN which is trained on the classes in $\mathcal{C}$, mapping an image $x\in\mathcal{X}\subseteq [0,1]^{|\mathcal{H}|\times|\mathcal{W}|\times 3}$ onto its softmax probabilities for each pixel $z\in\mathcal{H}\times\mathcal{W}$. Then, $f_{z,c}(x) \in (0,1)$ denotes the probability with which the model $f$ assigns some pixel $z$ to a class $c\in\mathcal{C}$. As decision rule, we apply the $\argmax$ function, \ie we obtain the semantic segmentation mask $m(x)\in \mathcal{C}^{|\mathcal{H}|\times|\mathcal{W}|}$ with $m_z(x) = \argmax_{c\in\mathcal{C}} f_{z,c}(x)$. In the following, we will estimate the prediction quality on a segment-level instead of pixel-wise, employing a meta regression approach that was first introduced in \citet{rottmann2019prediction}. On that account, we denote a segment, \ie a connected component of pixels that share the same class in $m(x)$, as $k\in\mathcal{K}(x)$.

\paragraph{Meta Regressor.}
As model for the meta regressor we apply the gradient boosting from the \texttt{scikit-learn v.0.24.2} library using the standard settings. The training datasets contain from $67$ to $75$ uncertainty metrics depending on the number of classes. We train on $313,720$ to $946,318$ segments. Further details on the definition of the segment-wise metrics, the exact size of the training data and the tree models obtained are provided in the Appendix. For any predicted segment $k$, the gradient boosting regressor, via clipping, outputs a value between $0$ and $1$, where a value close to $0$ expresses low, a value close to $1$ high prediction quality.

The motivation to use a segment-wise meta regression framework is to identify segments with low predicted IoU as candidate segments that potentially stem from OoD objects.

\paragraph{Uncertainty Metrics and Prediction Quality Estimation.}
We consider novelties as \emph{none-of-the-known} objects, \ie they differ semantically from the model's training data. Assuming that the segmentation DNN produces unstable predictions on these unexplored entities, various measurable phenomena occur. For instance, the model exhibits a high prediction uncertainty. This is quantified by dispersion measures as the softmax entropy, probability margin or variation ratio, which we compute pixel-wise via
\begin{equation}
    E_z(f(x)) = - \frac{1}{\log(|\mathcal{C}|)} \sum\limits_{c\in\mathcal{C}} f_{z,c}(x)\log(f_{z,c}(x)) \; ,
\end{equation}
\begin{equation}
    D_z(f(x)) = 1 - \max_{c\in\mathcal{C}} f_{z,c}(x) + \max_{c\in\mathcal{C}\setminus\{m_z(x)\}} f_{z,c}(x) \; ,
\end{equation}
\begin{equation}
    V_z(f(x)) = 1 - \max_{c\in\mathcal{C}} f_{z,c}(x) \; ,
\end{equation}
respectively. These are then averaged over the segments $k\in\mathcal{K}(x)$ or over the segment boundary. Moreover, we examine some geometrical properties of the segments, such as their size, \ie the number of pixels $|k|$ contained in $k$, their shape or their position in the image. For in-depth details on the constructed metrics, we refer to \citet{rottmann2019prediction} and the appendix.
By feeding these metrics into a meta regression model, we obtain prediction quality estimates for each segment $k\in\mathcal{K}(x)$, which we denote by $s(k)\in [0,1]$. These quality estimates approach the true segment-wise \emph{Intersection over Union} (IoU) with reasonably high accuracy \citep{rottmann2019prediction}. To fit the meta regressor, we compute the metrics plus the true IoU values of all segments included in the training data of the segmentation network. This meta model is then applied to unseen data, \ie data that was not included in the training of $f$, for the purpose of anomaly segmentation. Here, we consider a segment $k$ to be anomalous, if its quality score is below some predefined threshold $\tau\in [0,1]$, \ie if $s(k) < \tau$. By that, we identify individual segments as unknown, however, the semantic segmentation of unknown objects usually consists of several segments, \ie of different predicted classes. As we can uniquely assign each pixel $z$ to a segment $k(z)$, we obtain a binary pixel-wise classification mask $a \in \{0,1\}^{|\mathcal{H}|\times|\mathcal{W}|}$ via
\begin{equation}
    a_z = \mathbbm{1}_{\{s(k(z)) < \tau\}} ~~ \forall z \in \mathcal{H}\times\mathcal{W} \; , \label{eq:binary}
\end{equation}
where the class label $\mathbbm{1}_{\{s(k(z)) < \tau\}} = 1$ indicates anomalous pixels. Finally, the connected components in the anomaly mask $a$ merge adjacent anomalous segments into suspicious objects. Under ideal conditions,
\begin{enumerate}
    \item the semantic segmentation network performs perfectly on in-distribution data,
    \item the meta model detects all (but only) unknowns, and
    \item novel objects of different classes are separable.
\end{enumerate}

\paragraph{Embedding and Clustering of Image Patches.} 
Image clustering usually takes place in a lower dimensional latent space due to the curse of dimensionality. To this end, we feed image patches tailored to the suspicious objects into an image classification DenseNet201 \cite{huang2018densely}, which is trained on the ImageNet dataset \citep{deng2009imagenet} with 1000 classes. The patches are not equally sized. That nevertheless the DenseNet feature extractor returns features of equal size ($1,920$) for each patch is a consequence of the application of the AdaptiveAvgPool2d layer that is applied as the last layer after the fully convolutional and depthwise interconnected layers of the DenseNet. Put shortly, this last layer pools over both spatial dimension of the feature maps and thereby the output is not dependent on the size of the input, that is transported through the fully convolutional layers.
Their feature representations are further compressed, resulting in a two-dimensional embedding space as illustrated in \cref{fig:overview} (bottom left). 
We apply two commonly used dimensionality reduction techniques. For complexity reasons, we compute the first 50 principal components \citep{FRSLIIIOL} before deploying the better performing \emph{t-SNE} method \citep{Maaten2008VisualizingDU} with Euclidean distance as similarity measure.

This procedure for image embedding is adopted from \citet{oberdiek2020detection}, where the authors evaluated several feature extractors, distance metrics and feature dimensions. We employ the best performing setup in this quantitative analysis to obtain clusters of visually related image patches.
Beyond that, we identify these clusters using the \emph{DBSCAN} \citep{dbscan} algorithm. This clustering method requires two hyperparameters, namely the radius $\varepsilon\in\mathbb{R}$ that defines a neighborhood $B_\varepsilon(\cdot)$ and a threshold $N_\mathrm{min} \in \mathbb{N}$ regarding the number of data points within this $\varepsilon$-neighborhood. Let $\mathcal{E} = \{e_1,e_2,\ldots\} \subset \mathbb{R}^2$ denote the set of the embedded features. Then, an embedding is considered a core point, if and only if it has at least $N_\mathrm{min}$ neighbors, \ie
\begin{align}
    \begin{split}
        e_i \in \mathcal{E} &\text{ is core point } \Leftrightarrow \\
        &|\{e_j\in\mathcal{E}:~e_j\in B_\varepsilon(e_i)\}| \geq N_\mathrm{min} \; .
    \end{split}
\end{align}
The algorithm further distinguishes between border points, \ie embeddings that are not core points themselves, but belong to a core point's neighborhood, and noise else. To mitigate the risk of failures, \ie objects from a different category in the novel clusters, we only consider the core points. We further reject embeddings representing image patches that are smaller than some predefined size. The cluster with the most remaining core points (or all clusters that involve ``enough'' core points) will be used to extend the segmentation network by new classes (\cref{fig:overview}, bottom).

\paragraph{Novelty Segmentation.} 
Using pseudo labels instead of manually annotated targets is a cost-efficient (in the sense of human effort) method of training neural networks on unlabeled data. 
For the sake of simplicity we assume that exactly one cluster is returned by the aforementioned procedure.
For some image $x\in\mathcal{X}$, we denote the predicted segmentation mask by $m(x)$ and the respective segments by $\mathcal{K}(x)$. Let $\mathcal{K}^\mathrm{novel}(x) \subseteq \mathcal{K}(x)$ describe the set of segments $k\in\mathcal{K}(x)$ that are also included in the considered cluster. If $\mathcal{K}^\mathrm{novel}(x)\neq \emptyset$, \ie image $x$ (probably) contains the novel class, we include the tuple $(x,\tilde{y}(x))\in\mathcal{X}\times\{1,\ldots,C+1\}^{|\mathcal{H}|\times|\mathcal{W}|}$ into the re-training data $\mathcal{D}^{C+1}$ for learning the novel class $C+1$. Here, $\tilde{y}(x)$ denotes the pseudo label, where 
\begin{equation}\label{eq:anomalyseg}
    \tilde{y}_z(x) = \begin{cases} C + 1 & \mathrm{,~if~} k(z)\in \mathcal{K}^\mathrm{novel}(x) \\
    m_z(x) & \mathrm{,~otherwise}
    \end{cases} \; ,
\end{equation}
\ie a pixel $z$ is either assigned to the novel class ID $C+1$, or to the class $c\in\mathcal{C}$ that was predicted by the initial model $f$.
An example for acquiring pseudo ground truth for one image is given in \cref{fig:pseudolabel}.
In the following section we extend the segmentation DNN $f$ by fine-tuning it on $\mathcal{D}^{C+1}$.

\section{Extension of the Model's Semantic Space}
\label{sec:incremental-learning}

\begin{figure}[t]
    \captionsetup[subfigure]{labelformat=empty, position=top}
    \centering
    \subfloat[novelty pseudo ground truth]{\includegraphics[width=0.235\textwidth]{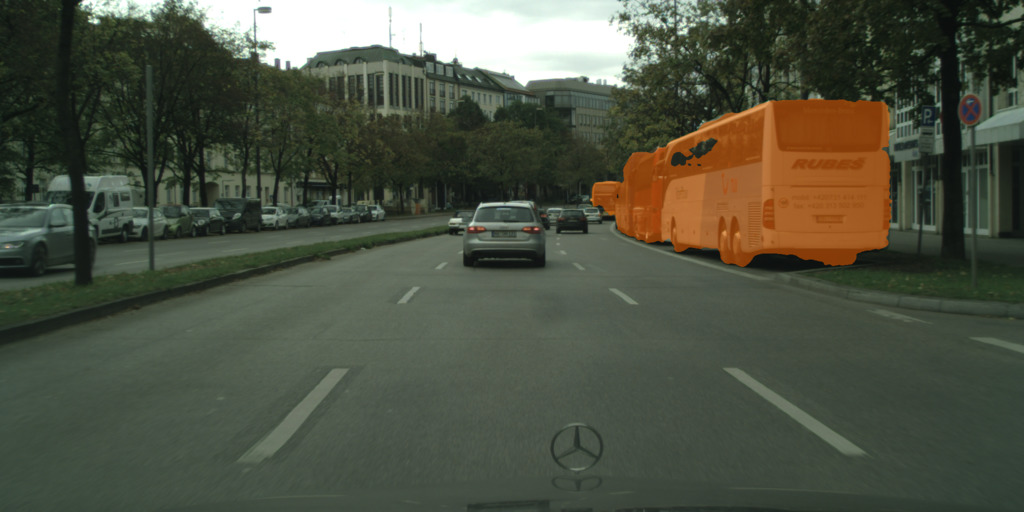}}~
    \subfloat[classes predicted by initial DNN]{\includegraphics[width=0.235\textwidth]{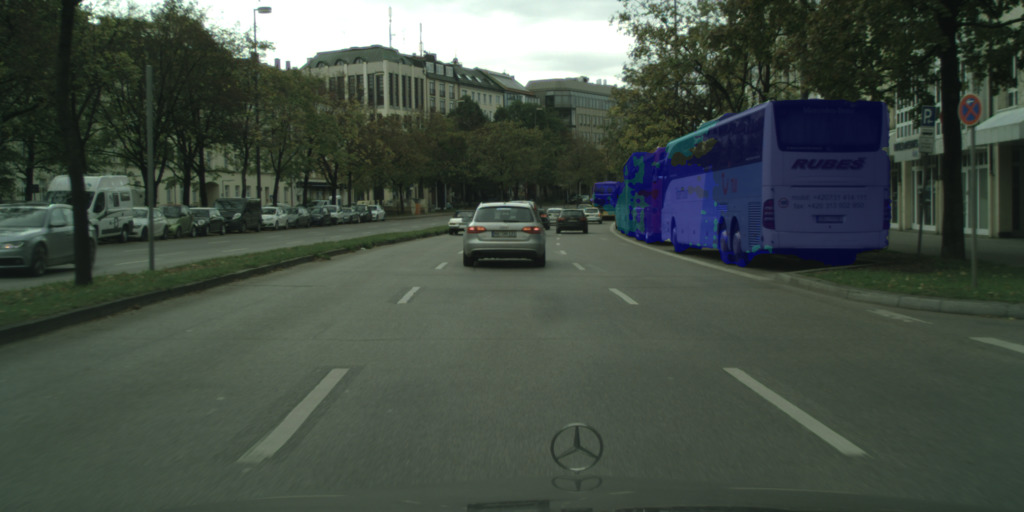}}\\
    \captionsetup[subfigure]{labelformat=empty, position=bottom}
    \subfloat[]{\includegraphics[width=0.33\textwidth]{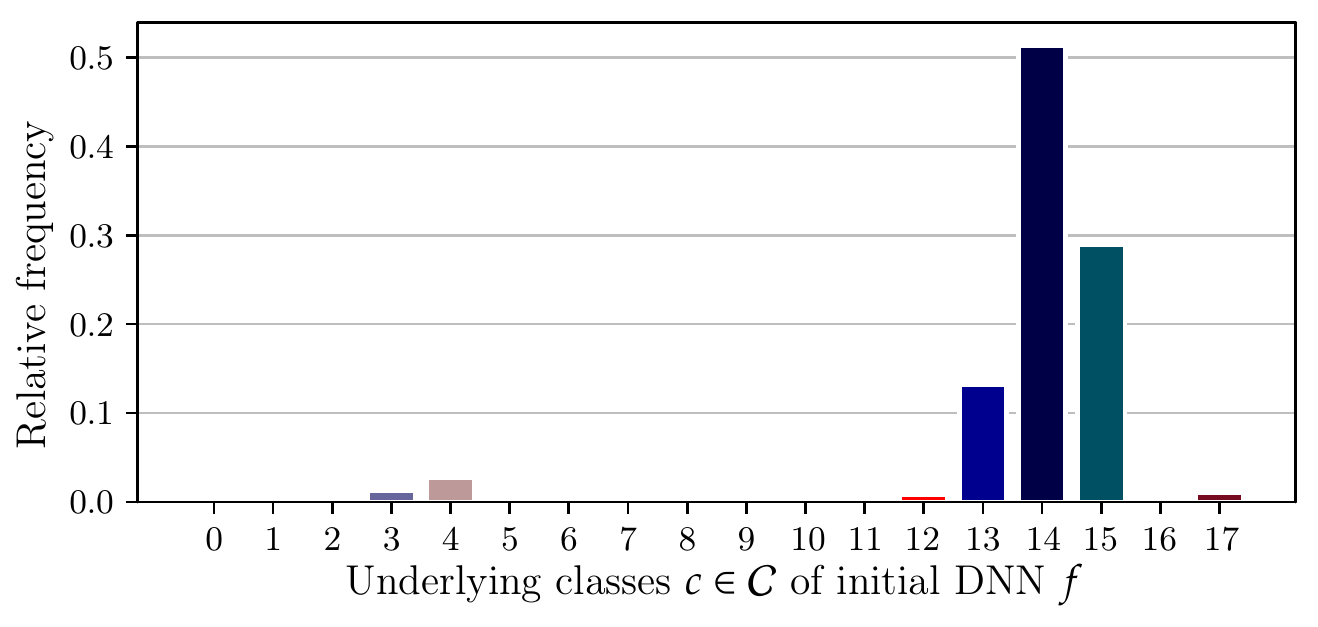}}~ 
    \subfloat[]{\includegraphics[trim={0 -1cm 20cm 0},clip,width=0.14\textwidth]{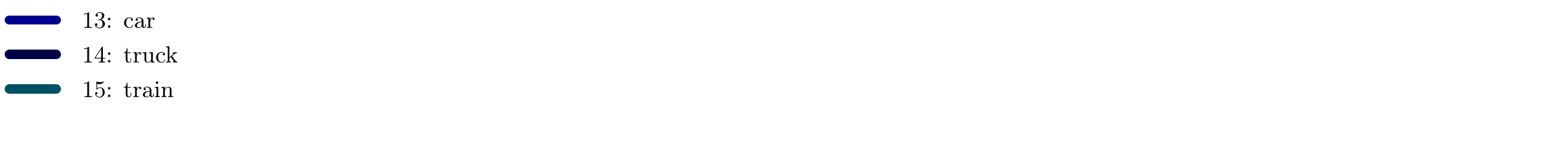}}
    \caption{ Bar plot showing the relative frequencies of predicted classes for instances of the novel class, together with an exemplary image.
    }
    \label{fig:related-classes}
\end{figure}

In this section we describe our approach to semantic incremental learning with the pseudo ground truth acquired by novelty segmentation. Starting from our initial segmentation model $f$, we are seeking an extended model $g:\mathcal{X}\to(0,1)^{|\mathcal{H}|\times|\mathcal{W}|\times (C+1)}$ that retains the knowledge of $f$ while additionally learning the novel class $C+1$. Denote the extended semantic space by $\mathcal{C}^+ = \mathcal{C}\cup\{C+1\}$. In more detail, we replace the ultimate layer of $f$ and reinitialize only the affected weights to obtain the initial model $g$ for re-training, \ie the model we train on the newly collected data $\mathcal{D}^{C+1}$. As loss function we apply a weighted cross entropy loss \citep{1434171}, denoted by $l_{\mathrm{ce},\omega}$. The class-wise weights $\omega_c\in(0,1]$, $c\in\mathcal{C}^+$, are recalculated for each batch based on the inverse class frequency to alleviate class imbalances.

To mitigate the problem of catastrophic forgetting \citep{McCloskey1989CatastrophicII}, we pursue two strategies, namely knowledge distillation \citep{hinton2015distilling} and rehearsal \citep{rehearsal}.

Knowledge distillation in class-incremental learning aims at minimizing variations of the softmax output restricted to only the old classes $c\in\mathcal{C}$. This is realized by an additional distillation loss function \citep{Michieli2021KnowledgeDF} $l_{\mathrm{d}}$, where
\begin{align}
\begin{split}
    l_{\mathrm{d}}(g(x),f(x)) &\\
    := -\frac{1}{|\mathcal{H}||\mathcal{W}|} &\sum_{z\in\mathcal{H}\times\mathcal{W}}\sum_{c\in\mathcal{C}} f_{z,c}(x)\log(g_{z,c}(x)) \; .
\end{split}
\end{align}
Overall, we aim at minimizing the objective
\begin{align}
\label{eq:loss}
\begin{split}
    L :=~\lambda~\mathbb{E}[l_{\mathrm{ce},\omega}&(g(x),\Tilde{y}(x))]\\
    + & ( 1 -  \lambda)~\mathbb{E}[l_\mathrm{d}(g(x),f(x))],~~\lambda\in [0,1]
\end{split}
\end{align}
with $\lambda$ regulating the impact of the distillation loss.

Rehearsal methods propose to replay (some of) the data $\mathcal{D}^\mathrm{train}\subset\mathcal{X}\times\mathcal{C}^{|\mathcal{H}|\times|\mathcal{W}|}$ seen during the training of the initial model $f$. We select a subset $\mathcal{D}^\mathrm{known}\subseteq\mathcal{D}^\mathrm{train}$ that contains as much data as $\mathcal{D}^{C+1}$. This subset is chosen largely at random, but in such a way that it involves classes, that are
\begin{enumerate}
    \item  not or rarely present in $\mathcal{D}^{C+1}$ (class frequency), or
    \item  similar or related to the novel class.
\end{enumerate}
As there is no measure for the second case, we identify those classes by considering the frequency, with which a class is predicted by $f$ on pixels assigned to the novel class. This is, for all data $(x,\Tilde{y}(x))\in\mathcal{D}^{C+1}$ and classes $c\in\mathcal{C}$, we sum up the number of pixels $z\in\mathcal{H}\times\mathcal{W}$ where $\Tilde{y}_z(x)=C+1 \wedge m_z(x) = c$. An example is given in \cref{fig:related-classes}, where the classes \emph{truck}, \emph{train} and \emph{car} are the most frequently predicted classes for instances of the novel class \emph{bus}.

\section{Experimental Setup \& Evaluation}\label{sec:experiments}
We evaluate our approach on the task of detecting and incrementally learning novel classes in traffic scenes, for which there exist large datasets such as Cityscapes \citep{Cordts2016TheCD} and A2D2 \citep{geyer2020a2d2}. To this end, all evaluated segmentation DNN's were trained on a training split and only on a subset of all available classes. We then perform our experiments on a test split of the same dataset on which the DNN was trained in order to extent it by exactly one or even multiple novel classes. We measure the performance of the extended models computing the evaluation metrics \emph{intersection over union} (IoU), \emph{precision} and \emph{recall} for a validation set.

\paragraph{Experimental Setup.}

As segmentation DNNs we employ the DeepLabV3+ \citep{chen2018encoderdecoder} and the PSPNet \citep{Zhao2017PyramidSP}. The first is trained for different subsets of known classes on the Cityscapes dataset. Moreover, both models are pre-trained on Cityscapes with all 19 classes and then fine-tuned on the A2D2 dataset. Here we use a label mapping between both datasets through which 14 classes remain.

We perform five experiments: For the first three experiments, a DeepLabv3+ with a WideResNet38 backbone is trained on the Cityscapes dataset, where 1) the classes \emph{person} \& \emph{rider}, 2) the class \emph{bus} and 3) the classes \emph{person} \& \emph{rider}, \emph{bus} and \emph{car} are excluded. In a fourth experiment, a DeepLabv3+ as well as a PSPNet based on a ResNet50 backbone are fine-tuned on the A2D2 dataset, for which we specified subsets for training, testing and validation, including 2975, 1355 and 451 annotated images, respectively. Then, we also apply our method to the A2D2 dataset without prior fine-tuning, \ie under a domain shift, employing a DeepLabV3+ trained on Cityscapes. 
Our experiments follow a hierarchical structure with increasing complexity:
\begin{enumerate}
    \setlength\itemsep{0mm} 
    \item Construction of a ``well'' separated category (\emph{human}),
    \item Construction of a category in the midst of known similar categories (\emph{bus}),
    \item Construction of multiple novel categories (\emph{human}, \emph{bus} and \emph{car}),
    \item Construction of a new category under domain shift with ground truth for known classes (\emph{guardrail}, with fine-tuning),
    \item Construction of a new category under domain shift without ground truth (\emph{guardrail}, without fine-tuning).
\end{enumerate}
Each of those initial DNNs is employed to predict the semantic segmentation masks for the images contained in the respective test set. For the segment-wise prediction quality estimation introduced in \cref{sec:discovery}, we apply a gradient boosting model to obtain the quality scores $s(k)\in[0,1]$ for each segment $k\in\mathcal{K}(x)$ and image $x$ in the test set. The threshold in  \cref{eq:binary} is set to $\tau = 0.5$, \ie a segment $k\in\mathcal{K}$ is considered as anomalous, if $s(k)<0.5$. To extract features of the suspicious objects, we employ a DenseNet201 \citep{huang2018densely}, trained on the ImageNet dataset \citep{deng2009imagenet} with 1000 classes. Note that the DBSCAN hyperparameters have to be selected dependent on the density of the desired clusters.

For the class-incremental extension of an initial DNN $f$, we replace its final layer to obtain a larger DNN $g$ (see \cref{sec:incremental-learning}). Only the decoder of this model is trained for 70 epochs on the newly collected data $\mathcal{D}^{C+1}$ together with the replayed data $\mathcal{D}^\mathrm{known}$. We use random crops of size $1000\times1000$ pixels, the Adam optimizer with a learning rate of $5\cdot10^{-5}$ and a weight decay of $10^{-4}$. Further, the learning rate is adjusted after every iteration via a polynomial learning rate policy \citep{chen2017deeplab}. The distillation loss and the cross-entropy loss are weighted equally in the overall loss function defined in \cref{eq:loss}, \ie $\lambda=0.5$ (analogously to \citet{michieli2019incremental}).

As the five experiments struggle with different issues, the experimental setup slightly differs. For the first case, we construct the novel category \emph{human}, which is ``well'' separable from all known classes, to enhance the purity of the ``human cluster'' and to simplify the learning of novel objects. However, we observe that the DNN tends to ``overlook'' many humans, \ie they are assigned to the class predicted in the background, \eg to the \emph{road} class. As a consequence, the segment-wise anomaly detection fails to detect such persons, which is why these will be assigned to other classes in our acquired pseudo ground truth. To not distract the extended segmentation network, we modify the pseudo labels by ignoring all known classes $c\in\mathcal{C}$ during the incremental training procedure. 
The \emph{bus} class added in the second experiment is closely related to other classes in the vehicle category, such as \emph{truck}, \emph{train} and \emph{car}, which complicates the construction of pure clusters. We mitigate the impact of objects from similar classes by discarding all objects from the cluster that consist of only one segment in the predicted segmentation. 
Experiment three extends the previous ones by facing multiple unknown classes, namely \emph{human}, \emph{bus} and \emph{car}.
The last two experiments deal with an additional domain shift from urban street scenes in Cityscapes to countryside and highway scenes in A2D2. To bridge this gap, we fine-tune the initial DNN on our A2D2 training set, which, however, requires A2D2 ground truth for the known classes. Without fine-tuning, the prediction quality and thereby the quality of our pseudo ground truth suffers. On that account, we discard images that are generally rated as badly predicted, \ie where the relative amount of pixels with a low quality estimate exceeds $1/3$ of the image in total. Moreover, we renounce the replay of previously-seen data, since this prevents the DNN from adapting to the new domain.

\paragraph{Evaluation of Results.}

\begin{table}[t]
    \centering
    \begin{adjustbox}{width=0.44\textwidth}
    \begin{tabular}{l||cc|c}
        \hline
        Model & mIoU$_\mathcal{C}$ & IoU$_\mathrm{novelty}$ & mIoU$_{\mathcal{C}^+}$ \\\hline\hline
        \rowcolor{maroon!10} \textbf{1.\ experiment:} Cityscapes, human & \multicolumn{3}{c}{DeepLabV3+}\\\hline\hline
        initial DNN & 68.63 & 00.00 & 64.82 \\
        \rowcolor{Gray} extended DNN (ours) & 68.53 & 39.80 & 66.94 \\
        extended DNN (supervised) & 69.43 & 59.33 & 68.87  \\
        oracle & 71.05  & 72.85 & 71.15 \\\hline\hline
        \rowcolor{maroon!10} \textbf{2.\ experiment:} Cityscapes, bus & \multicolumn{3}{c}{DeepLabV3+}\\\hline\hline
        initial DNN & 66.94 & 00.00 & 63.42 \\
        \rowcolor{Gray} extended DNN (ours) & 67.07 & 44.73 & 65.89 \\
        extended DNN (supervised) & 66.74 & 41.40 & 65.41 \\
        oracle & 69.48 & 76.66 & 69.86 \\\hline\hline
        \rowcolor{maroon!10} \textbf{3.\ experiment:} Cityscapes, multi & \multicolumn{3}{c}{DeepLabV3+}\\\hline\hline
        initial DNN & 56.99 & 00.00 \& 00.00 & 50.29 \\
        \rowcolor{Gray} extended DNN (ours) & 57.52 & 40.22 \& 81.27 & 57.90 \\
        oracle & 77.28 & 81.90 \& 94.94 & 78.59 \\\hline\hline
        \rowcolor{maroon!10} \textbf{4.\ experiment (a):} A2D2, guardrail & \multicolumn{3}{c}{DeepLabV3+ (fine-tuned)}\\\hline\hline
        initial DNN & 75.77 & 00.00 & 70.72 \\
        \rowcolor{Gray} extended DNN (ours) & 72.07 & 46.10 & 70.34 \\
        oracle & 75.23 & 74.58 & 75.19 \\\hline\hline
        \rowcolor{maroon!10} \textbf{4.\ experiment (b):} A2D2, guardrail & \multicolumn{3}{c}{PSPNet (fine-tuned)}\\\hline\hline
        initial DNN & 68.77 & 00.00 & 64.19 \\
        \rowcolor{Gray} extended DNN (ours) & 64.54 & 32.79 & 62.42  \\
        oracle & 67.71 & 69.08 & 67.80 \\\hline\hline
        \rowcolor{maroon!10} \textbf{5.\ experiment:} A2D2, guardrail & \multicolumn{3}{c}{DeepLabV3+ (not fine-tuned)}\\\hline\hline
        initial DNN & 59.38 & 00.00 & 55.42 \\
        \rowcolor{Gray} extended DNN (ours) & 60.48 & 20.90 & 57.84  \\\hline
    \end{tabular}
    \end{adjustbox}
    \caption{Comparing overview of all evaluated models, where the results for our extended DNNs are highlighted in gray. As performance metrics, we provide the mean IoU over the old and new classes, denoted by mIoU$_\mathcal{C}$ and mIoU$_{\mathcal{C}^+}$, respectively, and the IoU value of the novel class(es), IoU$_\mathrm{novelty}$.}
    \label{tab:comparison-results}
    
\end{table}

\begin{table}[t]
    \centering
    \begin{adjustbox}{width=0.47\textwidth}
    \begin{tabular}{l||ccc|ccc}
        \hline
          & IoU  & precision  & recall & IoU  & precision  & recall\\\hline\hline
        \rowcolor{maroon!10} \textbf{1.\ experiment:} & \multicolumn{6}{c}{DeepLabV3+}\\\hhline{~|---|---}
        \rowcolor{maroon!10} Cityscapes, human &\multicolumn{3}{c|}{initial} & \multicolumn{3}{c}{extended}\\\hline\hline
        \rowcolor{Gray} human &  00.00 & 00.00 & 00.00 & 39.80 & 60.60 & 53.72 \\ \hline
        mean over $\mathcal{C}$ &  68.63 & 79.79 & 80.94 & 68.53 & 83.32 & 77.17 \\ \hline
        mean over ${\mathcal{C}^+}$ &  64.82 & 75.36 & 76.44 & 66.94 & 82.05 & 75.86 \\        \hline\hline
        \rowcolor{maroon!10} \textbf{2.\ experiment:} & \multicolumn{6}{c}{DeepLabV3+}\\\hhline{~|---|---}
        \rowcolor{maroon!10} Cityscapes, bus &\multicolumn{3}{c|}{initial} & \multicolumn{3}{c}{extended}\\\hline\hline
        \rowcolor{Gray} bus &  00.00 & 00.00 & 00.00 & 44.73 & 58.33 & 66.15 \\ \hline
        mean over $\mathcal{C}$ & 66.94 & 79.32 & 79.55 & 67.07 & 82.46 & 76.31 \\\hline
        mean over ${\mathcal{C}^+}$ & 63.42 & 75.15 & 75.36 & 65.89 & 81.19 & 75.78 \\\hline\hline
        \rowcolor{maroon!10} \textbf{3.\ experiment:} & \multicolumn{6}{c}{DeepLabV3+}\\\hhline{~|---|---}
        \rowcolor{maroon!10} Cityscapes, multi &\multicolumn{3}{c|}{initial} & \multicolumn{3}{c}{extended}\\\hline\hline
        \rowcolor{Gray} human &  00.00 & 00.00 & 00.00 & 40.22 & 68.74 & 49.65 \\ \hline
        \rowcolor{Gray} car &  00.00 & 00.00 & 00.00 & 81.27 & 86.56 & 93.05 \\ \hline
        mean over $\mathcal{C}$ & 56.99 & 65.75 & 80.88 & 57.52 & 78.53 & 65.77 \\\hline
        mean over ${\mathcal{C}^+}$ & 50.29 & 58.01 & 71.37 & 57.90 & 78.43 & 66.43 \\\hline\hline
        \rowcolor{maroon!10} \textbf{4.\ experiment (a):} & \multicolumn{6}{c}{DeepLabV3+}\\\hhline{~|---|---}
        \rowcolor{maroon!10} A2D2, guardrail &\multicolumn{3}{c|}{initial} & \multicolumn{3}{c}{extended}\\\hline\hline
        \rowcolor{Gray} guardrail &  00.00 & 00.00 & 00.00 & 46.10 & 80.41 & 52.09 \\ \hline
        mean over $\mathcal{C}$ & 75.77 & 87.86 & 83.47 & 72.07 & 89.01 & 78.44 \\\hline
        mean over ${\mathcal{C}^+}$ & 70.72 & 82.00 & 77.90 & 70.34 & 88.44 & 76.69 \\\hline\hline
        \rowcolor{maroon!10} \textbf{4.\ experiment (b):} & \multicolumn{6}{c}{PSPNet}\\\hhline{~|---|---}
        \rowcolor{maroon!10} A2D2, guardrail &\multicolumn{3}{c|}{initial} & \multicolumn{3}{c}{extended}\\\hline\hline
        \rowcolor{Gray} guardrail &  00.00 & 00.00 & 00.00 & 32.79 & 70.75 & 38.04 \\ \hline
        mean over $\mathcal{C}$ & 68.77 & 84.57 & 76.79 & 64.54 & 86.41 & 71.22 \\\hline
        mean over ${\mathcal{C}^+}$ & 64.19 & 78.93 & 71.67 & 62.42 & 85.36 & 69.01 \\\hline\hline
        \rowcolor{maroon!10} \textbf{5.\ experiment:}& \multicolumn{6}{c}{DeepLabV3+}\\\hhline{~|---|---}
        \rowcolor{maroon!10} A2D2, guardrail &\multicolumn{3}{c|}{initial} & \multicolumn{3}{c}{extended}\\\hline\hline
        \rowcolor{Gray} guardrail &  00.00 & 00.00 & 00.00 & 20.90 & 77.12 & 22.32 \\ \hline
        mean over $\mathcal{C}$ &  59.38 & 79.50 & 68.14 & 60.48 & 84.08 & 66.61 \\ \hline
        mean over ${\mathcal{C}^+}$ & 55.42 & 74.20 & 63.60 & 57.84 & 83.61 & 63.66 \\ \hline
    \end{tabular}
    \end{adjustbox}
    \caption{Direct comparison of the initial and the extended DNNs for all conducted experiments. We report the IoU, precision and recall values for the novel class (highlighted with gray rows), respectively, as well as averaged over the previously-known and the extended class spaces $\mathcal{C}$ and $\mathcal{C}^+$.}
    \label{tab:detailed-results}
\end{table}

In the following, all evaluation values belonging to our extended models are averaged over five runs of the respective experiment. For in-depth details we refer to the appendix. We provide a qualitative comparison of different models for all conducted experiments in \cref{tab:comparison-results}, reporting the mean IoU over the known classes and over the extended class set, denoted as mIoU$_\mathcal{C}$ and  mIoU$_{\mathcal{C}^+}$, respectively, as well as the IoU value of the novel classes (IoU$_\mathrm{novelty}$). The models considered in this comparison are the initial and the extended DNN, where the class space is extended via our method. For the first and second experiment we further compare our approach with a baseline, where a DNN is extended using a self-training approach. That is, we employ a so-called teacher network, which is already trained on the extended semantic space $\mathcal{C}^+$, to produce pseudo labels for some student network. Thereby, we obtain a high quality pseudo ground truth. Apart from this, the baseline DNN is extended analogously to ours. In addition, for the first four experiments we provide results of an \emph{oracle}, \ie a DNN, that is initially trained on the extended class set $\mathcal{C}^+$ and only with human-annotated ground truth. 
In the fifth experiment, we extend the initial DNN by a novel class derived from a different dataset. To some extent, the oracle from experiment four (a) can serve as a coarse reference for experiment five.
In \cref{tab:detailed-results} we give a more detailed overview about all experiments, reporting not only the IoU, but also the precision and recall values of the novel class as well as averaged over $\mathcal{C}$ and $\mathcal{C}^+$. Note that the fourth experiment is evaluated twice, once for (a) the DeepLabV3+ and once for (b) the PSPNet. For class-wise evaluation results and visualizations, we refer to \cref{sec:models}.

In general, we observe that our approach succeeds in incrementally extending a DNN by a novel class, while the performance on previously-known classes remains stable. On Cityscapes, we achieve IoU values for the novel classes human and bus of IoU$_\mathrm{human}=39.80 \pm 0.73 \%$ and IoU$_\mathrm{bus}=44.73 \pm 1.46\%$, respectively. For the third experiment with two novel classes, we obtain similar results for the \emph{human} class with IoU$_\mathrm{human}=40.22 \pm 1.77 \%$ and for the \emph{car} class even IoU$_\mathrm{car}=81.27 \pm 1.16\%$. While these IoU values are a considerable achievement for a method working without ground truth, the distinct gaps to the oracle's IoU values still leave room for further improvement. Compared to the baseline DNN, we do not achieve competitive performance in the first experiment, while in the second experiment, our approach actually performs slightly better.
This is explained by the fact, that the pseudo ground truth for the \emph{human} class incorporates much more noise than that for the \emph{bus} class.
In the fourth experiment we mitigate the domain shift from Cityscapes to A2D2 by prior fine-tuning of the networks, using A2D2 ground truth. By that, we obtain IoU values of IoU$_\mathrm{guardrail}=46.10 \pm 4.8\%$ for the DeepLabV3+ and IoU$_\mathrm{guardrail}=32.79 \pm 3.48\%$ for the PSPNet. We conclude, that our approach achieves better results for models which are initially better-performing. Without fine-tuning the DeepLabV3+ on A2D2, we obtain IoU$_\mathrm{guardrail}=20.90 \pm 1.73\%$, while the mean IoU over the previously-known classes $\mathcal{C}$ slightly increases from $59.38\%$ to $60.48 \pm 0.47\%$.

\section{Conclusion \& Outlook}
In this work, we have introduced a new and modular procedure for the class-incremental extension of a semantic segmentation network, where novel classes are detected, annotated and learned in an unsupervised fashion. While there already exists an unsupervised open world approach for semantic segmentation \citep{Nakajima2019IncrementalCD}, we are the first in this field to extend a neural network's semantic space by robust novel classes. We performed five hierarchically structured experiments with an increasing level of difficulty. We demonstrated that our approach can deal with novelties that are either ``well'' separated or related to known categories, and that it is even applicable when the test data is sampled from a slightly different distribution than the DNN was trained on. Moreover, we applied two different models in the fourth experiment, where the initial DeepLabV3+ already outperformed the initial PSPNet. This performance gap is also reflected in the model's ability to learn the novel class, thus we conclude that our method benefits significantly from high performance networks.

For future work, we plan to improve the extension of a neural network by multiple classes at once. On that account, suitable datasets are in demand. Two datasets for the task of anomaly segmentation were recently published in \citet{chan2021segmentmeifyoucan}, however, these show a wide variety of anomalous objects. To advance the research in class-incremental learning, it requires datasets where novel objects, \ie objects that do not appear in the training data, appear frequently in the test data. 

We are currently working on a synthetic dataset tailored to our approach. This data is generated using the CARLA 0.9.12 simulator \cite{Dosovitskiy17}, similar as extensively described in \cite{kowol2022aeye}. The data include annotated street scene images, generated on the same maps for training and testing. Since we aim at detecting novel classes in the test data, these images are enriched by several \textbf{never-seen} object classes, \eg \emph{deer}, \emph{construction vehicle} or \emph{portable toilet} (examples provided in  \cref{sec:carla-dataset}).

Besides, we plan to adapt our approach to video instead of image data, where anomaly detection includes anomaly tracking over multiple frames.

Our source code is publicly available on github under \href{https://github.com/SUhlemeyer/novelty-learning}{https://github.com/SUhlemeyer/novelty-learning}.

\section{Limitations \& Negative Impact}

With the procedure presented in this work, we are taking a first step towards a new machine learning problem. This first step is highly experimental and our method has not the technology readiness level to be applied to real-world problems in a fully automated fashion. Especially from the safety point of view, a neural network should not be modified without any supervision, since we can not guarantee to avoid significant performance drops.

\begin{acknowledgements} 
    This work is funded by the German Federal Ministry for Economic Affairs and Energy, within the project ``KI Delta Learning'', grant no.\ 19A19013Q. We thank the consortium for the successful cooperation. The authors gratefully also acknowledge the
    Gauss Centre for Supercomputing e.V. (\href{https://www.gausscentre.eu}{https://www.gausscentre.eu}) for funding this project by providing computing time through the John von Neumann Institute for Computing (NIC) on the GCS Supercomputer JUWELS at Jülich Supercomputing Centre (JSC).
\end{acknowledgements}

\bibliography{uai2022-template}
\newpage
\appendix

\section{Evaluated Models}\label{sec:models}

We performed six experiments that differ in terms of underlying datasets, network architectures and novelties. In this section we provide a class-wise evaluation of each initial and extended DNN, as well as example images for all evaluated models, \ie also for the baseline and the oracle DNNs. For the extended models, we report the mean and standard deviation of the evaluation metrics for five runs, respectively, using the random seeds 14, 123, 666, 375 and 693.

\subsection{Experiment 1}

\begin{figure}[t]
    \captionsetup[subfigure]{labelformat=empty, position=top}
    \centering
    \captionsetup[subfigure]{labelformat=empty, position=bottom}
    \subfloat[]{\includegraphics[width=0.33\textwidth]{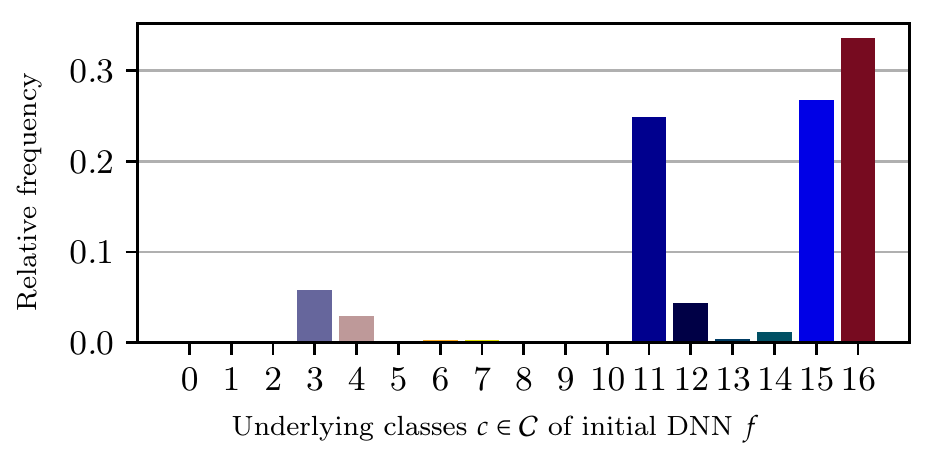}}~ 
    \subfloat[]{\includegraphics[trim={0 -1cm 20cm 0},clip,width=0.14\textwidth]{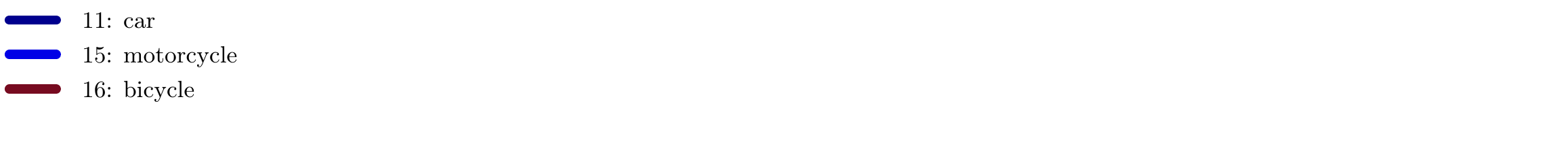}}
    \vspace{-0.75cm}
    \caption{ Bar plot showing the relative frequencies of predicted classes for instances of the novel class \emph{human}.
    }
    \label{fig:related-classes-human}
\end{figure}

\begin{figure}[t]
    \captionsetup[subfigure]{labelformat=empty, position=top}
    \centering
    \captionsetup[subfigure]{labelformat=empty, position=bottom}
    \subfloat[\centering image patch]{\includegraphics[width=0.15\textwidth]{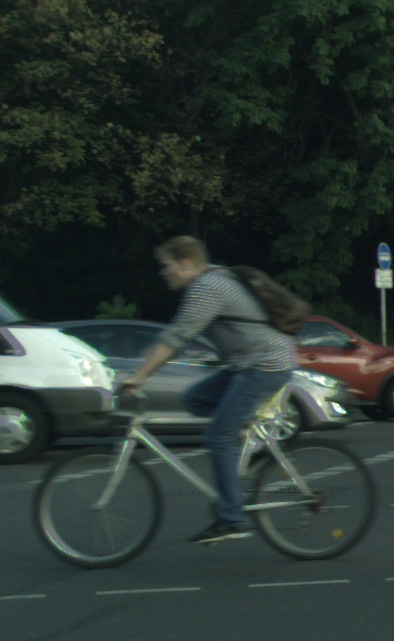}}~ 
    \subfloat[\centering predicted segmentation]{\includegraphics[width=0.15\textwidth]{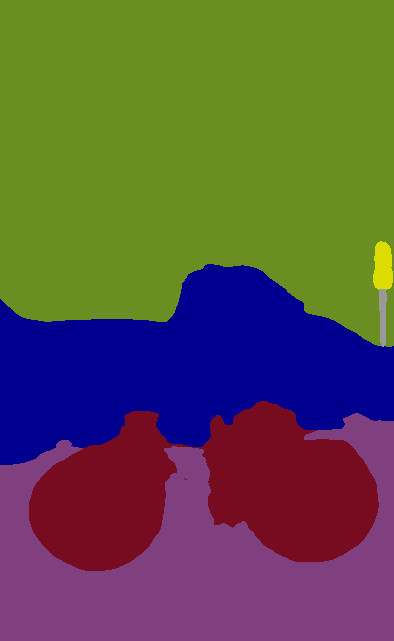}}~ 
    \subfloat[\centering quality estimation]{\includegraphics[width=0.15\textwidth]{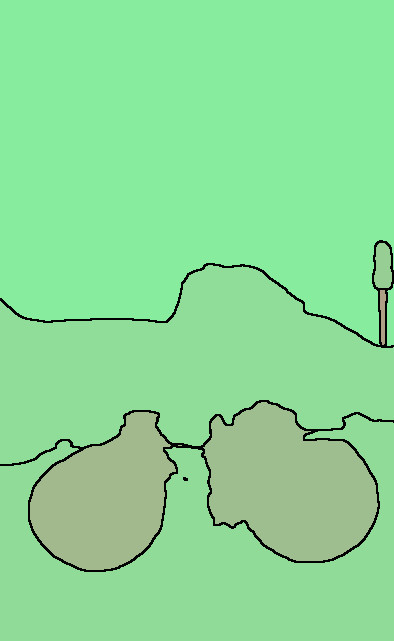}}
    \caption{Image patch, semantic segmentation and prediction quality estimation for a scene, where a cyclist is overlooked by the initial DNN.
    }
    \label{fig:overlooked-human}
\end{figure}

For the first experiment, we trained a DeepLabV3+ on the Cityscapes dataset, excluding the classes \emph{pedestrian} and \emph{rider}, both together constituting the class \emph{human}. This novelty is well separable from all the known classes as these belong to different, non-organic categories. As there are no similar classes, humans are either totally ``overlooked'' by the segmentation DNN, \ie assigned to the class predicted in their background, or predicted as related classes, \eg as \emph{bicycle}, \emph{motorcycle} or \emph{car} (cf.~\cref{fig:related-classes-human}). Since our anomaly detection method fails to spot overlooked persons, these remain mislabeled even in the pseudo ground truth, thus negatively affecting the incremental training procedure. For an example we refer to \cref{fig:overlooked-human}, where a cyclist is assigned to the background classes \emph{road} and \emph{car}. To prevent this issue, we ignore all known classes $c\in\mathcal{C}$ present in the pseudo labels. Our newly collected data $\mathcal{D}^{C+1}$ contains 76 pseudo-labeled images. The replayed training data is selected such that at least 25\% - 35\% of the images contain cars, motorcycles and bicycles, respectively.

We evaluated the initial and the extended DNN on the Cityscapes validation data. Class-wise results are provided in \cref{tab:cs_human}. Besides the novel class, which achieves an IoU value of nearly 40\% with approximately 50-60\% precision and recall, the incremental training has only little impact on previously-known classes. For many classes, however, we observe an improvement in precision at the expense of the corresponding recall values, \eg for the classes \emph{fence}, \emph{truck} and \emph{train}. This is also reflected in the mean precision and recall values over $\mathcal{C}$, \ie while precision increases by 3.53\%, recall decreases by 3.77\%. Especially the classes \emph{motorcycle} and \emph{bicycle} gain performance regarding the IoU and precision, which is mainly due to human pixels initially assigned to those classes, while the proportion of bikes (motor- or bicycles) that are predicted correctly drops significantly. 

\begin{figure*}[t]
    \captionsetup[subfigure]{labelformat=empty}
    \centering
    \subfloat[image \& annotation]{\includegraphics[width=0.19\textwidth]{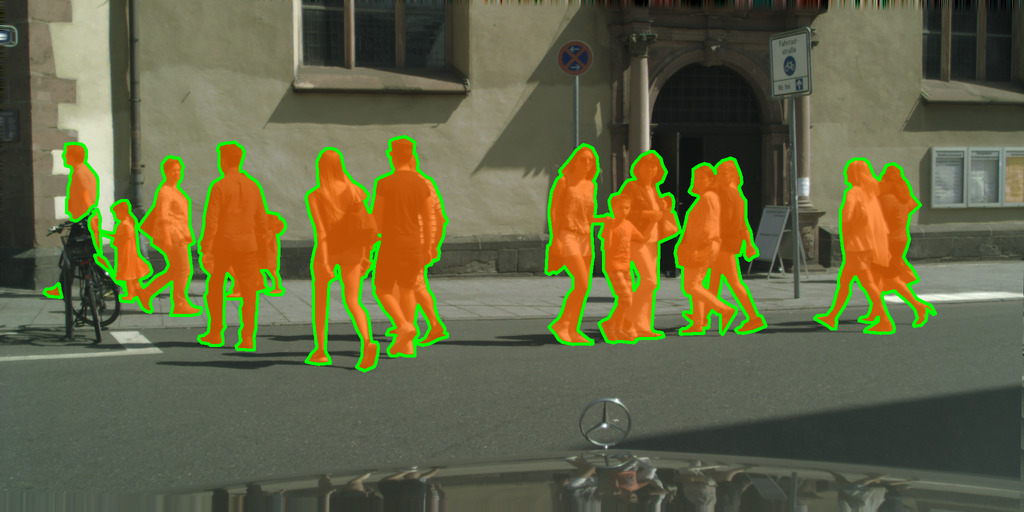}}~
    \subfloat[initial DNN]{\includegraphics[width=0.19\textwidth]{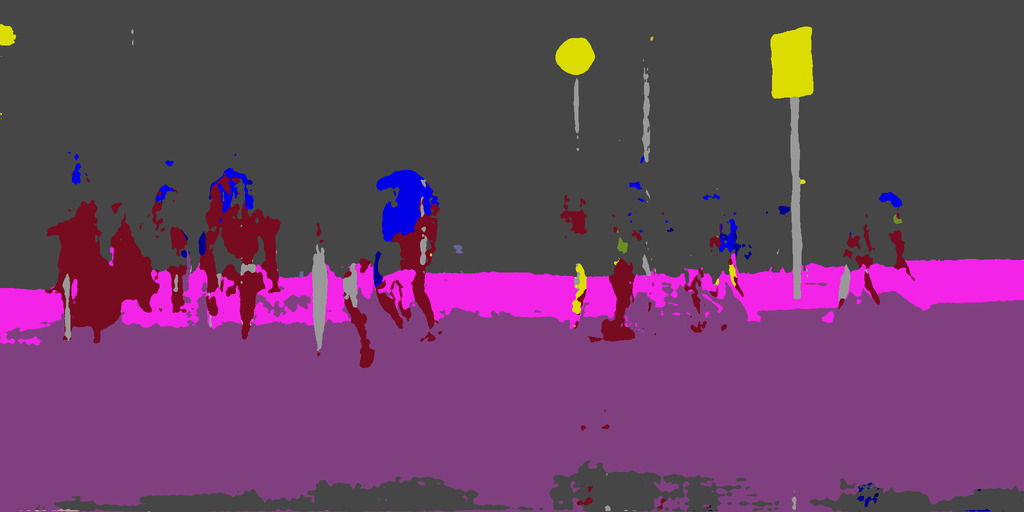}}~
    \subfloat[extended DNN]{\includegraphics[width=0.19\textwidth]{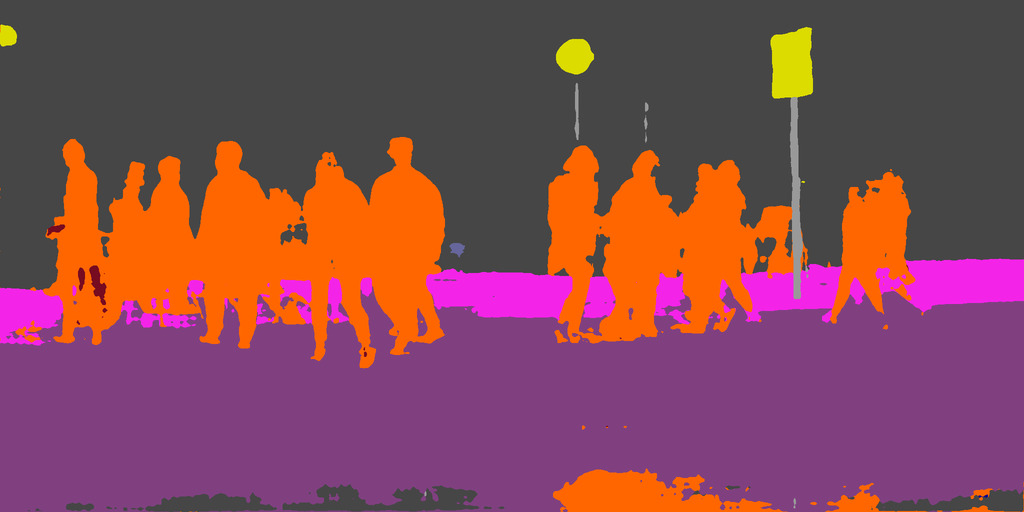}}~
    \subfloat[baseline]{\includegraphics[width=0.19\textwidth]{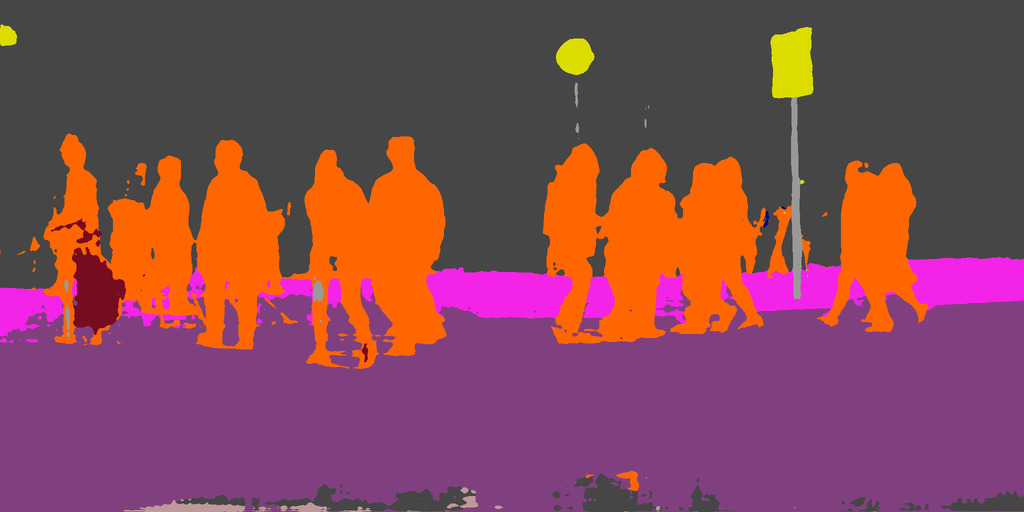}}~
    \subfloat[oracle]{\includegraphics[width=0.19\textwidth]{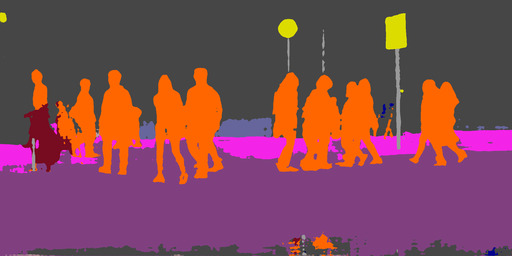}}
    \caption{Comparison of the semantic segmentation predictions of all DNNs evaluated in the first experiment for an exemplary scene from the Cityscapes validation data. 
    }
    \label{fig:results-1}
\end{figure*}

\begin{table}[t]
    \centering
    \resizebox{0.47\textwidth}{!}{
    \begin{tabular}{l||ccc|ccc}
        \hline
        \rowcolor{maroon!10} \textbf{1.\ experiment} & \multicolumn{6}{c}{DeepLabV3+} \\\hhline{~||------}
        \rowcolor{maroon!10} Cityscapes, human & \multicolumn{3}{c|}{initial} & \multicolumn{3}{c}{extended}\\\hline\hline
        Class  & IoU  & precision  & recall & IoU  & precision  & recall\\\hline\hline
        road  &  97.34 & 98.35 & 98.96 & 97.43 $\pm$ 0.05 & 98.54 $\pm$ 0.12 & 98.86 $\pm$ 0.08 \\ \hline
        sidewalk & 80.63 & 89.39 & 89.16 & 80.51 $\pm$ 0.23 & 89.50 $\pm$ 0.50 & 88.91 $\pm$ 0.67 \\ \hline
        building  & 88.91 & 92.80 & 95.50 & 89.40 $\pm$ 0.05 & 93.42 $\pm$ 0.20 & 95.42 $\pm$ 0.24 \\ \hline
        wall & 47.24 & 74.57 & 56.32 & 47.74 $\pm$ 0.57 & 78.92 $\pm$ 0.49 & 54.71 $\pm$ 0.77 \\ \hline
        fence  & 51.03 & 66.76 & 68.41 & 49.20 $\pm$ 0.44 & 70.06 $\pm$ 1.55 & 62.33 $\pm$ 1.26 \\ \hline
        pole  & 52.90 & 72.68 & 66.02 & 53.30 $\pm$ 0.39 & 74.42 $\pm$ 1.41 & 65.31 $\pm$ 1.64 \\ \hline
        traffic light  &  55.44 & 75.04 & 67.98 & 55.33 $\pm$ 0.19 & 75.49 $\pm$ 1.24 & 67.47 $\pm$ 1.21 \\ \hline
        traffic sign  &  66.66 & 86.22 & 74.61 & 66.32 $\pm$ 0.62 & 87.54 $\pm$ 1.41 & 73.27 $\pm$ 1.67 \\ \hline
        vegetation  & 89.95 & 93.60 & 95.85 & 90.15 $\pm$ 0.03 & 94.01 $\pm$ 0.22 & 95.65 $\pm$ 0.22 \\ \hline
        terrain  &  56.29 & 77.66 & 67.17 & 55.29 $\pm$ 0.47 & 75.88 $\pm$ 1.67 & 67.14 $\pm$ 1.77 \\ \hline
        sky  &  93.76 & 96.38 & 97.18 & 93.60 $\pm$ 0.11 & 96.01 $\pm$ 0.26 & 97.39 $\pm$ 0.19 \\ \hline
        \rowcolor{Gray} human &  00.00 & 00.00 & 00.00 & 39.80 $\pm$ 0.73 & 60.60 $\pm$ 1.20 & 53.72 $\pm$ 1.42 \\ \hline
        car  &  90.61 & 92.97 & 97.27 & 91.16 $\pm$ 0.21 & 95.25 $\pm$ 0.50 & 95.50 $\pm$ 0.47 \\ \hline
        truck  &  69.66 & 80.23 & 84.09 & 68.98 $\pm$ 0.56 & 84.92 $\pm$ 2.35 & 78.70 $\pm$ 1.97 \\ \hline
        bus  &  76.90 & 88.59 & 85.35 & 71.57 $\pm$ 0.60 & 87.25 $\pm$ 1.33 & 79.95 $\pm$ 1.15 \\ \hline
        train  &  70.35 & 83.33 & 81.87 & 63.11 $\pm$ 3.17 & 89.63 $\pm$ 1.61 & 68.13 $\pm$ 3.93 \\ \hline
        motorcycle  &  24.45 & 28.57 & 62.92 & 32.92 $\pm$ 1.13 & 53.91 $\pm$ 2.07 & 45.89 $\pm$ 2.21 \\ \hline
        bicycle  &  54.57 & 59.30 & 87.24 & 59.01 $\pm$ 0.61 & 71.62 $\pm$ 2.43 & 77.20 $\pm$ 3.38 \\ \hline\hline
        mean over $\mathcal{C}$ &  68.63 & 79.79 & 80.94 & 68.53 $\pm$ 0.27 & 83.32 $\pm$ 0.28 & 77.17 $\pm$ 0.60 \\ \hline
        mean over ${\mathcal{C}^+}$ &  64.82 & 75.36 & 76.44 & 66.94 $\pm$ 0.27 & 82.05 $\pm$ 0.25 & 75.86 $\pm$ 0.55 \\ \hline
    \end{tabular}}
    \caption{In-depth evaluation on the Cityscapes validation data for the first experiment, where we incrementally extend a DeepLabV3+ by the novel class \emph{human} on the Cityscapes dataset. We provide IoU, precision and recall values obtained for both, the initial and the extended DNN, on a class-level as well as averaged over the classes in $\mathcal{C}$ and $\mathcal{C}^+$, respectively.}
    \label{tab:cs_human}
\end{table}

A comparison of all evaluated models in the first experiment is illustrated for an example image in \cref{fig:results-1}. We observe a reduction of noise in the model's predictions, starting from the initial DNN, to the extended DNN, the baseline and the oracle. Nonetheless, the predicted segmentation of our extended DNN comes close to those predicted by the comparative models that both require ground truth for the novel class.

\subsection{Experiment 2}

The setup of the second experiment is the same as in the first one (DeepLabV3+, Cityscapes dataset), but excluding busses from the set of known classes instead of humans. This novelty belongs to the vehicle category, thus being akin to other vehicle classes as \emph{train} or \emph{truck}. These are also the classes the objects declared as novel were predicted for the most part, as we illustrated in \cref{fig:related-classes}. On that account, at least 50\% of the 55 images in $\mathcal{D}^{C+1}$ contain trucks, 30\% trains. As a consequence of the visual relatedness, trucks and trains that exhibit a low prediction quality, \ie that are treated as anomalies, contaminate the cluster of busses in the two-dimensional embedding space. We observed, that the segmentation network predicts most of these ``detected'' trucks and trains correctly, while it assigns multiple classes, \ie multiple segments in the semantic segmentation prediction, to a bus. Thus, we delete anomalies from the embedding space, whose predicted segmentation consists of only one segment (ignoring segments with less than 500 pixels).

\begin{figure*}[h]
    \captionsetup[subfigure]{labelformat=empty}
    \centering
    \subfloat[image \& annotation]{\includegraphics[width=0.19\textwidth]{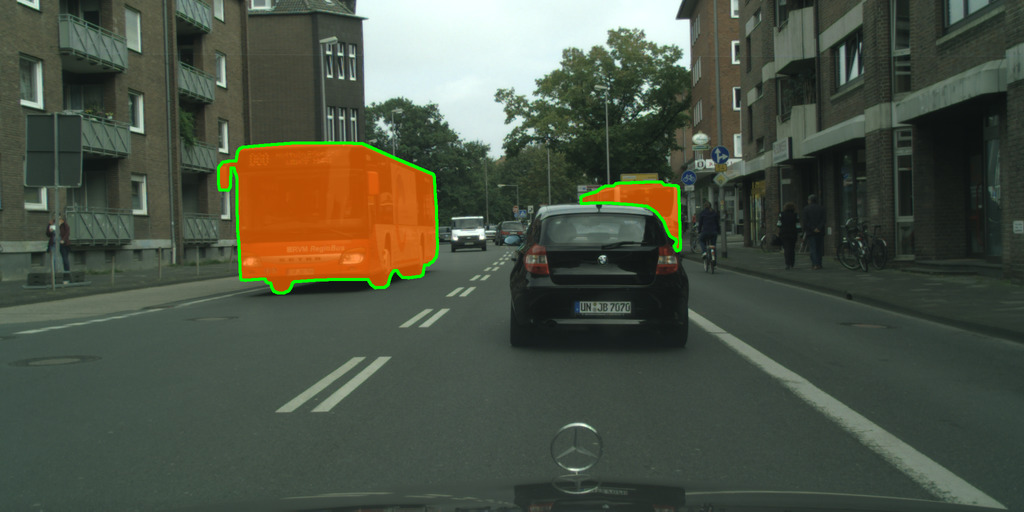}}~
    \subfloat[initial DNN]{\includegraphics[width=0.19\textwidth]{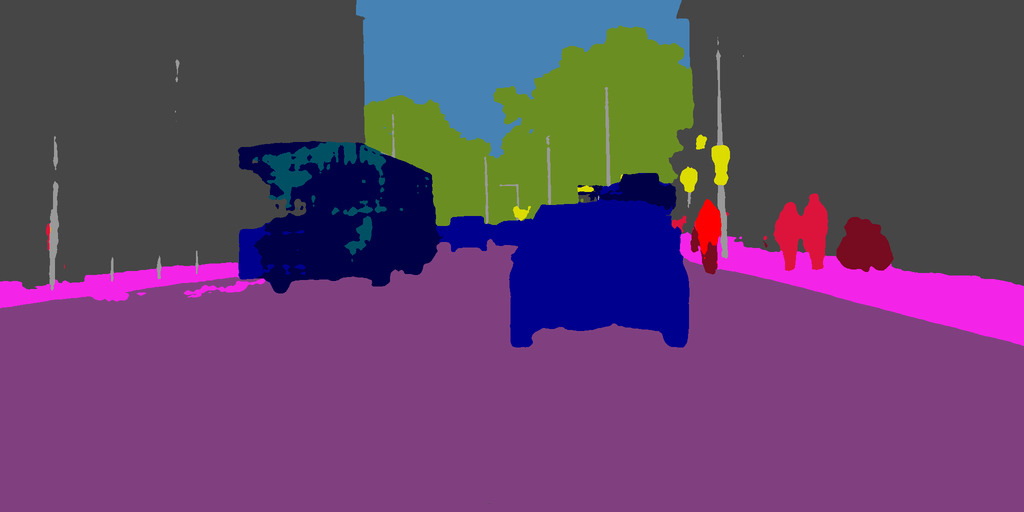}}~
    \subfloat[extended DNN]{\includegraphics[width=0.19\textwidth]{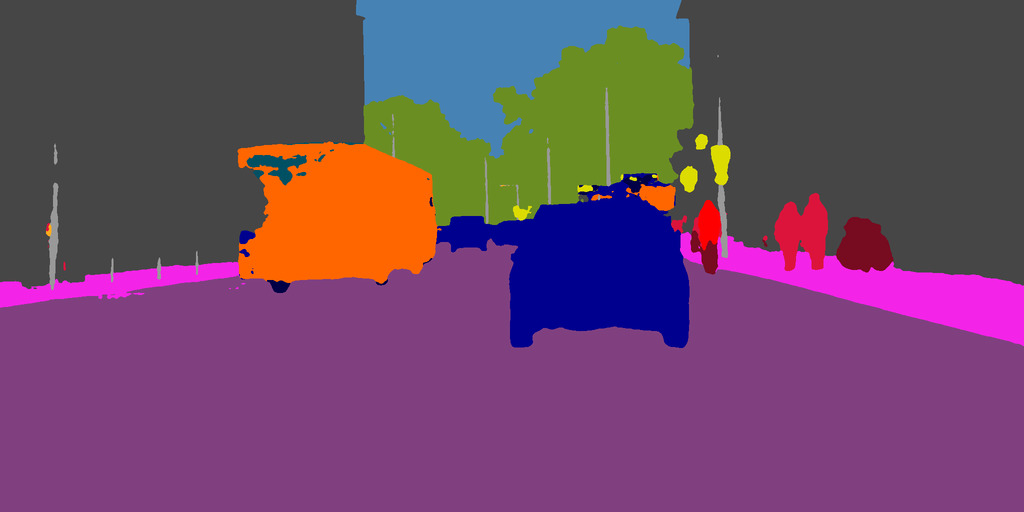}}~
    \subfloat[baseline]{\includegraphics[width=0.19\textwidth]{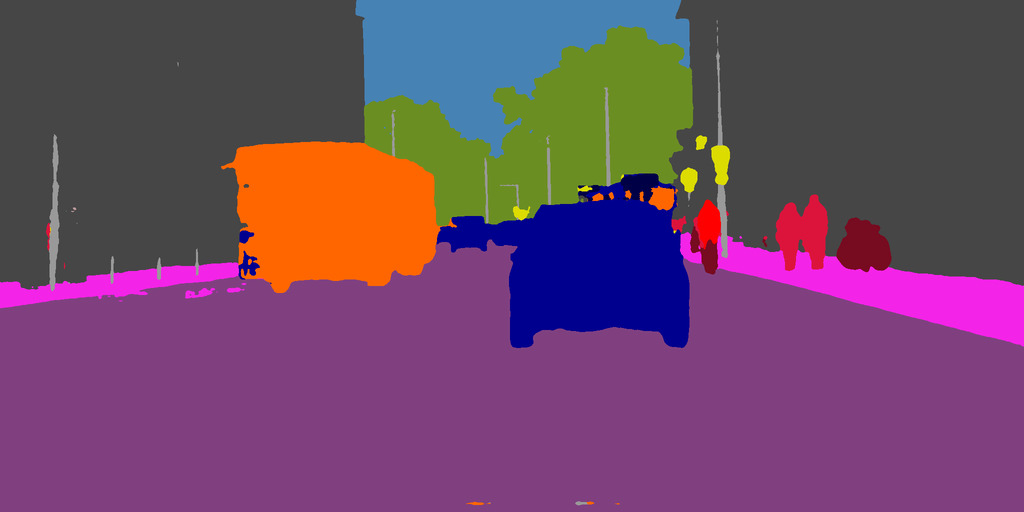}}~
    \subfloat[oracle]{\includegraphics[width=0.19\textwidth]{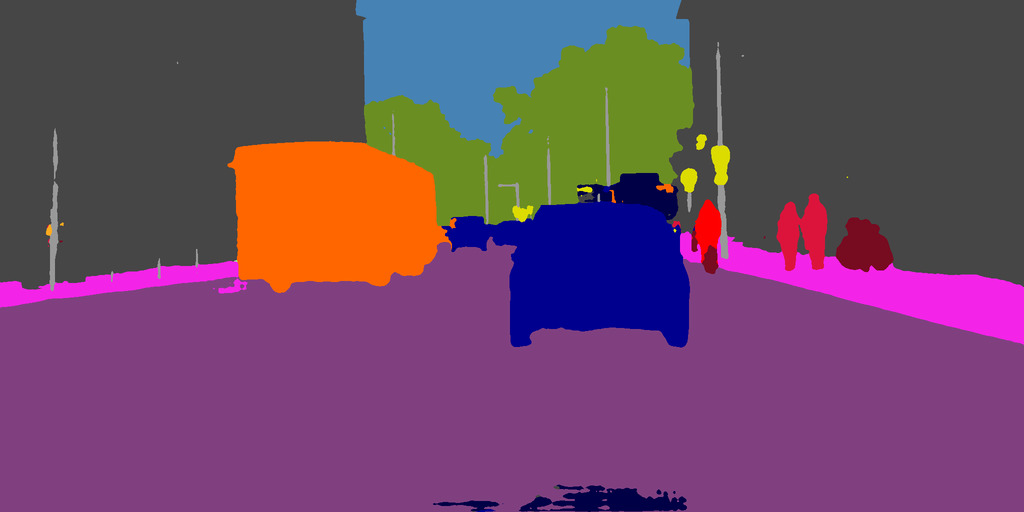}}
    \caption{Comparison of the semantic segmentation predictions of all DNNs evaluated in the second experiment for an example image from the Cityscapes validation data. 
    }
    \label{fig:results-2}
\end{figure*}

Again, we provide a class-wise evaluation on the Cityscapes validation split in \cref{tab:cs_bus} and present a comparison of different models for one exemplary street scene in \cref{fig:results-2}. Here, large parts of the bus in the foreground are predicted correctly by our extended DNN. The bus in the background is even better recognized by our network than by the baseline and oracle.
Analogous to the first experiment, the most similar classes \emph{truck} and \emph{train} show increasing IoU and precision, but decreasing recall values. Averaged over the known classes $c\in\mathcal{C}$, we again observe improvement in IoU and precision with a concurrent drop in recall. Averaged over the extended class set $\mathcal{C}^+$, all three performance measures increase after class-incremental learning.

\begin{table}[t]
    \centering
    \resizebox{0.47\textwidth}{!}{
    \begin{tabular}{l||ccc|ccc}
        \hline
        \rowcolor{maroon!10} \textbf{2.\ experiment} & \multicolumn{6}{c}{DeepLabV3+} \\\hhline{~||------}
        \rowcolor{maroon!10} Cityscapes, bus & \multicolumn{3}{c|}{initial} & \multicolumn{3}{c}{extended}\\\hline\hline
        Class  & IoU  & precision  & recall & IoU  & precision  & recall\\\hline\hline
        road & 97.63 & 98.81 & 98.80 & 97.57 $\pm$ 0.03 & 98.76 $\pm$ 0.09 & 98.79 $\pm$ 0.08 \\ \hline
        sidewalk & 81.60 & 89.65 & 90.09 & 81.57 $\pm$ 0.10 & 90.07 $\pm$ 0.46 & 89.63 $\pm$ 0.45 \\ \hline
        building  &  90.19 & 94.50 & 95.19 & 89.90 $\pm$ 0.10 & 94.22 $\pm$ 0.26 & 95.15 $\pm$ 0.25 \\ \hline
        wall & 48.77& 78.07 & 56.51 & 44.89 $\pm$ 3.11 & 79.23 $\pm$ 1.36 & 50.94 $\pm$ 4.20 \\ \hline
        fence  & 53.86 & 70.97 & 69.08 & 51.74 $\pm$ 0.81 & 71.82 $\pm$ 0.62 & 64.92 $\pm$ 1.27 \\ \hline
        pole  & 55.03 & 75.71 & 66.83 & 54.05 $\pm$ 0.61 & 77.62 $\pm$ 1.11 & 64.06 $\pm$ 1.54 \\ \hline
        traffic light  & 55.87 & 77.29 & 66.84 & 54.70 $\pm$ 0.92 & 80.15 $\pm$ 2.02 & 63.35 $\pm$ 2.46 \\ \hline
        traffic sign  & 68.21 & 87.02 & 75.94 & 67.88 $\pm$ 0.32 & 87.87 $\pm$ 0.98 & 74.91 $\pm$ 1.08 \\ \hline
        vegetation  & 90.35 & 93.98 & 95.91 & 90.21 $\pm$ 0.09 & 93.70 $\pm$ 0.33 & 96.04 $\pm$ 0.26 \\ \hline
        terrain & 54.03 & 79.90 & 62.53 & 52.77 $\pm$ 0.46 & 75.06 $\pm$ 1.14 & 64.00 $\pm$ 1.01 \\ \hline
        sky  & 93.64 & 96.14 & 97.30 & 93.26 $\pm$ 0.29 & 95.55 $\pm$ 0.63 & 97.49 $\pm$ 0.36 \\ \hline
        person  & 71.65 & 83.27 & 83.70 & 71.02 $\pm$ 0.21 & 82.22 $\pm$ 0.87 & 83.92 $\pm$ 0.65 \\ \hline
        rider  & 48.77 & 68.86 & 62.58 & 47.15 $\pm$ 0.73 & 70.85 $\pm$ 1.32 & 58.55 $\pm$ 1.99 \\ \hline
        car  & 91.90 & 94.65 & 96.94 & 91.76 $\pm$ 0.11 & 95.35 $\pm$ 0.61 & 96.07 $\pm$ 0.62 \\ \hline
        truck  & 47.51 & 51.19 & 86.87 & 54.14 $\pm$ 1.85 & 69.81 $\pm$ 4.17 & 71.09 $\pm$ 5.25 \\ \hline
        \rowcolor{Gray} bus  & 00.00 & 00.00 & 00.00 & 44.73 $\pm$ 1.46 & 58.33 $\pm$ 3.13 & 66.15 $\pm$ 5.16 \\ \hline
        train & 43.57 & 48.58 & 80.88 & 55.46 $\pm$ 1.64 & 74.35 $\pm$ 5.75 & 69.19 $\pm$ 5.46 \\ \hline
        motorcycle  & 44.35 & 61.76 & 61.13 & 41.66 $\pm$ 1.17 & 71.22 $\pm$ 1.70 & 50.16 $\pm$ 2.38 \\ \hline
        bicycle  & 68.00 & 77.42 & 84.82 & 67.52 $\pm$ 0.28 & 76.38 $\pm$ 0.64 & 85.35 $\pm$ 0.44 \\ \hline \hline
        mean over $\mathcal{C}$ & 66.94 & 79.32 & 79.55 & 67.07 $\pm$ 0.12 & 82.46 $\pm$ 0.56 & 76.31 $\pm$ 0.46 \\ \hline
        mean over ${\mathcal{C}^+}$ & 63.42 & 75.15 & 75.36 & 65.89 $\pm$ 0.10 & 81.19 $\pm$ 0.54 & 75.78 $\pm$ 0.34 \\ \hline
    \end{tabular}}
    \caption{In-depth evaluation on the Cityscapes validation data for the second experiment, where we incrementally extend a DeepLabV3+ by the novel class \emph{bus} on the Cityscapes dataset. We provide IoU, precision and recall values obtained for both, the initial and the extended DNN, on a class-level as well as averaged over the classes in $\mathcal{C}$ and $\mathcal{C}^+$, respectively.}
    \label{tab:cs_bus}
\end{table}

\begin{table}[t]
    \centering
    \resizebox{0.47\textwidth}{!}{
    \begin{tabular}{l||ccc|ccc}
        \hline
        \rowcolor{maroon!10} \textbf{3.\ experiment} & \multicolumn{6}{c}{DeepLabV3+} \\\hhline{~||------}
        \rowcolor{maroon!10} Cityscapes, multi & \multicolumn{3}{c|}{initial} & \multicolumn{3}{c}{extended}\\\hline\hline
        Class  & IoU  & precision  & recall & IoU  & precision  & recall\\\hline\hline
        road & 95.43 & 96.41 & 98.95 & 96.62 $\pm$ 0.07 & 98.29 $\pm$ 0.20 & 98.27 $\pm$ 0.22 \\ \hline
        sidewalk & 77.23 & 83.84 & 90.74 & 76.42 $\pm$ 0.26 & 84.27 $\pm$ 0.98 & 89.16 $\pm$ 0.91 \\ \hline
        building & 87.21 & 91.05 & 95.39 & 87.42 $\pm$ 0.12 & 92.66 $\pm$ 0.30 & 93.92 $\pm$ 0.40 \\ \hline
        wall & 45.86 & 68.38 & 58.20 & 40.36 $\pm$ 0.59 & 76.67 $\pm$ 1.57 & 46.03 $\pm$ 1.07 \\ \hline
        fence & 47.86 & 59.63 & 70.79 & 41.15 $\pm$ 1.47 & 69.23 $\pm$ 2.40 & 50.44 $\pm$ 2.54 \\ \hline
        pole & 51.63 & 69.15 & 67.09 & 48.68 $\pm$ 0.48 & 73.74 $\pm$ 1.13 & 58.93 $\pm$ 1.42 \\ \hline
        traffic light & 55.61 & 77.70 & 66.17 & 45.62 $\pm$ 0.47 & 72.64 $\pm$ 0.85 & 55.09 $\pm$ 1.07 \\ \hline
        traffic sign & 64.84 & 80.37 & 77.04 & 58.34 $\pm$ 0.74 & 86.84 $\pm$ 0.70 & 64.01 $\pm$ 1.23 \\ \hline
        vegetation & 88.26 & 91.27 & 96.40 & 88.61 $\pm$ 0.22 & 91.80 $\pm$ 0.43 & 96.22 $\pm$ 0.21 \\ \hline
        terrain & 53.22 & 72.42 & 66.74 & 45.43 $\pm$ 0.77 & 79.11 $\pm$ 1.55 & 51.66 $\pm$ 1.67 \\ \hline
        sky & 93.58 & 96.11 & 97.27 & 92.41 $\pm$ 0.16 & 95.56 $\pm$ 0.19 & 96.56 $\pm$ 0.10 \\ \hline
        \rowcolor{Gray} human & 00.00 & 00.00 & 00.00 & 40.22 $\pm$ 1.77 & 68.74 $\pm$ 4.84 & 49.65 $\pm$ 4.80 \\ \hline
        \rowcolor{Gray} car & 00.00 & 00.00 & 00.00 & 81.27 $\pm$ 1.16 & 86.56 $\pm$ 2.20 & 93.05 $\pm$ 1.12 \\ \hline
        truck & 9.31 & 9.41 & 89.35 & 25.59 $\pm$ 7.41 & 61.27 $\pm$ 5.50 & 30.77 $\pm$ 9.90 \\ \hline
        train & 41.70 & 45.05 & 84.87 & 49.87 $\pm$ 5.21 & 60.85 $\pm$ 8.56 & 73.99 $\pm$ 2.61 \\ \hline
        motorcycle & 4.03 & 4.12 & 66.09 & 14.30 $\pm$ 2.72 & 63.79 $\pm$ 3.44 & 15.64 $\pm$ 3.31 \\ \hline
        bicycle & 39.13 & 41.30 & 88.15 & 51.97 $\pm$ 1.58 & 71.26 $\pm$ 1.98 & 65.95 $\pm$ 4.30 \\ \hline\hline
        mean over $\mathcal{C}$ & 56.99 & 65.75 & 80.88 & 57.52 $\pm$ 0.80 & 78.53 $\pm$ 1.20 & 65.78 $\pm$ 1.00 \\ \hline
        mean over ${\mathcal{C}^+}$ & 50.29 & 58.01 & 71.37 & 57.90 $\pm$ 0.68 & 78.43 $\pm$ 1.10 & 66.43 $\pm$ 0.94 \\ \hline
    \end{tabular}}
    \caption{In-depth evaluation on the Cityscapes validation data for the third experiment, where we incrementally extend a DeepLabV3+ by the novel classes \emph{human} and \emph{car} on the Cityscapes dataset. We provide IoU, precision and recall values obtained for both, the initial and the extended DNN, on a class-level as well as averaged over the classes in $\mathcal{C}$ and $\mathcal{C}^+$, respectively.}
    \label{tab:cs_human_and_car}
\end{table}

\begin{table}[t]
    \centering
    \resizebox{0.47\textwidth}{!}{
    \begin{tabular}{l||ccc|ccc}
        \hline
        \rowcolor{maroon!10} \textbf{4.\ experiment (a)} & \multicolumn{6}{c}{DeepLabV3+} \\\hhline{~||------}
        \rowcolor{maroon!10} A2D2, guardrail & \multicolumn{3}{c|}{initial} & \multicolumn{3}{c}{extended}\\\hline\hline
        Class  & IoU  & precision  & recall & IoU  & precision  & recall\\\hline\hline
        road  & 95.59 & 97.21 & 98.29 & 95.93 $\pm$ 0.06 & 97.94 $\pm$ 0.18 & 97.91 $\pm$ 0.15 \\ \hline
        sidewalk & 72.01 & 86.73 & 80.92 & 72.08 $\pm$ 0.41 & 85.29 $\pm$ 0.84 & 82.33 $\pm$ 1.28 \\ \hline
        building & 87.82 & 93.58 & 93.44 & 85.75 $\pm$ 0.67 & 93.13 $\pm$ 0.53 & 91.54 $\pm$ 1.01 \\ \hline
        fence & 59.35 & 81.59 & 68.53 & 56.76 $\pm$ 0.37 & 79.89 $\pm$ 2.40 & 66.29 $\pm$ 1.63 \\ \hline
        pole  & 56.13 & 76.39 & 67.91 & 54.31 $\pm$ 0.24 & 77.86 $\pm$ 0.52 & 64.23 $\pm$ 0.66 \\ \hline
        traffic light & 68.41 & 85.10 & 77.72 & 65.48 $\pm$ 0.19 & 84.21 $\pm$ 0.77 & 74.65 $\pm$ 0.83 \\ \hline
        traffic sign & 76.34 & 86.78 & 86.38 & 74.53 $\pm$ 0.38 & 89.98 $\pm$ 1.11 & 81.30 $\pm$ 1.19 \\ \hline
        vegetation & 91.61 & 94.01 & 97.29 & 92.00 $\pm$ 0.23 & 94.81 $\pm$ 0.38 & 96.89 $\pm$ 0.17 \\ \hline
        sky & 97.96 & 98.72 & 99.22 & 97.81 $\pm$ 0.03 & 98.57 $\pm$ 0.07 & 99.22 $\pm$ 0.04 \\ \hline
        person & 67.60 & 79.28 & 82.11 & 64.27 $\pm$ 0.58 & 87.70 $\pm$ 0.87 & 70.65 $\pm$ 1.21 \\ \hline
        car & 93.19 & 96.73 & 96.22 & 92.42 $\pm$ 0.11 & 96.04 $\pm$ 0.35 & 96.08 $\pm$ 0.35 \\ \hline
        truck & 84.99 & 88.51 & 95.53 & 80.98 $\pm$ 2.66 & 84.75 $\pm$ 3.29 & 94.82 $\pm$ 0.69 \\ \hline
        motorcycle & 48.68 & 84.71 & 53.37 & 26.05 $\pm$ 2.72 & 90.18 $\pm$ 2.09 & 26.85 $\pm$ 3.04 \\ \hline
        bicycle & 61.08 & 80.65 & 71.57 & 50.65 $\pm$ 3.27 & 85.78 $\pm$ 2.10 & 55.43 $\pm$ 4.78 \\ \hline
        \rowcolor{Gray} guardrail & 00.00 & 00.00 & 00.00 & 46.10 $\pm$ 4.79 & 80.41 $\pm$ 2.12 & 52.09 $\pm$ 6.42 \\ \hline  \hline
        mean over $\mathcal{C}$ & 75.77 & 87.86 & 83.47 & 72.07 $\pm$ 0.39 & 89.01 $\pm$ 0.48 & 78.44 $\pm$ 0.52 \\ \hline
        mean over ${\mathcal{C}^+}$ & 70.72 & 82.00 & 77.90 & 70.34 $\pm$ 0.50 & 88.44 $\pm$ 0.40 & 76.69 $\pm$ 0.47 \\ \hline
    \end{tabular}}
    \caption{In-depth evaluation on the A2D2 validation data for the fourth experiment, where we first fine-tune and then incrementally extend a DeepLabV3+ by the novel class \emph{guardrail} on the A2D2 dataset. We provide IoU, precision and recall values obtained for both, the initial and the extended DNN, on a class-level as well as averaged over the classes in $\mathcal{C}$ and $\mathcal{C}^+$, respectively.}
    \label{tab:a2d2_guardrail-dl-fine-tuned}
\end{table}

\subsection{Experiment 3}

In the next experiment we extend the previous ones by enlarging the set of novel classes, withholding the classes \emph{pedestrian}\&\emph{rider}, \emph{bus} and \emph{car}. Again, we trained a DeepLabV3+ network on the Cityscapes dataset to learn the remaining, non-novel classes.
We reconsidered our approach to reject possibly known objects from the embedding space to improve the purity of novel object clusters. Instead of rejecting anomalous segments that consist of only one predicted segment in the semantic segmentation mask, we include a random choice of objects / segments from each known class into the embedding space. If an anomalous object can be assigned to an existing class, it is no longer taken into account in the further procedure. To decide whether an object is novel or known, we consider its 2.75-neighborhood. If this contains at least 10 known objects from which at least 80\% belong to the most frequent class, we assume the anomaly belongs to even this class, \ie we reject it. Consequently, we discard the detected bus segments since these are closely related to the classes \emph{truck} and \emph{train}. However, we obtain two clusters, one for the class \emph{car} (1375 segments) and one for the class \emph{human} (135 segments). We incrementally expand the model by these classes, achieving a similar IoU value (around 40\%) for the \emph{human} class as in experiment 1, where we only learned a single class. For the \emph{bus} class, we even get an IoU value of more than 80\%. Detailed results are provided in \cref{tab:cs_human_and_car}.

\begin{figure*}[t]
    \captionsetup[subfigure]{labelformat=empty}
    \centering
    \subfloat[image \& annotation]{\includegraphics[width=0.19\textwidth]{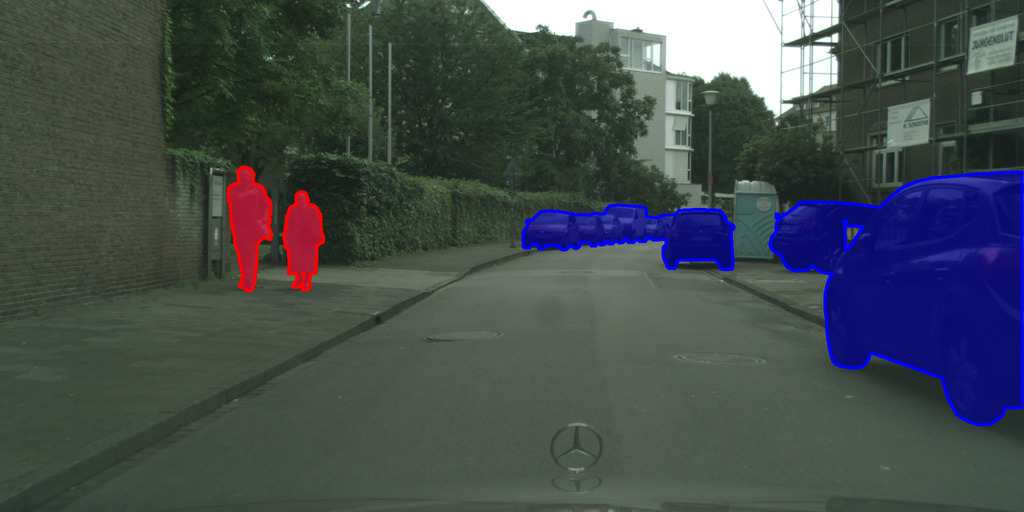}}~
    \subfloat[initial DNN]{\includegraphics[width=0.19\textwidth]{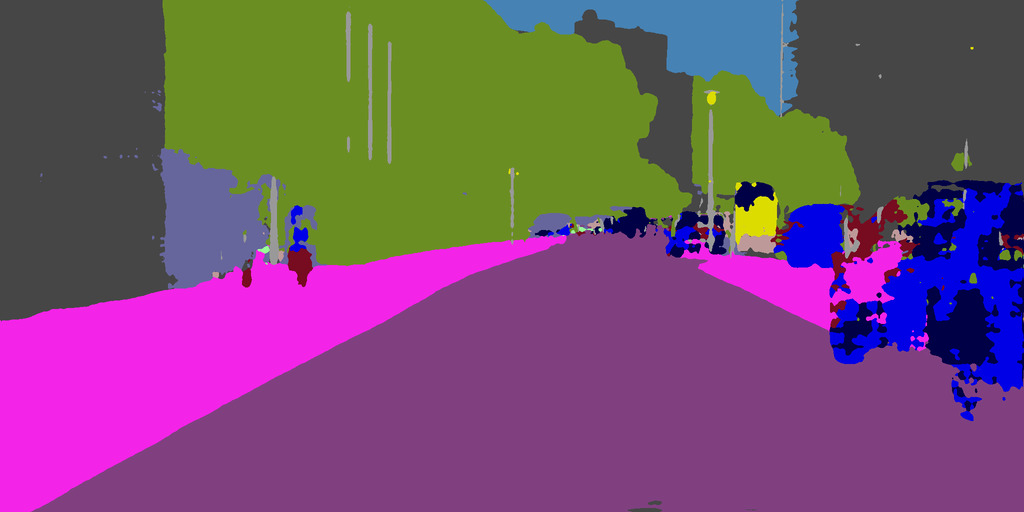}}~
    \subfloat[extended DNN]{\includegraphics[width=0.19\textwidth]{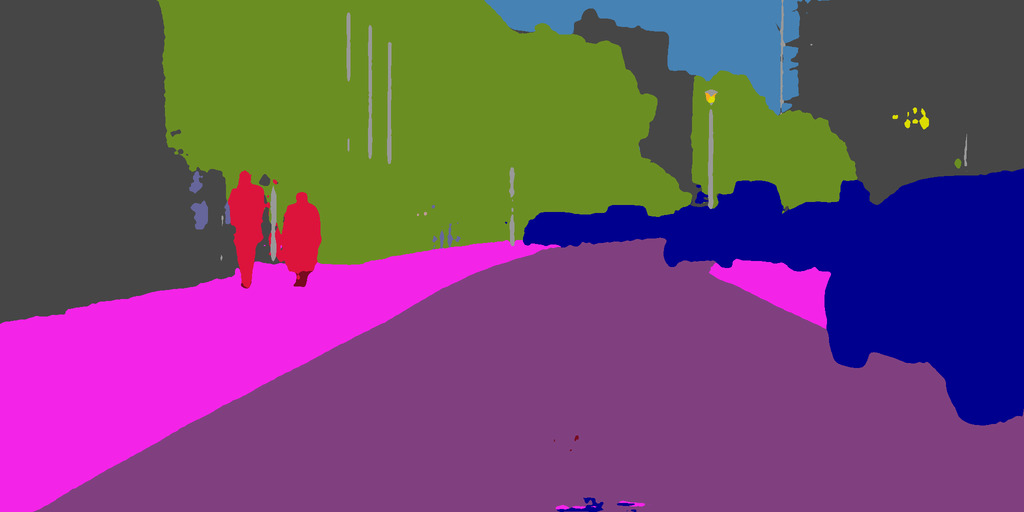}}~
    \subfloat[oracle]{\includegraphics[width=0.19\textwidth]{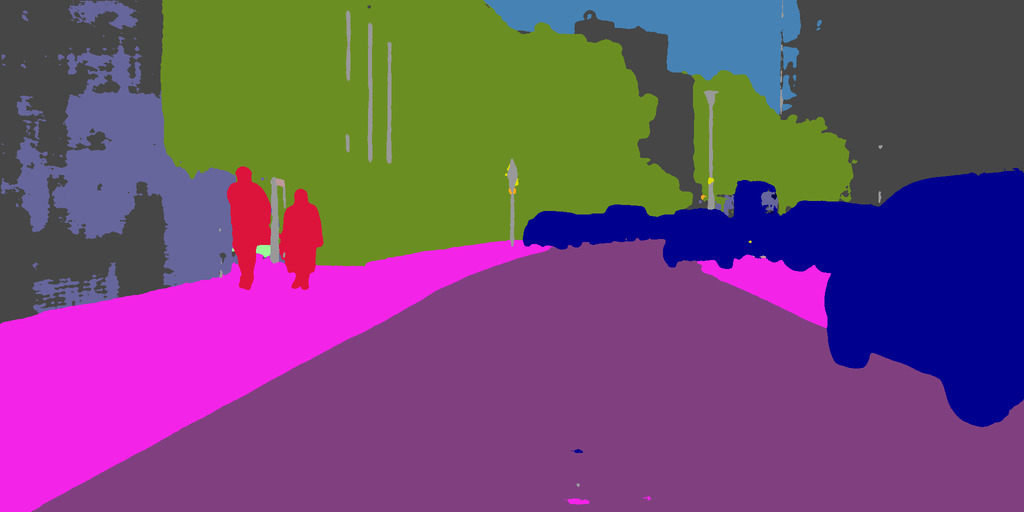}}
    \caption{Comparison of the semantic segmentation predictions of all DNNs evaluated in the third experiment for an example image from the Cityscapes validation data. 
    }
    \label{fig:results-3}
\end{figure*}

\subsection{Experiment 4(a)}

The fourth experiment involves two different network architectures. 
Results for the first one are shown in experiment 4(a), results for the other one in 4(b).
We start with a DeepLabV3+ network trained on the Cityscapes dataset and aim to detect and learn the \emph{guardrail} class using images taken from the A2D2 dataset. To mitigate a performance drop caused by the domain shift from Cityscapes to A2D2, we first fine-tune the decoder for 70 epochs on our A2D2 training split, applying the same hyperparameters we used for the incremental training (see \cref{sec:experiments}). By that, we improve the mean IoU of the initial network from 59.38\% to 75.77\%. The classes which suffer the most are \emph{person}, \emph{motorcycle} and \emph{bicycle}, which is presumably due to their rare occurrence on country roads and highways, and therefore, low frequency in the re-training data, which involves only 30 pseudo-labeled and 30 replayed images. Further details are provided in \cref{tab:a2d2_guardrail-dl-fine-tuned}.

\begin{figure*}[t]
    \captionsetup[subfigure]{labelformat=empty, position=top}
    \centering
    \subfloat[image \& novelty annotation]{\includegraphics[trim={0 0 0 124px},clip,width=0.23\textwidth]{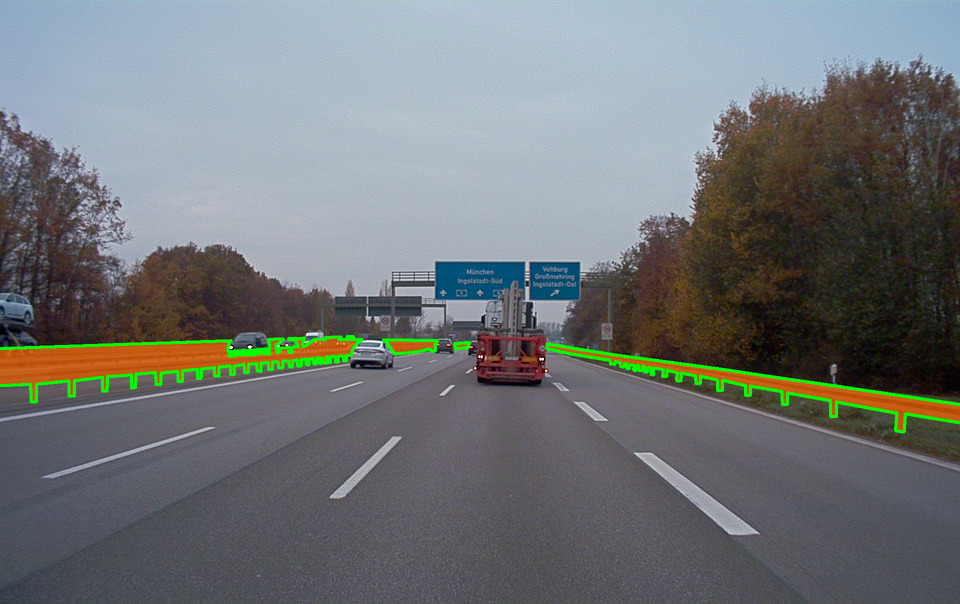}}~~
    \subfloat[]{\rotatebox[origin=lb]{90}{\scriptsize 4.\ experiment (a) ~}}~
    \subfloat[initial DNN]{\includegraphics[trim={0 0 0 124px},clip,width=0.23\textwidth]{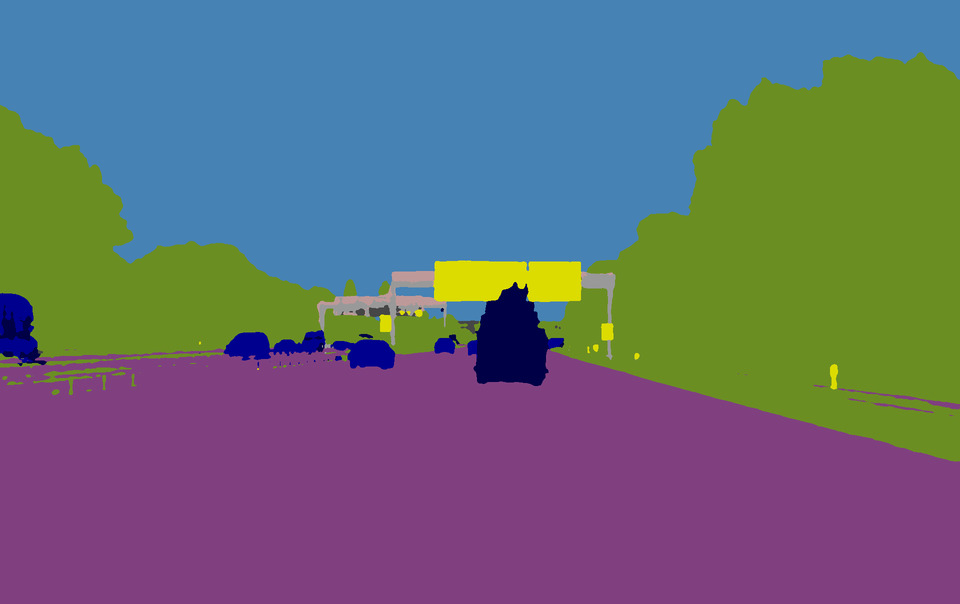}}~
    \subfloat[extended DNN]{\includegraphics[trim={0 0 0 124px},clip,width=0.23\textwidth]{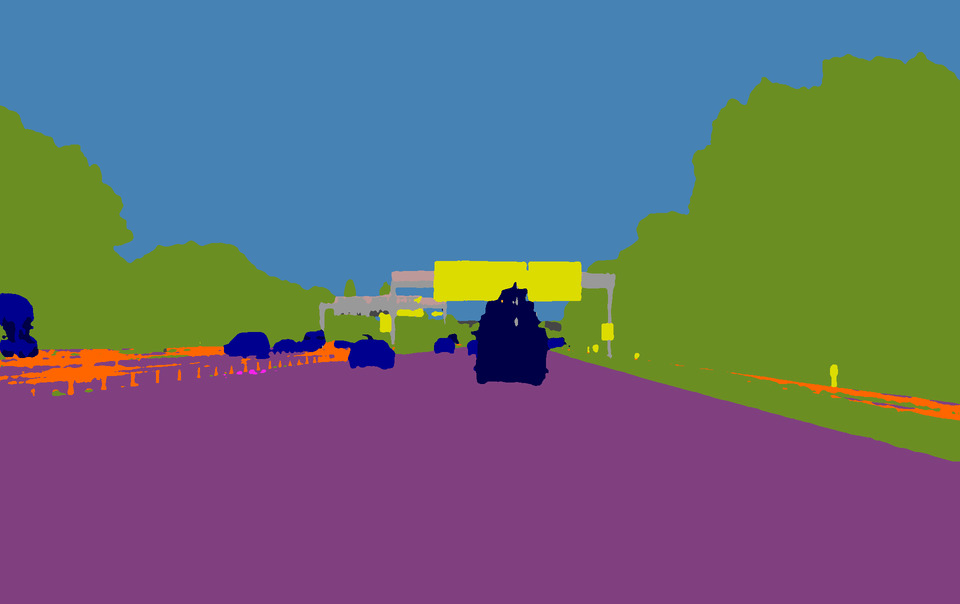}}~
    \subfloat[oracle]{\includegraphics[trim={0 0 0 124px},clip,width=0.23\textwidth]{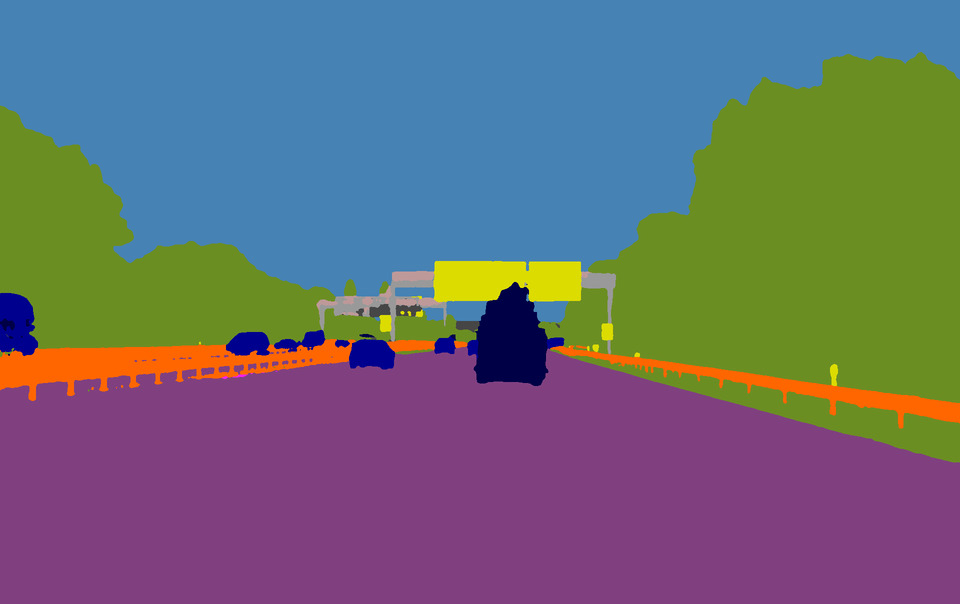}}
    \\
    \vspace{-0.3cm}
    \captionsetup[subfigure]{labelformat=empty, position=bottom}
    \subfloat[ground truth]{\includegraphics[trim={0 0 0 124px},clip,width=0.23\textwidth]{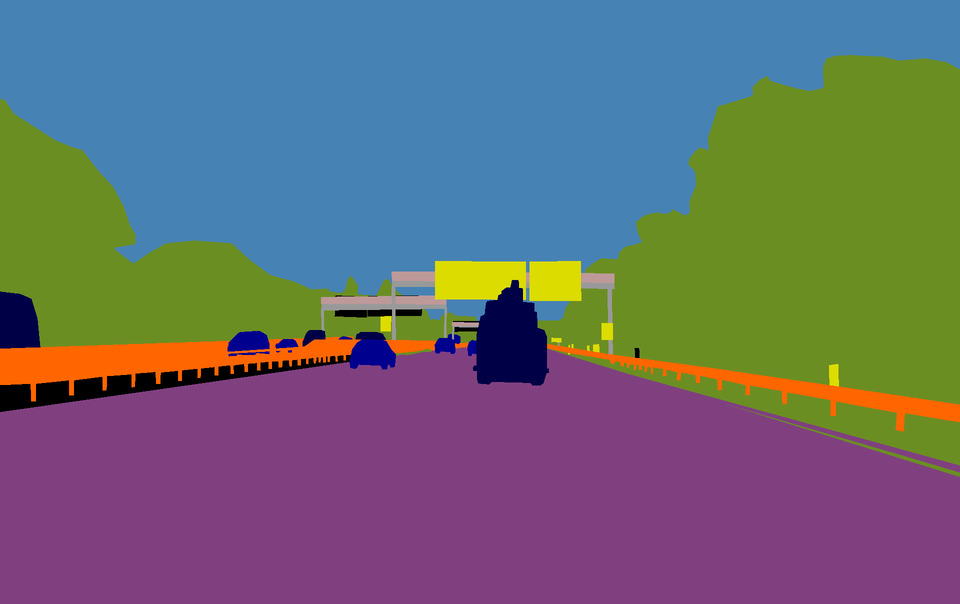}}~~
    \subfloat[]{\rotatebox[origin=lb]{90}{\scriptsize ~ 4.\ experiment (b) }}~
    \subfloat[]{\includegraphics[trim={0 0 0 124px},clip,width=0.23\textwidth]{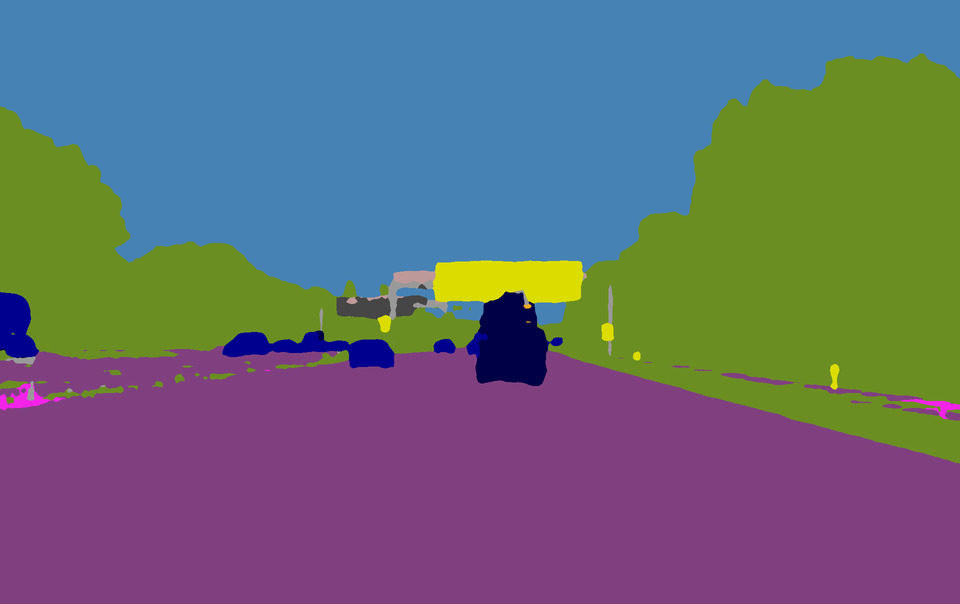}}~
    \subfloat[]{\includegraphics[trim={0 0 0 124px},clip,width=0.23\textwidth]{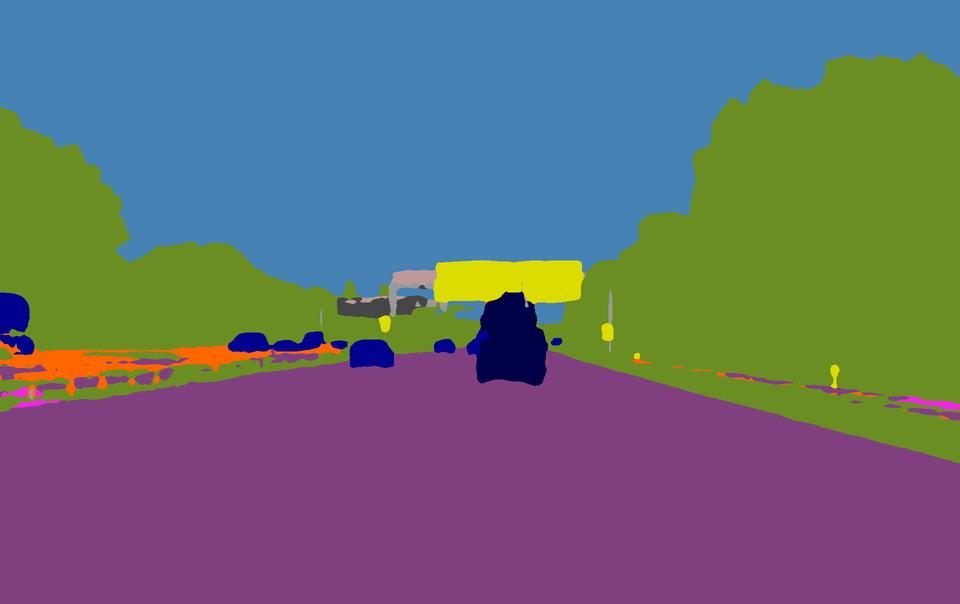}}~
    \subfloat[]{\includegraphics[trim={0 0 0 124px},clip,width=0.23\textwidth]{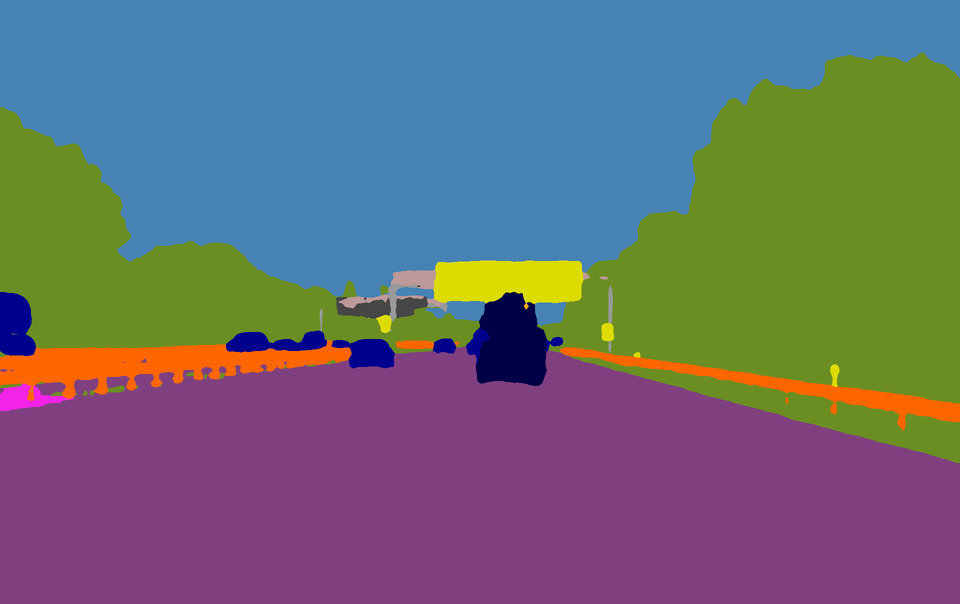}}\\
    \vspace{-0.9cm}
    \subfloat[]{\includegraphics[trim={0 0 0 124px},clip,width=0.23\textwidth]{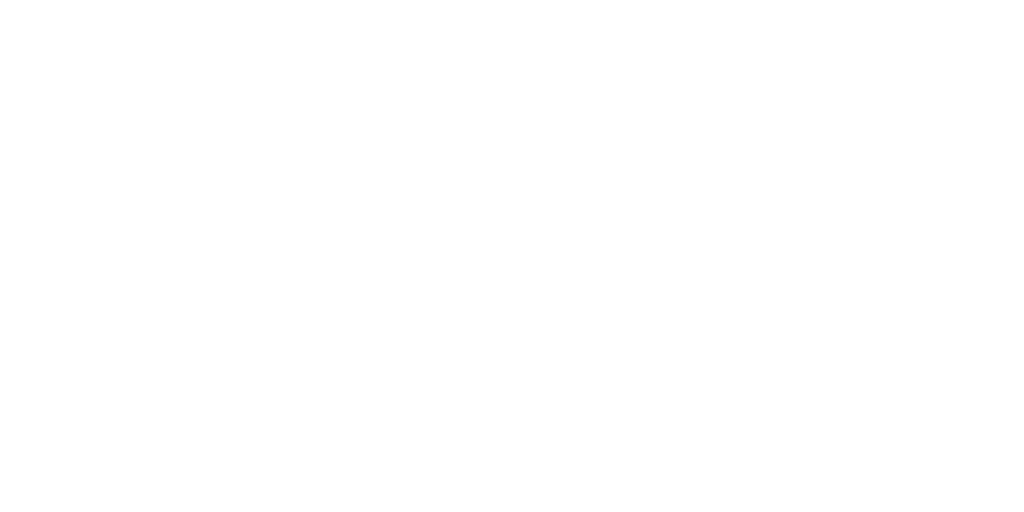}}~~
    \subfloat[]{\rotatebox[origin=lb]{90}{\scriptsize ~~ 5.\ experiment }}~
    \subfloat[]{\includegraphics[trim={0 0 0 124px},clip,width=0.23\textwidth]{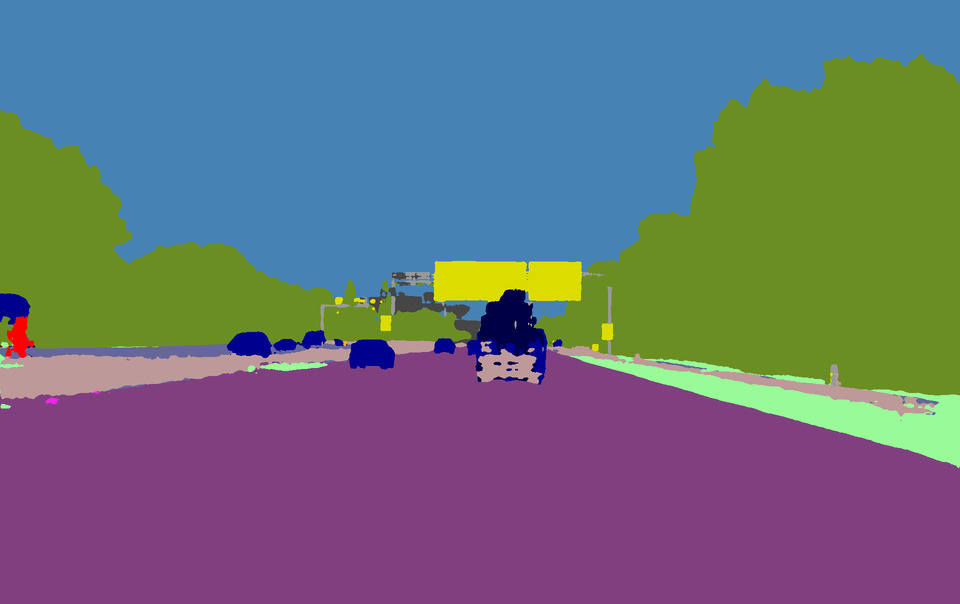}}~
    \subfloat[]{\includegraphics[trim={0 0 0 124px},clip,width=0.23\textwidth]{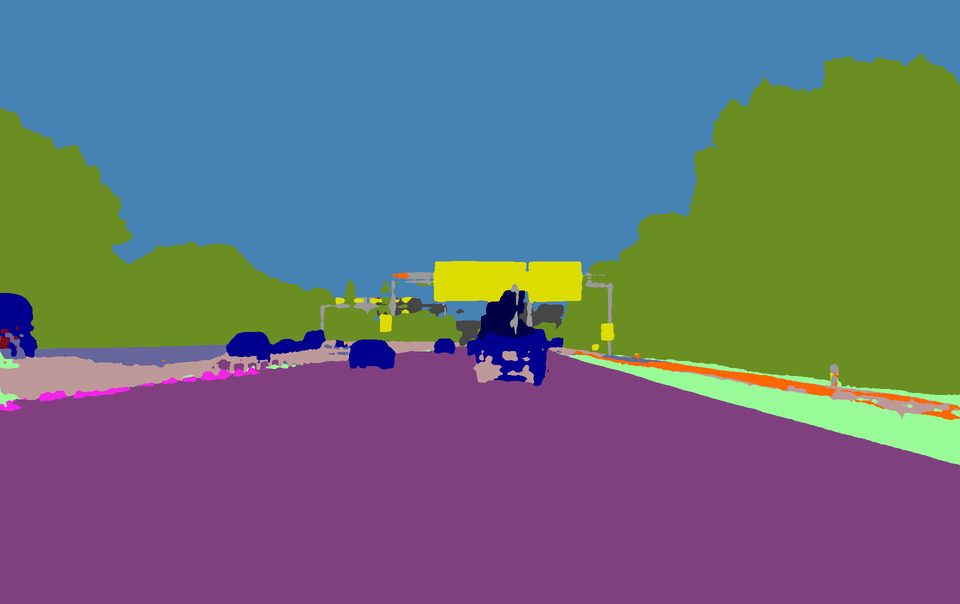}}~
    \subfloat[]{\includegraphics[width=0.23\textwidth]{figures/empty.png}}
    \\
    \caption{Comparison of the semantic segmentation predictions of all models incrementally extended by the \emph{guardrail} class for an example image from the A2D2 validation split. 
    }
    \label{fig:results-4and5}
\end{figure*}

\begin{figure}[t]
    \captionsetup[subfigure]{labelformat=empty}
    \centering
    \subfloat[]{\includegraphics[trim={0 0 0 124px},clip,width=0.23\textwidth]{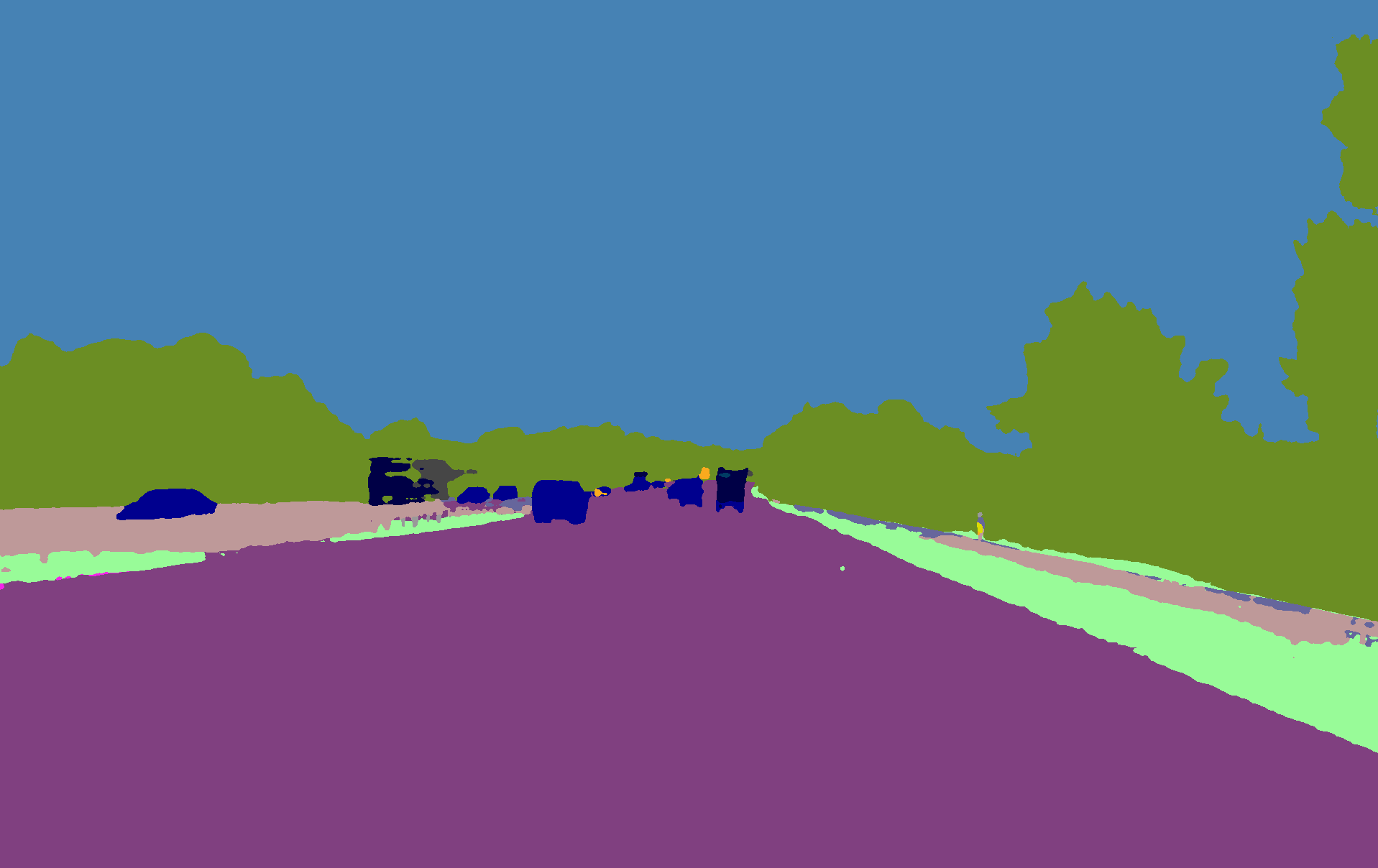}}~
    \subfloat[]{\includegraphics[trim={0 0 0 165px},clip,width=0.23\textwidth]{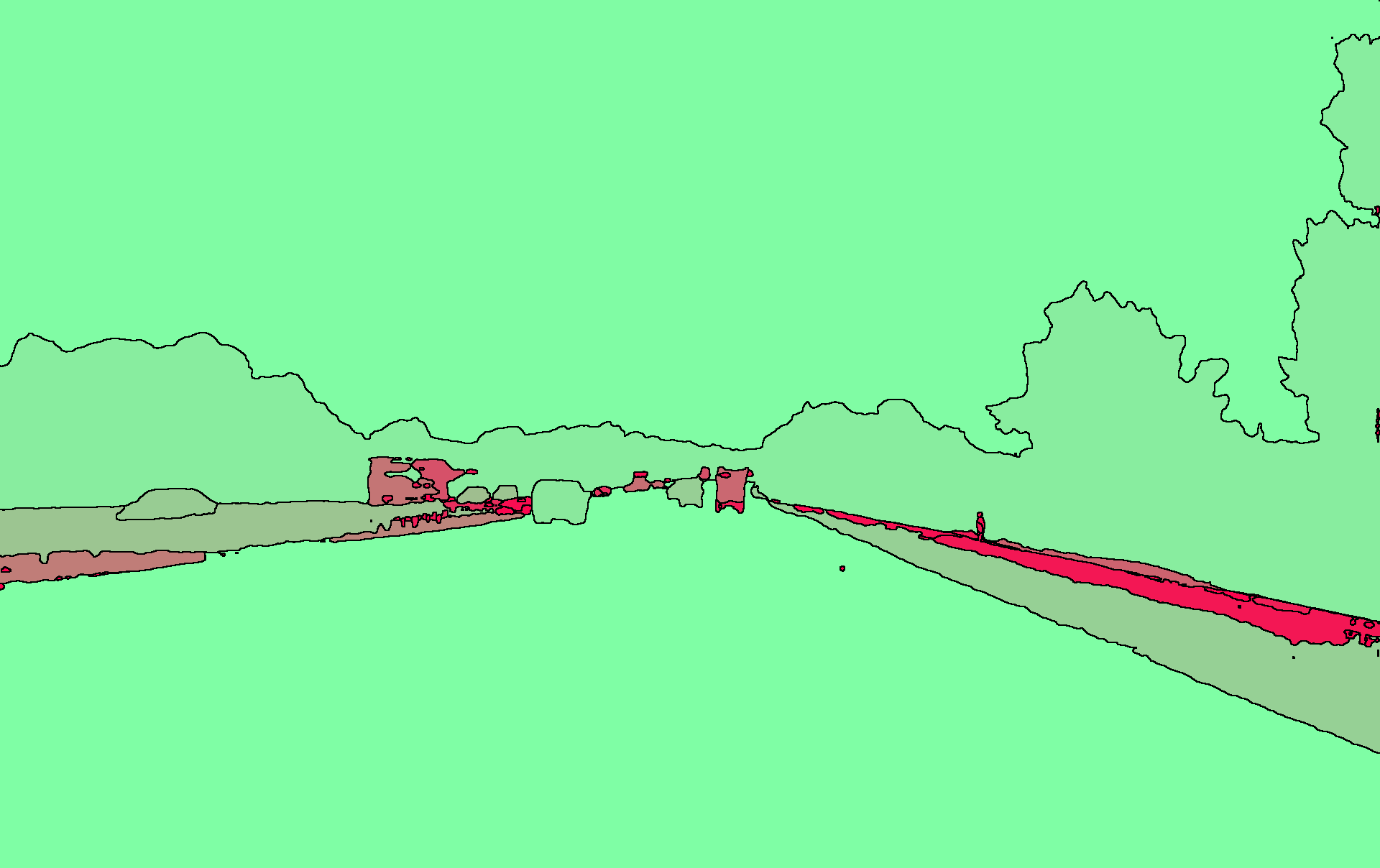}}\\
    \vspace{-0.75cm}
    \subfloat[predicted segmentation]{\includegraphics[trim={0 0 0 124px},clip,width=0.23\textwidth]{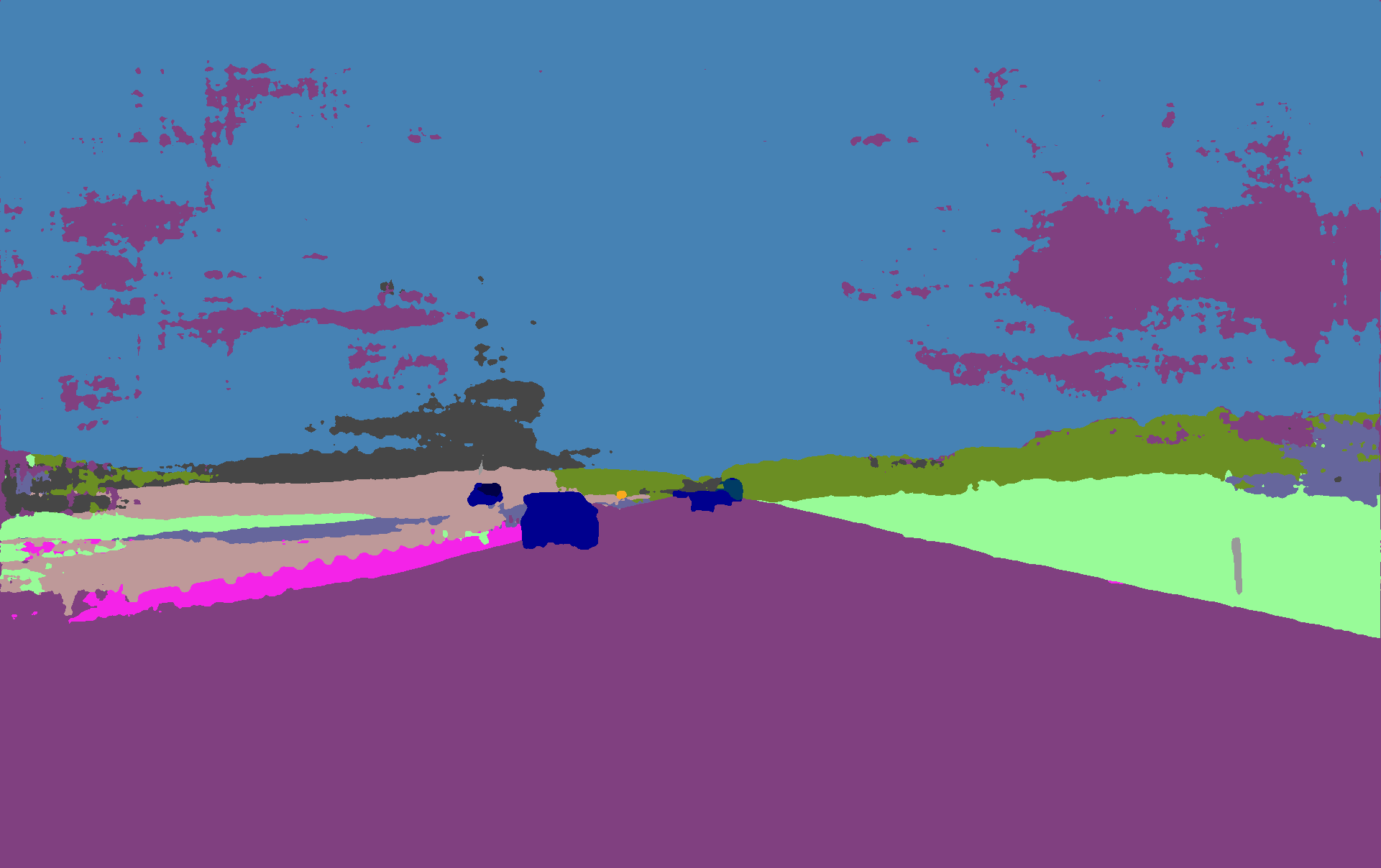}}~
    \subfloat[quality estimation]{\includegraphics[trim={0 0 0 166px},clip,width=0.23\textwidth]{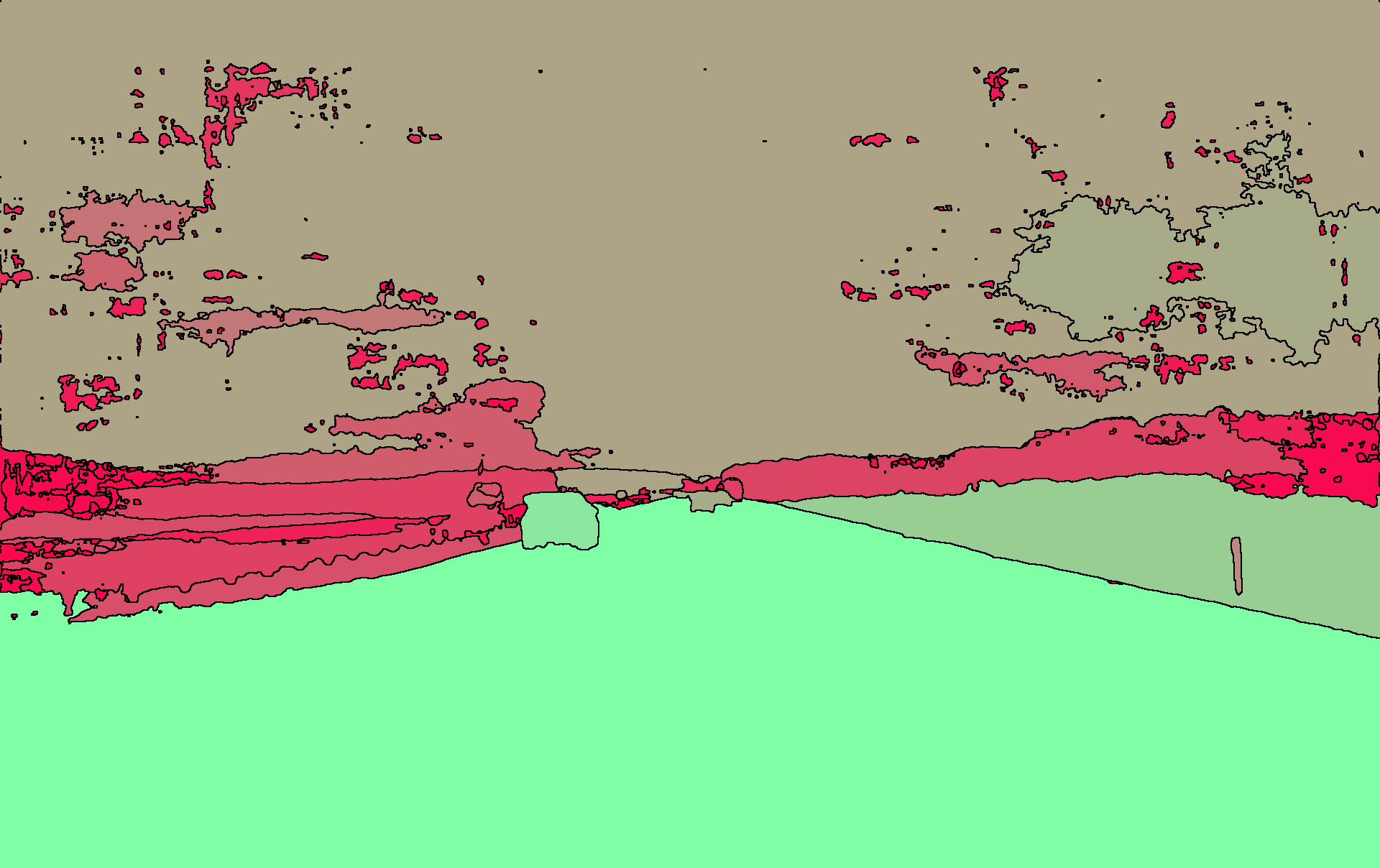}}
    \caption{Illustration of prediction quality differences (green color indicates high, red color low prediction quality), caused by the domain shift from Cityscapes to A2D2, mainly due to weather conditions.}
    \label{fig:domain-shift}
\end{figure}

\subsection{Experiment 4(b)}

\begin{table}[t]
    \centering
    \resizebox{0.47\textwidth}{!}{
    \begin{tabular}{l||ccc|ccc}
        \hline
        \rowcolor{maroon!10} \textbf{4.\ experiment (b)} & \multicolumn{6}{c}{PSPNet} \\\hhline{~||------}
        \rowcolor{maroon!10} A2D2, guardrail & \multicolumn{3}{c|}{initial} & \multicolumn{3}{c}{extended}\\\hline\hline
        Class  & IoU  & precision  & recall & IoU  & precision  & recall\\\hline\hline
        road  & 95.18 & 97.10 & 97.96 & 94.93 $\pm$ 0.21 & 96.94 $\pm$ 0.55 & 97.86 $\pm$ 0.34 \\ \hline
        sidewalk & 66.15 & 83.68 & 75.94 & 62.19 $\pm$ 2.28 & 82.28 $\pm$ 2.09 & 71.99 $\pm$ 4.75 \\ \hline
        building & 84.32 & 92.46 & 90.54 & 82.38 $\pm$ 0.46 & 90.78 $\pm$ 0.86 & 89.91 $\pm$ 1.04 \\ \hline
        fence & 54.48 & 76.84 & 65.18 & 50.67 $\pm$ 1.24 & 80.91 $\pm$ 1.85 & 57.62 $\pm$ 2.33 \\ \hline
        pole  & 44.60 & 63.94 & 59.59 & 42.15 $\pm$ 0.91 & 65.52 $\pm$ 2.19 & 54.31 $\pm$ 2.89 \\ \hline
        traffic light & 58.94 & 81.14 & 68.30 & 56.07 $\pm$ 0.17 & 80.65 $\pm$ 1.85 & 64.83 $\pm$ 1.37 \\ \hline
        traffic sign & 71.30 & 87.71 & 79.22 & 67.63 $\pm$ 0.47 & 87.61 $\pm$ 0.71 & 74.79 $\pm$ 0.56 \\ \hline
        vegetation & 90.68 & 93.12 & 97.18 & 90.65 $\pm$ 0.11 & 93.71 $\pm$ 0.41 & 96.53 $\pm$ 0.32 \\ \hline
        sky & 97.57 & 98.44 & 99.10 & 97.21 $\pm$ 0.12 & 98.06 $\pm$ 0.19 & 99.12 $\pm$ 0.10 \\ \hline
        person & 59.17 & 82.53 & 67.64 & 46.20 $\pm$ 1.13 & 82.99 $\pm$ 0.99 & 51.04 $\pm$ 1.60 \\ \hline
        car & 89.39 & 94.36 & 94.44 & 86.82 $\pm$ 0.34 & 93.90 $\pm$ 0.57 & 92.01 $\pm$ 0.60 \\ \hline
        truck & 77.83 & 84.05 & 91.31 & 73.53 $\pm$ 1.91 & 82.11 $\pm$ 2.40 & 87.58 $\pm$ 1.25 \\ \hline
        motorcycle & 19.73 & 76.72 & 20.99 & 7.00 $\pm$ 2.02 & 94.92 $\pm$ 3.73 & 7.04 $\pm$ 2.07 \\ \hline
        bicycle & 53.49 & 71.82 & 67.70 & 46.05 $\pm$ 1.37 & 79.31 $\pm$ 2.49 & 52.44 $\pm$ 2.71 \\ \hline
        \rowcolor{Gray} guardrail & 00.00 & 00.00 & 00.00 & 32.79 $\pm$ 3.47 & 70.75 $\pm$ 2.04 & 38.04 $\pm$ 4.90 \\ \hline\hline
        mean over $\mathcal{C}$ & 68.77 & 84.57 & 76.79 & 64.54 $\pm$ 0.28 & 86.41 $\pm$ 0.77 & 71.22 $\pm$ 0.69 \\ \hline
        mean over ${\mathcal{C}^+}$ & 64.19 & 78.93 & 71.67 & 62.42 $\pm$ 0.42 & 85.36 $\pm$ 0.78 & 69.01 $\pm$ 0.94 \\ \hline
    \end{tabular}}
    \caption{In-depth evaluation on the A2D2 validation data for the fourth experiment, where we first fine-tune and then incrementally extend a PSPNet by the novel class \emph{guardrail} on the A2D2 dataset. We provide IoU, precision and recall values obtained for both, the initial and the extended DNN, on a class-level as well as averaged over the classes in $\mathcal{C}$ and $\mathcal{C}^+$, respectively.}
    \label{tab:a2d2_guardrail-psp-fine-tuned}
\end{table}

In experiment 4(b), we employ a PSPNet instead of a DeepLabV3+, for the rest we proceed as in the previous subsection. Again, the training data consists of 30 images with pseudo ground truth and 30 labeled, replayed images (containing only old classes) from the A2D2 training split. Note that these 30 images are not the same as in experiment 4(a) due to the different network providing predictions of estimated low quality on different images.
In total, the initial and the extended PSPNet are outperformed by DeepLabV3+, however, both architectures show similar patterns:
\begin{itemize}
    \setlength\itemsep{0mm} 
    \item extended DNN exhibits a high precision$_\mathrm{guardrail}$ and a low recall$_\mathrm{guardrail}$
    \item classes that are mostly affected by re-training: \emph{person}, \emph{motorcycle}, \emph{bicycle}
    \item averaged over $\mathcal{C}$ and $\mathcal{C}^+$, respectively, IoU and recall values decrease, precision values increase
\end{itemize}
For more detailed information we refer to \cref{tab:a2d2_guardrail-psp-fine-tuned}.

\begin{table}[t]
    \centering
    \resizebox{0.47\textwidth}{!}{
    \begin{tabular}{l||ccc|ccc}
        \hline
        \rowcolor{maroon!10} \textbf{5.\ experiment} & \multicolumn{6}{c}{DeepLabV3+} \\\hhline{~||------}
        \rowcolor{maroon!10} A2D2, guardrail & \multicolumn{3}{c|}{initial} & \multicolumn{3}{c}{extended}\\\hline\hline
        Class  & IoU  & precision  & recall & IoU  & precision  & recall\\\hline\hline
        road  & 89.88 & 92.18 & 97.30 & 93.15 $\pm$ 0.19 & 94.89 $\pm$ 0.23 & 98.07 $\pm$ 0.12 \\ \hline
        sidewalk & 47.91 & 76.22 & 56.33 & 35.28 $\pm$ 2.43 & 86.95 $\pm$ 0.98 & 37.26 $\pm$ 2.67 \\ \hline
        building & 70.94 & 86.88 & 79.45 & 71.25 $\pm$ 1.46 & 90.51 $\pm$ 0.89 & 77.03 $\pm$ 2.21 \\ \hline
        fence & 26.08 & 35.30 & 49.94 & 26.20 $\pm$ 0.49 & 37.25 $\pm$ 1.46 & 46.99 $\pm$ 1.26 \\ \hline
        pole  & 42.59 & 59.24 & 60.25 & 42.77 $\pm$ 0.37 & 62.91 $\pm$ 0.73 & 57.21 $\pm$ 0.85 \\ \hline
        traffic light & 47.59 & 85.85 & 51.64 & 52.52 $\pm$ 0.70 & 89.21 $\pm$ 1.15 & 56.10 $\pm$ 1.19 \\ \hline
        traffic sign & 54.89 & 82.49 & 62.13 & 57.23 $\pm$ 0.25 & 87.34 $\pm$ 1.03 & 62.42 $\pm$ 0.43 \\ \hline
        vegetation & 69.15 & 96.68 & 70.83 & 73.42 $\pm$ 0.41 & 95.05 $\pm$ 0.62 & 76.35 $\pm$ 0.34 \\ \hline
        sky & 94.96 & 98.25 & 96.59 & 96.92 $\pm$ 0.09 & 97.81 $\pm$ 0.13 & 99.08 $\pm$ 0.05 \\ \hline
        person & 59.77 & 71.00 & 79.08 & 59.58 $\pm$ 1.23 & 84.68 $\pm$ 2.45 & 66.88 $\pm$ 2.89 \\ \hline
        car & 90.47 & 95.72 & 94.28 & 90.72 $\pm$ 0.16 & 96.14 $\pm$ 0.39 & 94.16 $\pm$ 0.53 \\ \hline
        truck & 62.64 & 83.61 & 71.40 & 71.10 $\pm$ 0.24 & 89.44 $\pm$ 0.51 & 77.62 $\pm$ 0.36 \\ \hline
        motorcycle & 28.39 & 70.82 & 32.15 & 32.77 $\pm$ 3.05 & 79.50 $\pm$ 3.43 & 35.96 $\pm$ 4.24 \\ \hline
        bicycle & 46.04 & 78.74 & 52.57 & 43.84 $\pm$ 1.01 & 85.43 $\pm$ 1.50 & 47.41 $\pm$ 1.56 \\ \hline
        \rowcolor{Gray} guardrail & 00.00 & 00.00 & 00.00 & 20.90 $\pm$ 1.73 & 77.12 $\pm$ 3.95 & 22.32 $\pm$ 2.07 \\ \hline\hline
        mean over $\mathcal{C}$ & 59.38 & 79.50 & 68.14 & 60.48 $\pm$ 0.47 & 84.08 $\pm$ 0.49 & 66.61 $\pm$ 0.64 \\ \hline
        mean over ${\mathcal{C}^+}$ & 55.42 & 74.20 & 63.60 & 57.84 $\pm$ 0.48 & 83.61 $\pm$ 0.68 & 63.66 $\pm$ 0.63 \\ \hline
    \end{tabular}}
    \caption{In-depth evaluation on the A2D2 validation data for the fifth experiment, where we incrementally extend a DeepLabV3+ (trained on Cityscapes) by the novel class \emph{guardrail} on the A2D2 dataset. We provide IoU, precision and recall values obtained for both, the initial and the extended DNN, on a class-level as well as averaged over the classes in $\mathcal{C}$ and $\mathcal{C}^+$, respectively.}
    \label{tab:a2d2_guardrail-domain-shift}
\end{table}

\subsection{Experiment 5}

Finally, we perform the same experiment as in 4(a) without prior fine-tuning the initial DNN on A2D2. Consequently, the domain shift causes many noisy predictions, exhibiting low prediction quality estimates. We exclude such images from the further process based on two criteria:
\begin{enumerate}
    \setlength\itemsep{0mm} 
    \item mean quality score (averaged over pixels) less than 0.7
    \item more than 1/3 of all pixels with quality estimate less than 0.9.
\end{enumerate}
If at least one criterion holds, we reject the image, as illustrated in the bottom row of \cref{fig:domain-shift}.

Applying our method, we obtain 70 pseudo-labeled images. The incorporation of data seen during training of the initial DNN, \ie the Cityscapes training data, restrains the network from adapting onto the new domain. We therefore decided to extend the model only on $\mathcal{D}^{C+1}$.

Class-wise evaluation results are reported in \cref{tab:a2d2_guardrail-domain-shift}. Even with a domain shift, we achieve an IoU of 20.90 $\pm$ 1.73\% for the novel class. This is less than the value obtained with prior fine-tuning. However, this DNN still outperforms the PSPNet from the previous experiment considering only the precision. The low recall values are tolerable since many guardrails are still assigned to the ``supercategory'' \emph{fence}.
For most other classes, the IoU values increase or remain roughly the same. In contrast to the other experiments, the \emph{motorcycle} class improves in IoU, precision and recall values. Only classes that are rare in rural street scenes, \eg \emph{sidewalk} or \emph{bicycle}, suffer from the incremental training.

A visual comparison of the experiments 4(a), 4(b) and 5 is provided in \cref{fig:results-4and5}. All three extended DNNs have learned to predict the novel class to some extent. The prior fine-tuned networks show similar predictions, though DeepLabV3+ is much more precise than the PSPNet and better recognizes the guardrail on the right. The model from the fifth experiment predicts the left guardrail as \emph{fence} (which is not totally mistaken), though it performs better on the right-hand guardrail than the others. Both oracles illustrate, that the \emph{guardrail} class is learnable with high accuracy, still leaving room for improvement of unsupervised methods.

\begin{figure}[t]
\captionsetup[subfigure]{labelformat=empty, position=top}
    \centering
    \subfloat[]{\includegraphics[width=0.23\textwidth]{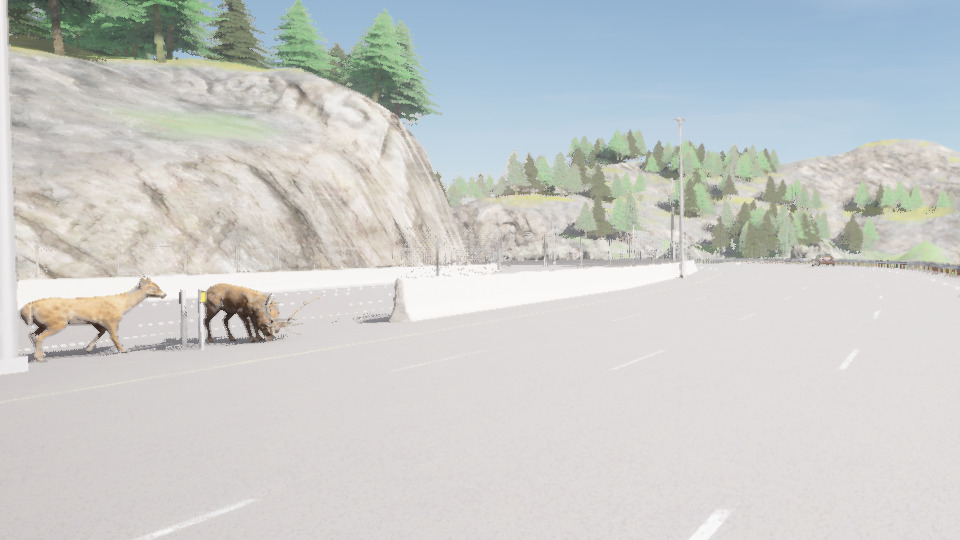}}~
    \subfloat[]{\includegraphics[width=0.23\textwidth]{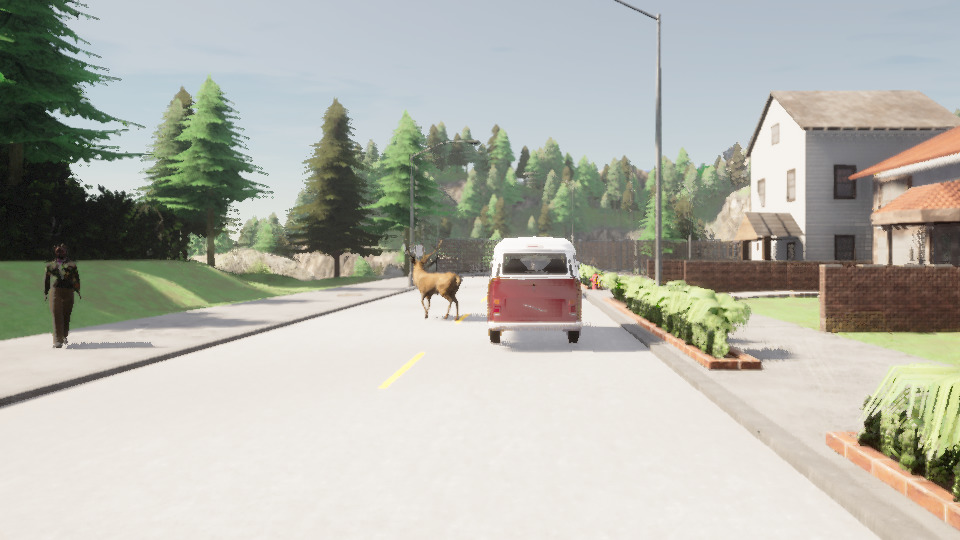}}
    \caption{Two examples from our CARLA test dataset including the novel class \emph{deer}.}
    \label{fig:carla}
\end{figure}

\begin{figure*}[t]
    \centering
    \includegraphics[width=0.95\textwidth]{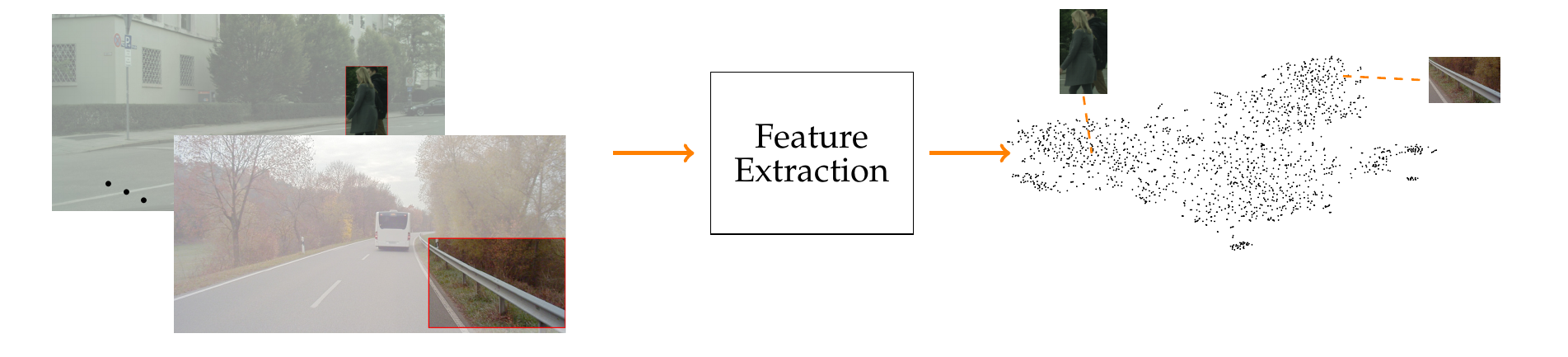}
    \caption{Coarse illustration of the feature extraction process. Detected unknown objects (here: human and guardrail) are cropped out (indicated by the red box). The image patches are fed into an encoder, the resulting feature vectors are then projected into a two dimensional space.}
    \label{fig:feature-extraction}
\end{figure*}

\section{Synthetic Dataset}\label{sec:carla-dataset}

We generated a synthetic dataset with the CARLA simulator, that contains novel classes such as \emph{deer} in the test data. Two examples are provided in \cref{fig:carla}. All classes considered as novel are never seen before, \ie they are not contained in the training data. Besides that, the street scenes for training and testing are recorded under identical conditions, \ie on the same maps, with the same weather conditions, camera angles etc., so that the segmentation network is not distracted by anything different than the novel objects.

\begin{table}[t]
    \centering
    \begin{tabular}{|l|c|c|}
        \hline
        \textbf{experiment} & \textbf{$\#$metrics} &  \textbf{$\#$segments in training set}\\\hline
        \textbf{1} & 71 & 608,906\\
        \textbf{2} & 73 & 571,853\\
        \textbf{3} & 67 & 946,318\\
        \textbf{4a} & 75 & 492,210\\
        \textbf{4b} & 75 & 313,720\\
        \textbf{5} & 75 & 535,457\\\hline
    \end{tabular}
    \caption{Overview about the training data of the meta regressor for each experiment. We report the number of metrics per segment $k$ (that depends on the number of classes $|\mathcal{C}|$) as well as the number of segments produced by the initial network during inference of the training data.}
    \label{tab:regression-data}
\end{table}

\begin{table*}[t]
    \centering
    \resizebox{0.95\textwidth}{!}{
    \begin{tabular}{|l|ccc|ccc|ccc|}
            \hline
            model & \multicolumn{3}{c|}{DenseNet201} & \multicolumn{3}{c|}{ResNet18} & \multicolumn{3}{c|}{ResNet152} \\\hline\hline
            metric & IoU & precision & recall & IoU & precision & recall & IoU & precision & recall\\\hline
        human & 39.80 $\pm$ 0.73 & \textbf{60.60} $\pm$ 1.20 & 53.72 $\pm$ 1.42 & \textbf{40.56} $\pm$ 0.95 & 54.80 $\pm$ 4.50 & 61.50 $\pm$ 4.12 & 40.30 $\pm$ 0.94 & 52.17 $\pm$ 1.59 & \textbf{63.97} $\pm$ 1.71 \\
        mean over $C$ & \textbf{68.53} $\pm$ 0.27 & 83.32 $\pm$ 0.28 & \textbf{77.17} $\pm$ 0.60 & 68.19 $\pm$ 0.56 & 84.44 $\pm$ 0.28 & 75.84 $\pm$ 0.90 & 67.44 $\pm$ 0.36 & \textbf{84.73} $\pm$ 0.36 & 74.58 $\pm$ 0.48\\
        mean over $C^+$ & \textbf{66.94} $\pm$ 0.27 & 82.05 $\pm$ 0.25 & \textbf{75.86} $\pm$ 0.55 & 66.65 $\pm$ 0.58 & 82.80 $\pm$ 0.22 & 75.05 $\pm$ 0.68 & 65.94 $\pm$ 0.31 & \textbf{82.92} $\pm$ 0.25 & 73.99 $\pm$ 0.38 \\\hline
    \end{tabular}}
    \caption{Ablation study for the feature extractor: we provide the IoU, precision and recall values for the first experiment, where we incrementally extend a DeepLabV3+ by the novel class \emph{human} on the Cityscapes dataset, using three different architectures for the feature extraction. For each feature extractor, we report the mean and standard deviation over five runs, respectively.}
    \label{tab:encoder}
\end{table*}

\begin{figure*}[h!]
    \captionsetup[subfigure]{labelformat=empty}
    \centering
    \subfloat[]{\rotatebox[origin=lb]{90}{\scriptsize ~~~~ 1.\ experiment}}~~
    \subfloat[]{\includegraphics[width=0.23\textwidth]{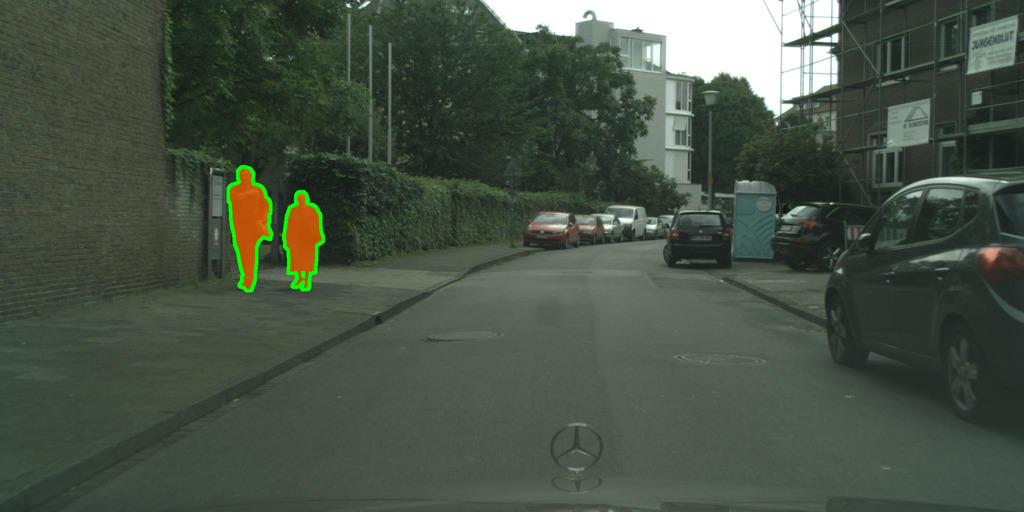}}
    \subfloat[]{\includegraphics[width=0.23\textwidth]{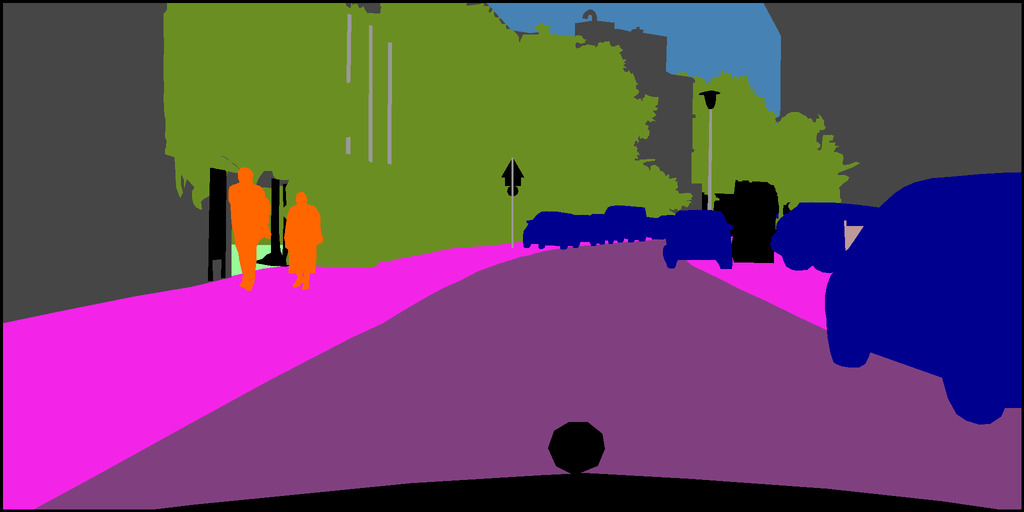}}
    \subfloat[]{\includegraphics[width=0.23\textwidth]{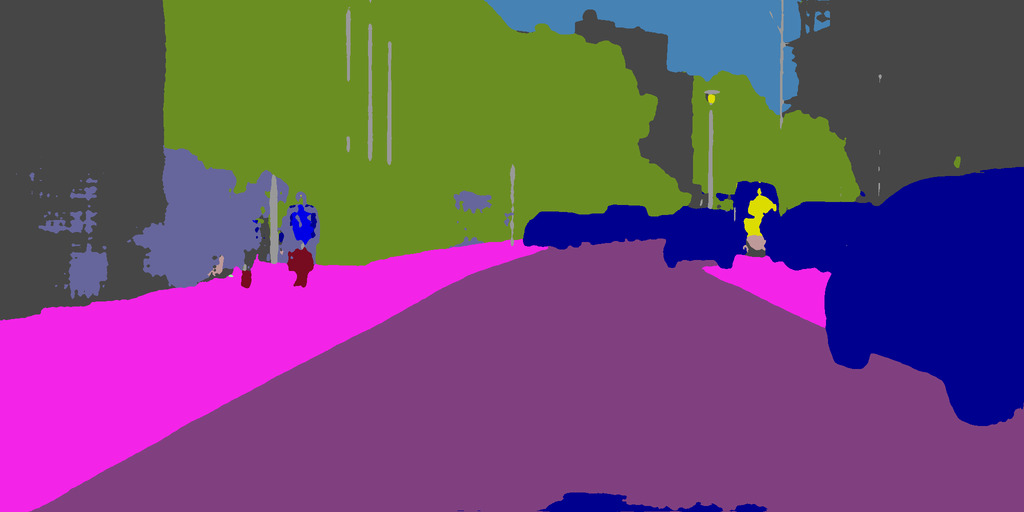}}
    \subfloat[]{\includegraphics[width=0.23\textwidth]{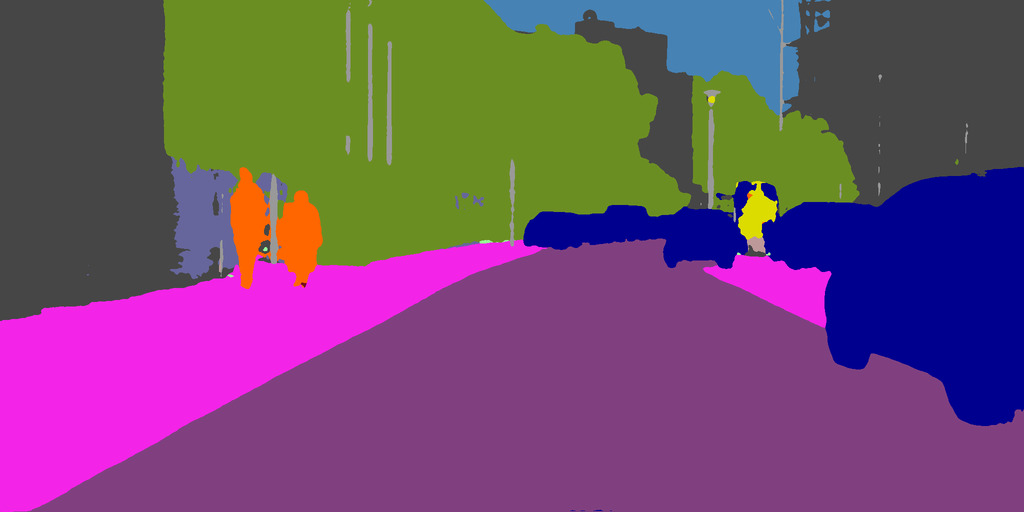}}\\
    \vspace{-0.9cm}
    \subfloat[]{\rotatebox[origin=lb]{90}{\scriptsize ~~~~ 2.\ experiment}}~~
    \subfloat[]{\includegraphics[width=0.23\textwidth]{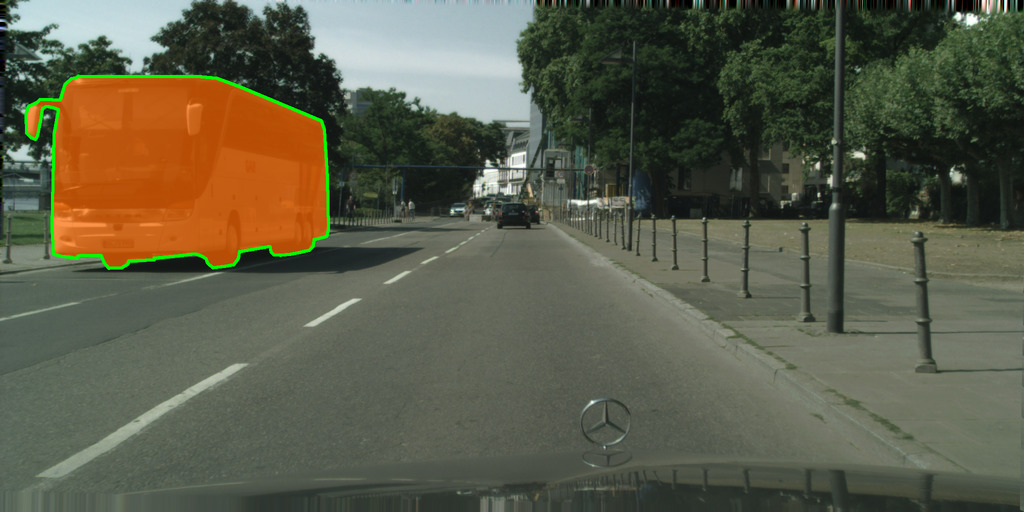}}
    \subfloat[]{\includegraphics[width=0.23\textwidth]{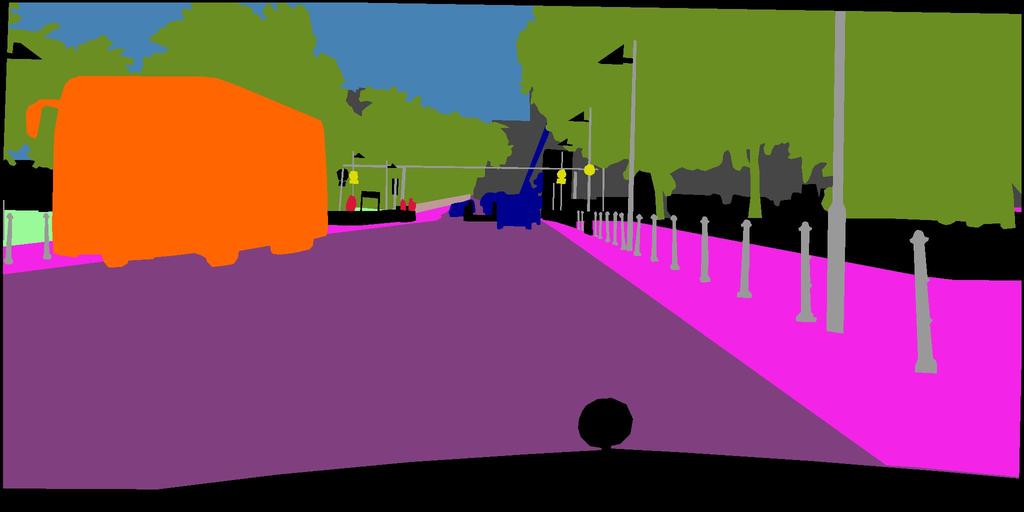}}
    \subfloat[]{\includegraphics[width=0.23\textwidth]{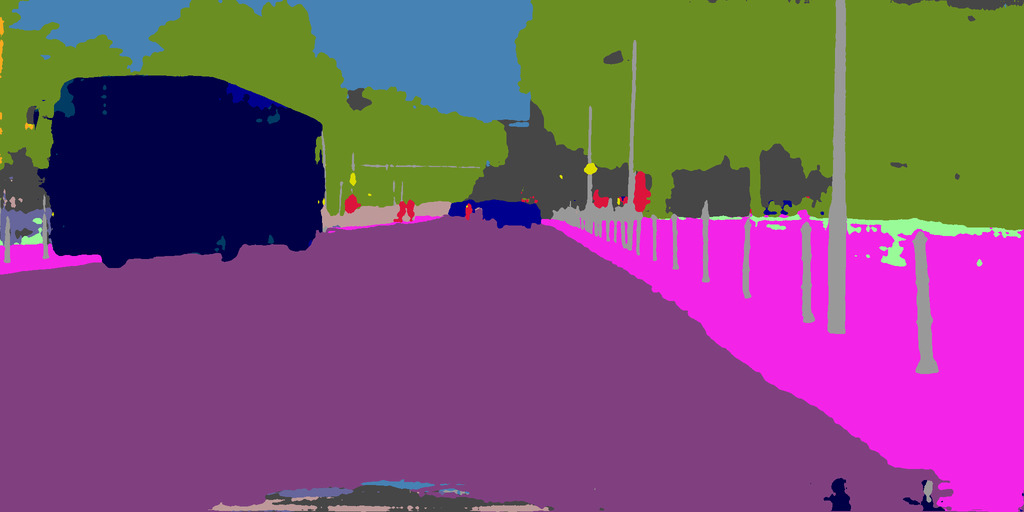}}
    \subfloat[]{\includegraphics[width=0.23\textwidth]{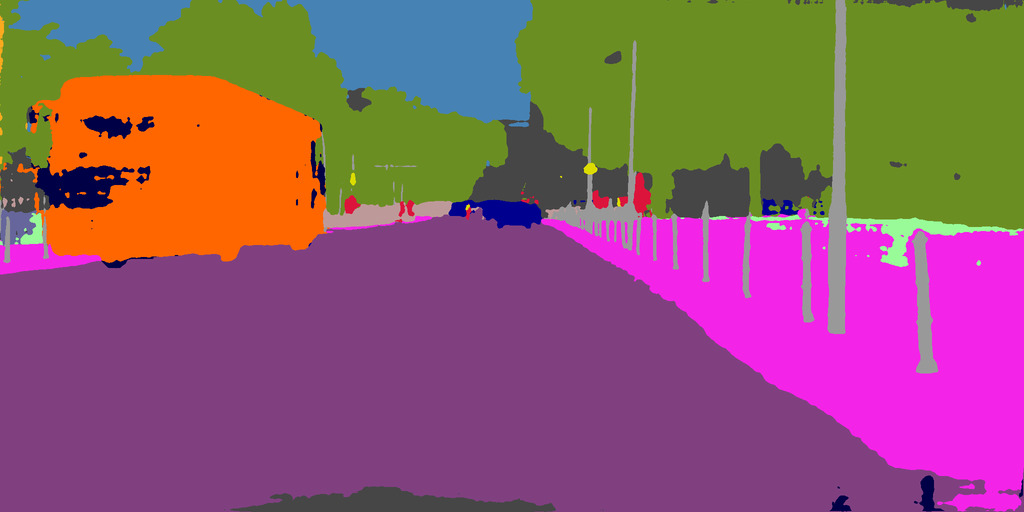}}\\
    \vspace{-0.9cm}
    \subfloat[]{\rotatebox[origin=lb]{90}{\scriptsize ~~~~ 3.\ experiment}}~~
    \subfloat[]{\includegraphics[width=0.23\textwidth]{figures/munster_000035_000019_exp3.jpg}}
    \subfloat[]{\includegraphics[width=0.23\textwidth]{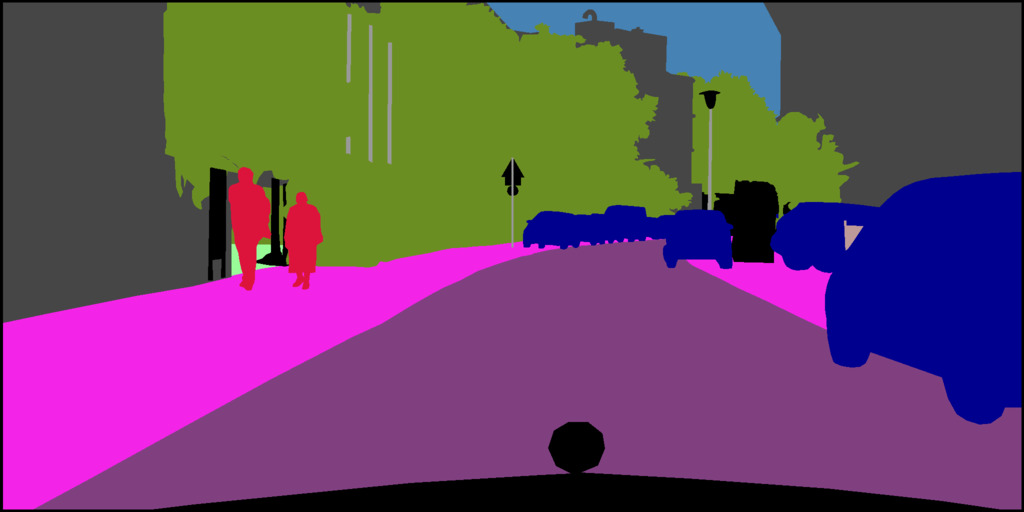}}
    \subfloat[]{\includegraphics[width=0.23\textwidth]{figures/munster_000035_000019_init.jpg}}
    \subfloat[]{\includegraphics[width=0.23\textwidth]{figures/munster_000035_000019_exp3_pred.jpg}}\\
    \vspace{-0.9cm}
    \subfloat[]{\rotatebox[origin=lb]{90}{\scriptsize ~ 4.\ experiment (a)}}~~
    \subfloat[]{\includegraphics[trim={0 0 0 124px},clip,width=0.23\textwidth]{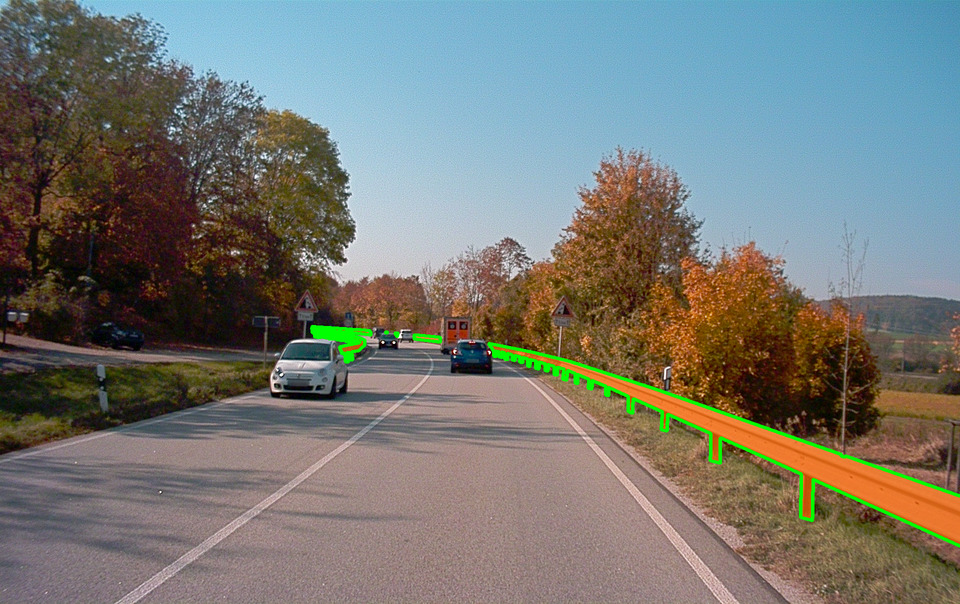}}
    \subfloat[]{\includegraphics[trim={0 0 0 124px},clip,width=0.23\textwidth]{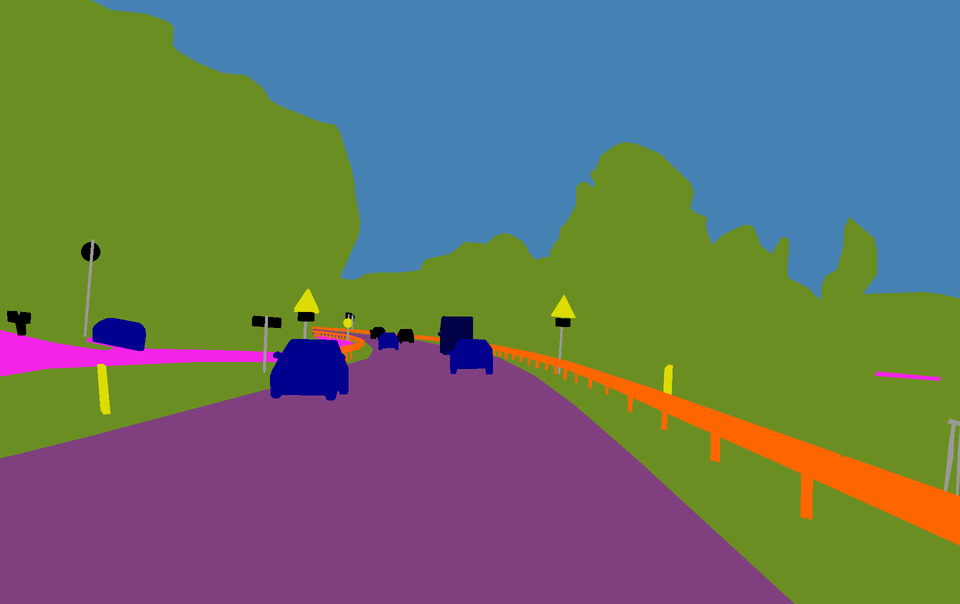}}
    \subfloat[]{\includegraphics[trim={0 0 0 124px},clip,width=0.23\textwidth]{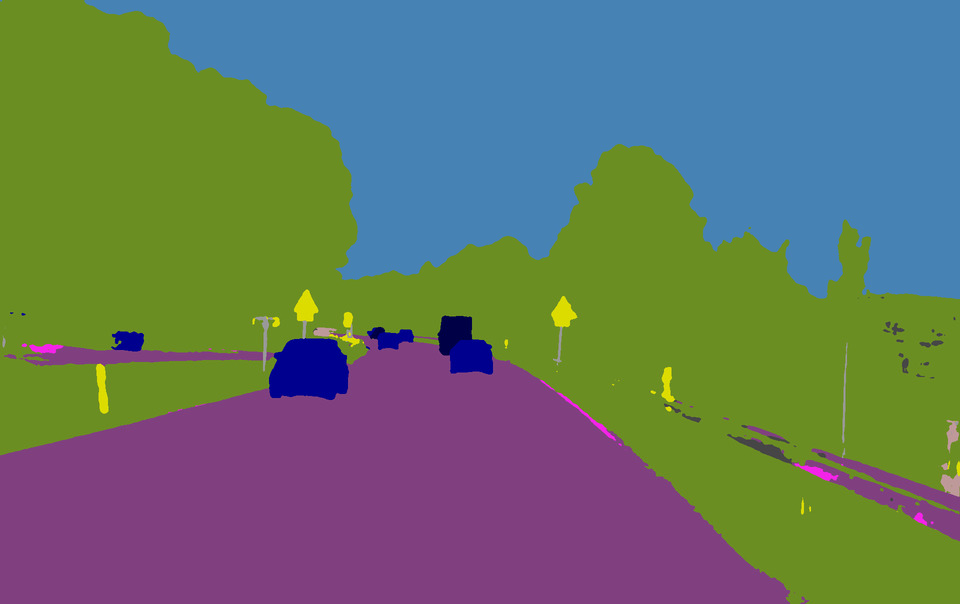}}
    \subfloat[]{\includegraphics[trim={0 0 0 124px},clip,width=0.23\textwidth]{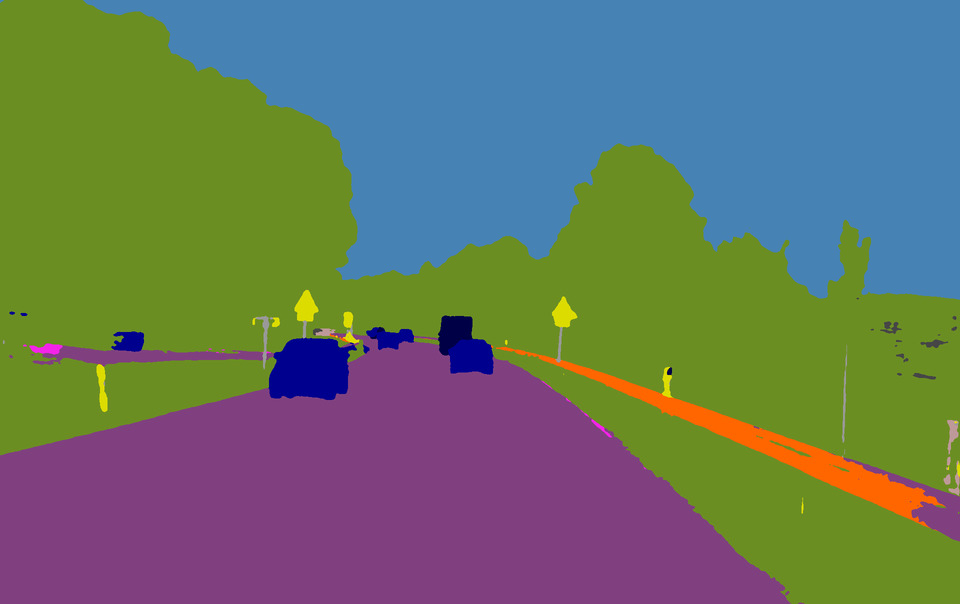}}\\
    \vspace{-0.9cm}
    \subfloat[]{\rotatebox[origin=lb]{90}{\scriptsize ~ 4.\ experiment (b)}}~~
    \subfloat[]{\includegraphics[trim={0 0 0 124px},clip,width=0.23\textwidth]{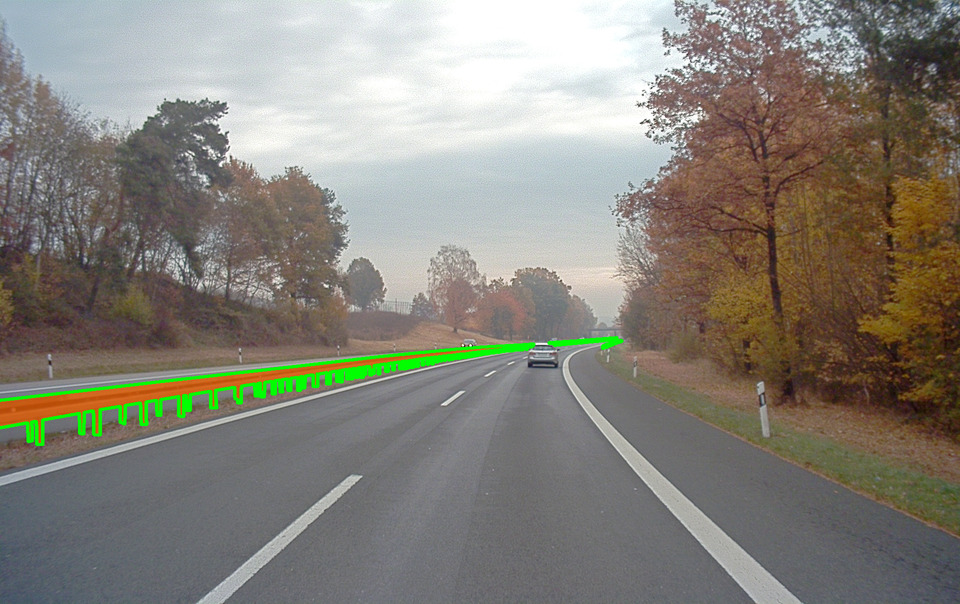}}
    \subfloat[]{\includegraphics[trim={0 0 0 124px},clip,width=0.23\textwidth]{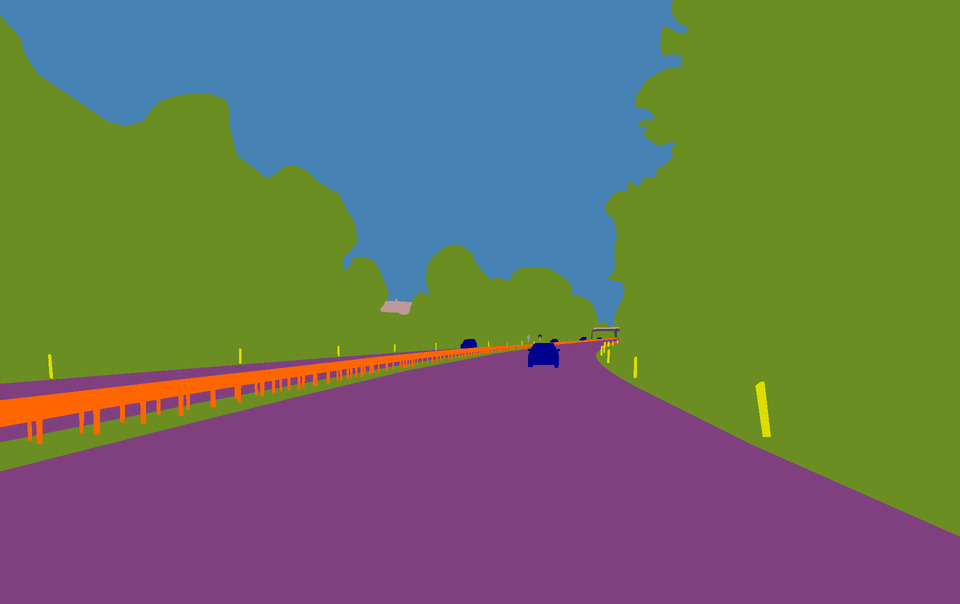}}
    \subfloat[]{\includegraphics[trim={0 0 0 124px},clip,width=0.23\textwidth]{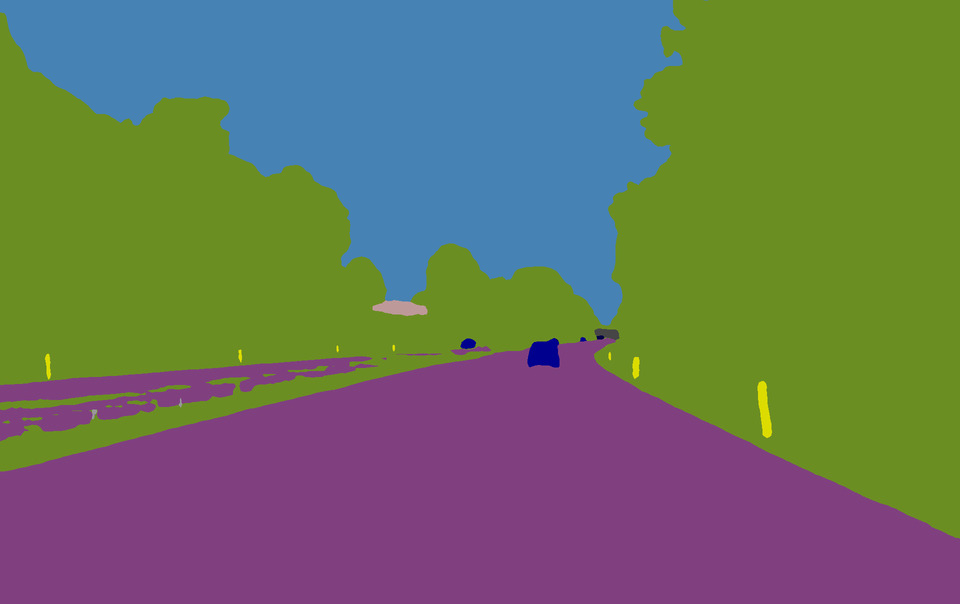}}
    \subfloat[]{\includegraphics[trim={0 0 0 124px},clip,width=0.23\textwidth]{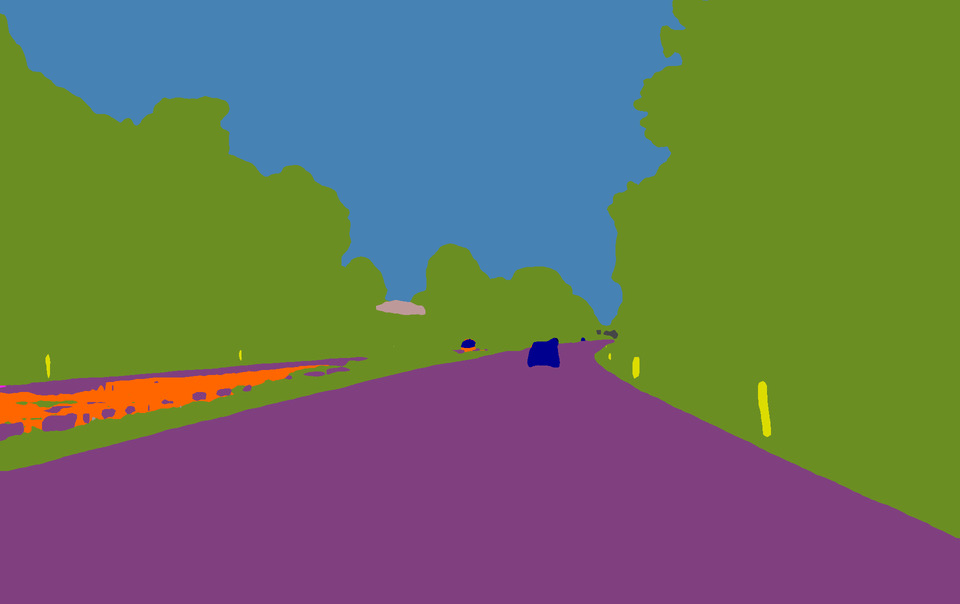}}\\
    \vspace{-0.9cm}
    \subfloat[]{\rotatebox[origin=lb]{90}{\scriptsize ~~~~ 5.\ experiment}}~~
    \subfloat[image \& novelty annotation]{\includegraphics[trim={0 0 0 124px},clip,width=0.23\textwidth]{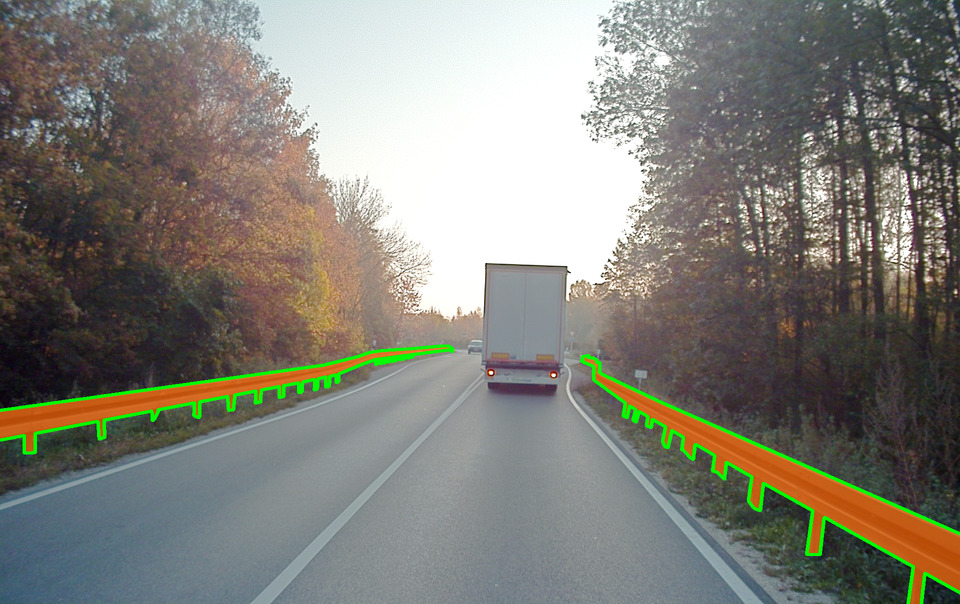}}
    \subfloat[ground truth]{\includegraphics[trim={0 0 0 124px},clip,width=0.23\textwidth]{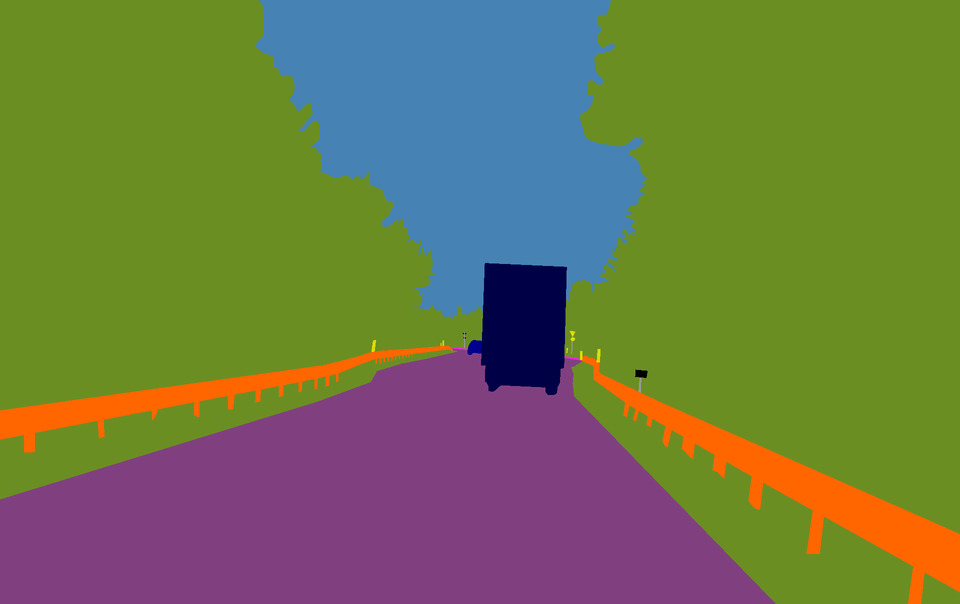}}
    \subfloat[prediction of initial DNN]{\includegraphics[trim={0 0 0 124px},clip,width=0.23\textwidth]{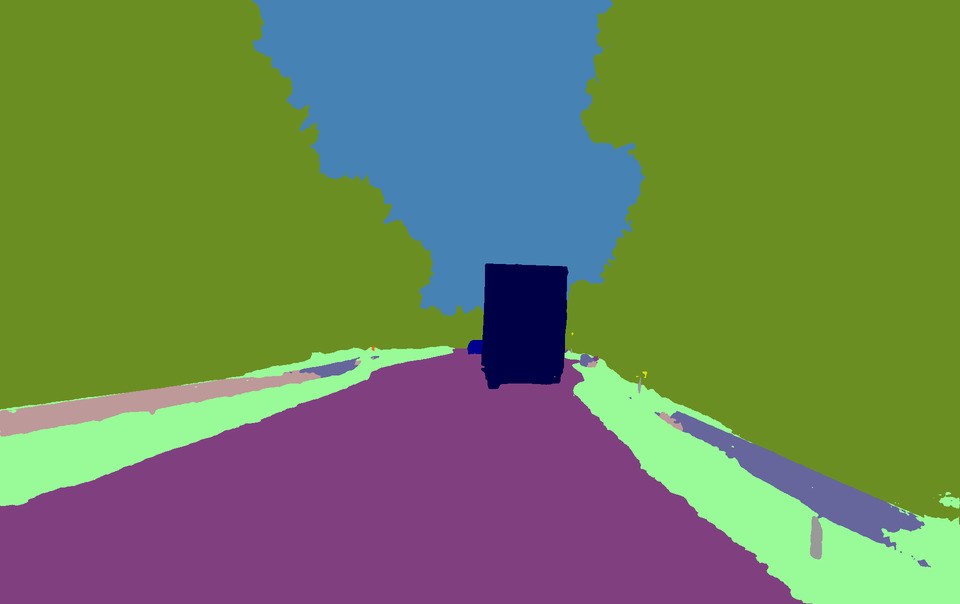}}
    \subfloat[prediction of extended DNN]{\includegraphics[trim={0 0 0 124px},clip,width=0.23\textwidth]{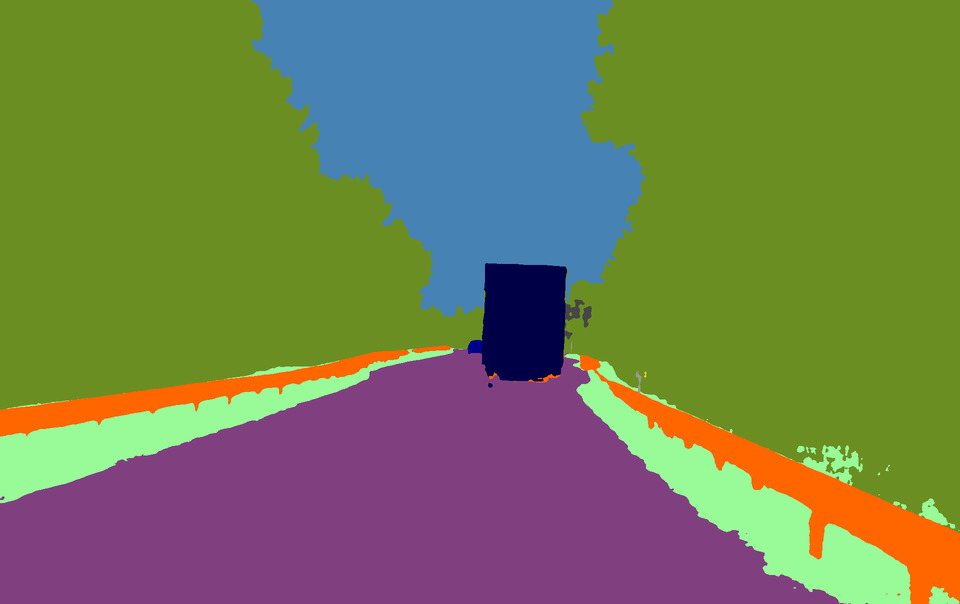}}
    \caption{Example images from the validation data for all conducted experiments, respectively.}
    \label{fig:results}
\end{figure*}

\section{Modules}

We present a modular procedure, this is, the individual modules can be modified or exchanged. In this section, we provide a deeper insight into the modules \textbf{meta regressor} and \textbf{feature extractor}. 

\subsection{Uncertainty metrics \& Meta regression}

For every segment $k\in\mathcal{K}(\mathcal{D}^\mathrm{train})$ we compute the following metrics:

\begin{itemize}

\item the size of the segment $k$, its interior $k^\mathrm{o}$ and its boundary $\partial k$:
\begin{equation*}
    S(k) = |k|,~ S^\mathrm{o}(k)=|k^\mathrm{o}|,~ \partial S(k) = |\partial k|
\end{equation*}
\item the relative sizes:
\begin{equation*}
    \tilde{S}(k) = S(k)/\partial S(k),~ \tilde{S}^\mathrm{o}(k)=S^\mathrm{o}(k)/\partial S(k)
\end{equation*}
\item several dispersion measures aggregated over $k$, $k^\mathrm{o}$ and $\partial k$, respectively:
\begin{equation*}
\begin{split}
    \Bar{D}(k) = \frac{1}{S} \sum_{z\in k}D_z(x),~ \Bar{D}^\mathrm{o}(k) = \frac{1}{S^\mathrm{o}} \sum_{z\in k^\mathrm{o}}D_z(x),\\
    \partial \Bar{D}(k) = \frac{1}{\partial S} \sum_{z\in \partial k}D_z(x)
\end{split}
\end{equation*}
where $D \in\{E,M,V\}$, \ie softmax entropy $E$, probability margin $M$ and variation ration $V$.
\item the relative dispersion measures:
\begin{equation*}
    \tilde{\Bar{D}}(k) = \Bar{D}(k)S(k),~ \tilde{\Bar{D}}^\mathrm{o}(k)=\Bar{D}^\mathrm{o}(k)\tilde{S}^\mathrm{o}(k)
\end{equation*}
$D \in\{E,M,V\}$.
\item the variance of the dispersion measures
\item the predicted class $c\in\mathcal{C}$
\item the mean softmax probabilities for each class $c\in\mathcal{C}$
\item the pixel position of the segment's geometric center
\item the ratio of the amount of pixels in the neighborhood of segment $k$ predicted to belong to class $c\in\mathcal{C}$ to the neighborhood size for each class $c\in\mathcal{C}$
\end{itemize}

Further, we compute the IoU (averaged over each segment), which is the only metric that requires ground truth and serves as target value for the meta regressor. The number of training metrics, \ie explanatory variables, is reported in \cref{tab:regression-data} for each experiment. This is, the training data for the meta regressor has a dimension of $|\mathcal{K}(\mathcal{D}^\mathrm{train})| \times \#$metrics.

\subsection{Feature extractor} 

We apply an image classification CNN, pre-trained on ImageNet, without the final classification layer to extract features of image patches as illustrated in \cref{fig:feature-extraction}. This feature extraction CNN can be exchanged arbitrarily, as long as the resulting feature vectors equally sized for different input dimensions. In \cref{tab:encoder} we compare the results for experiment 1, using three different feature extractors, namely DenseNet201, ResNet18 and ResNet152.

\begin{figure*}[t]
    \captionsetup[subfigure]{labelformat=empty}
    \centering
    \subfloat[experiment 1]{\includegraphics[width=0.4\textwidth]{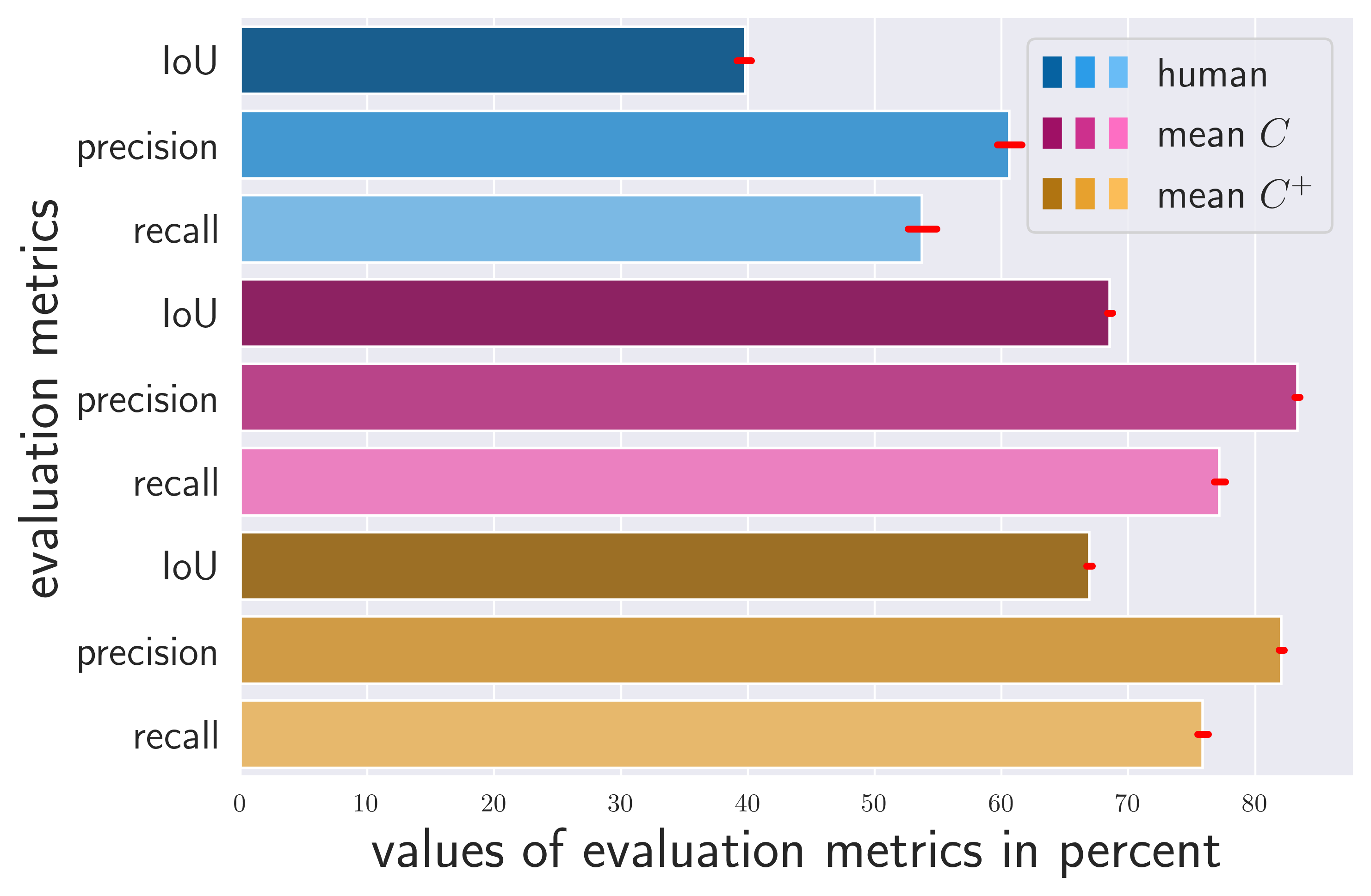}}~
    \subfloat[experiment 2]{\includegraphics[width=0.4\textwidth]{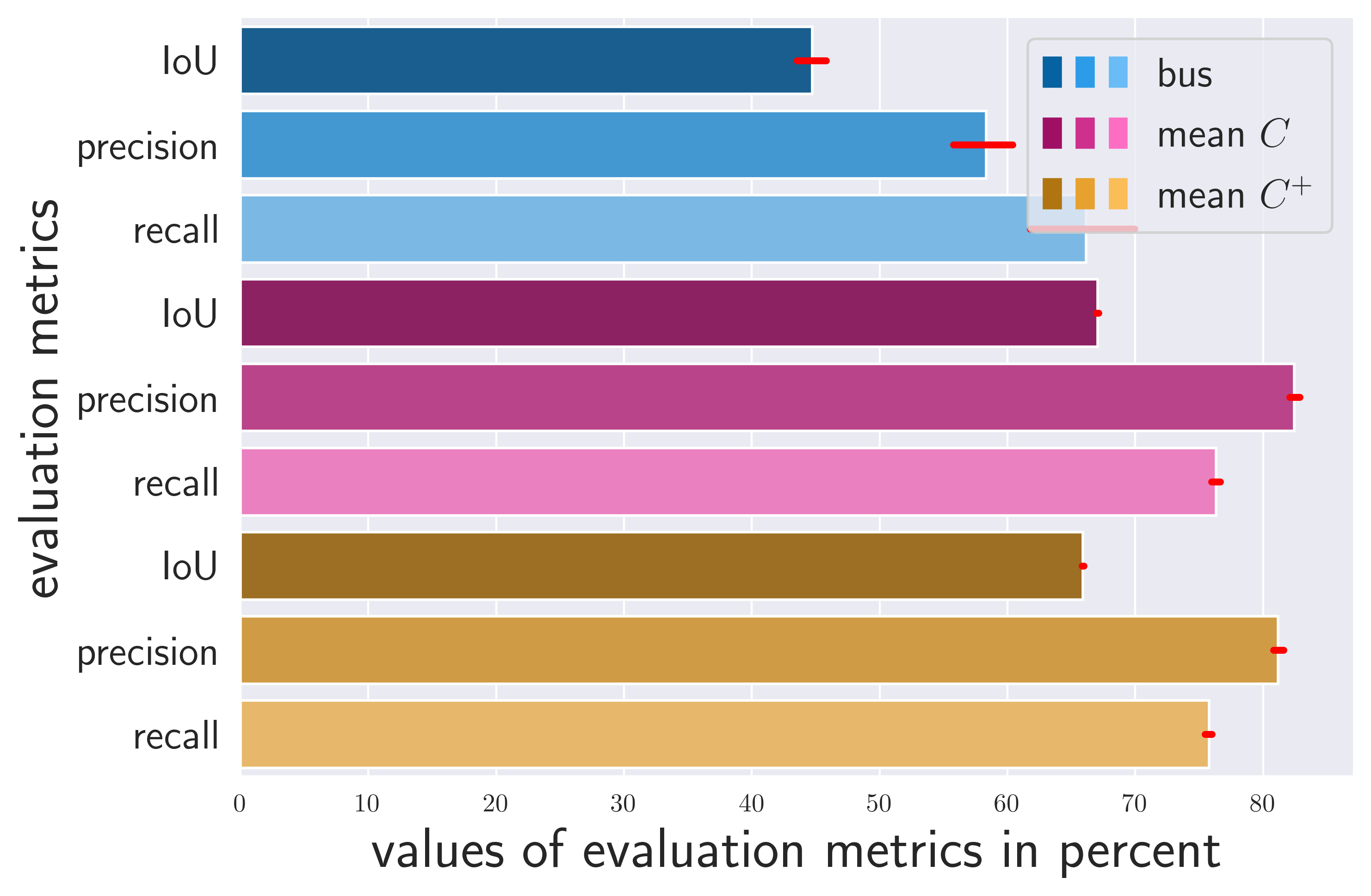}}\\
    \subfloat[experiment 3]{\includegraphics[width=0.4\textwidth]{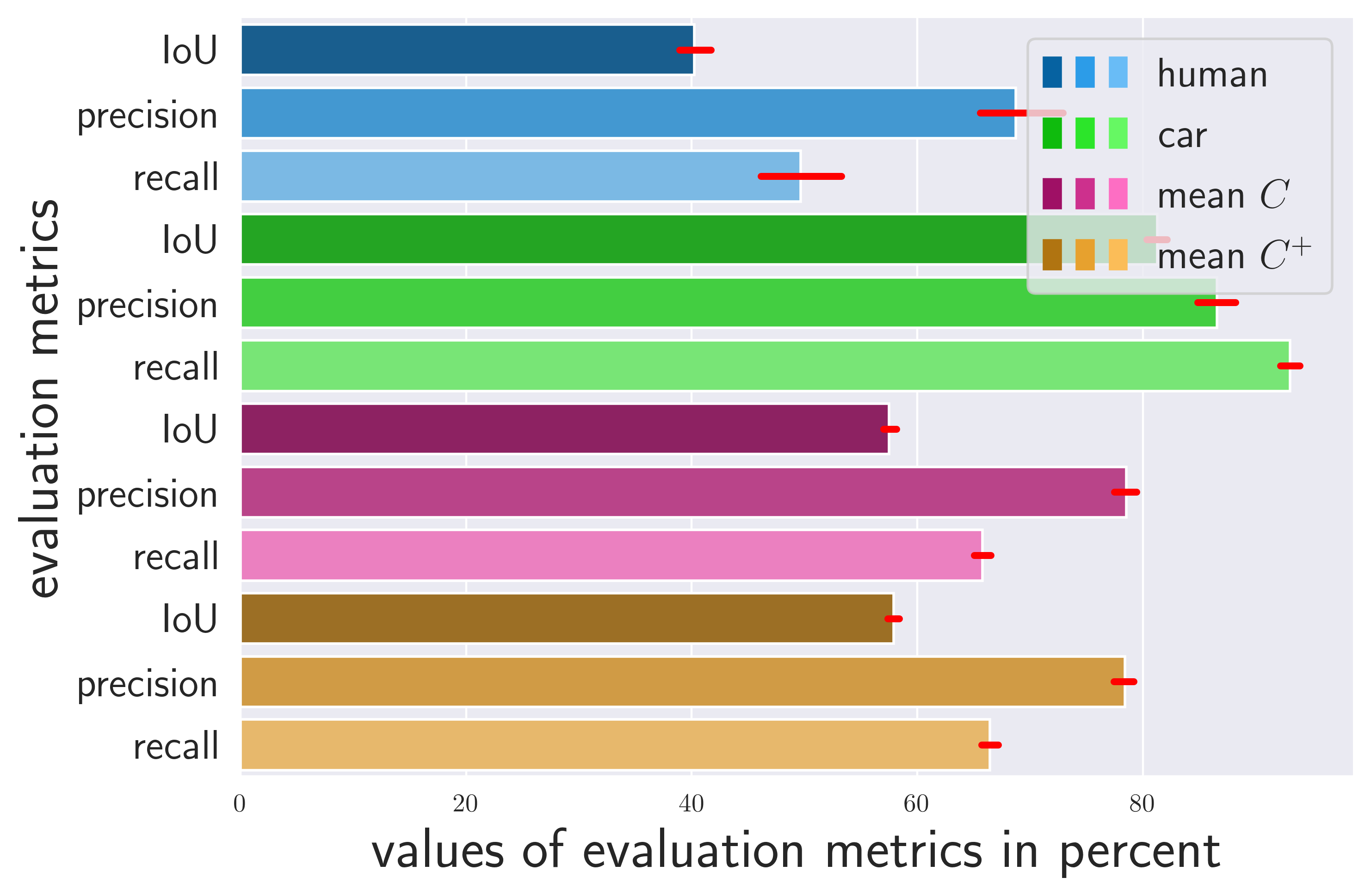}}~
    \subfloat[experiment 4a]{\includegraphics[width=0.4\textwidth]{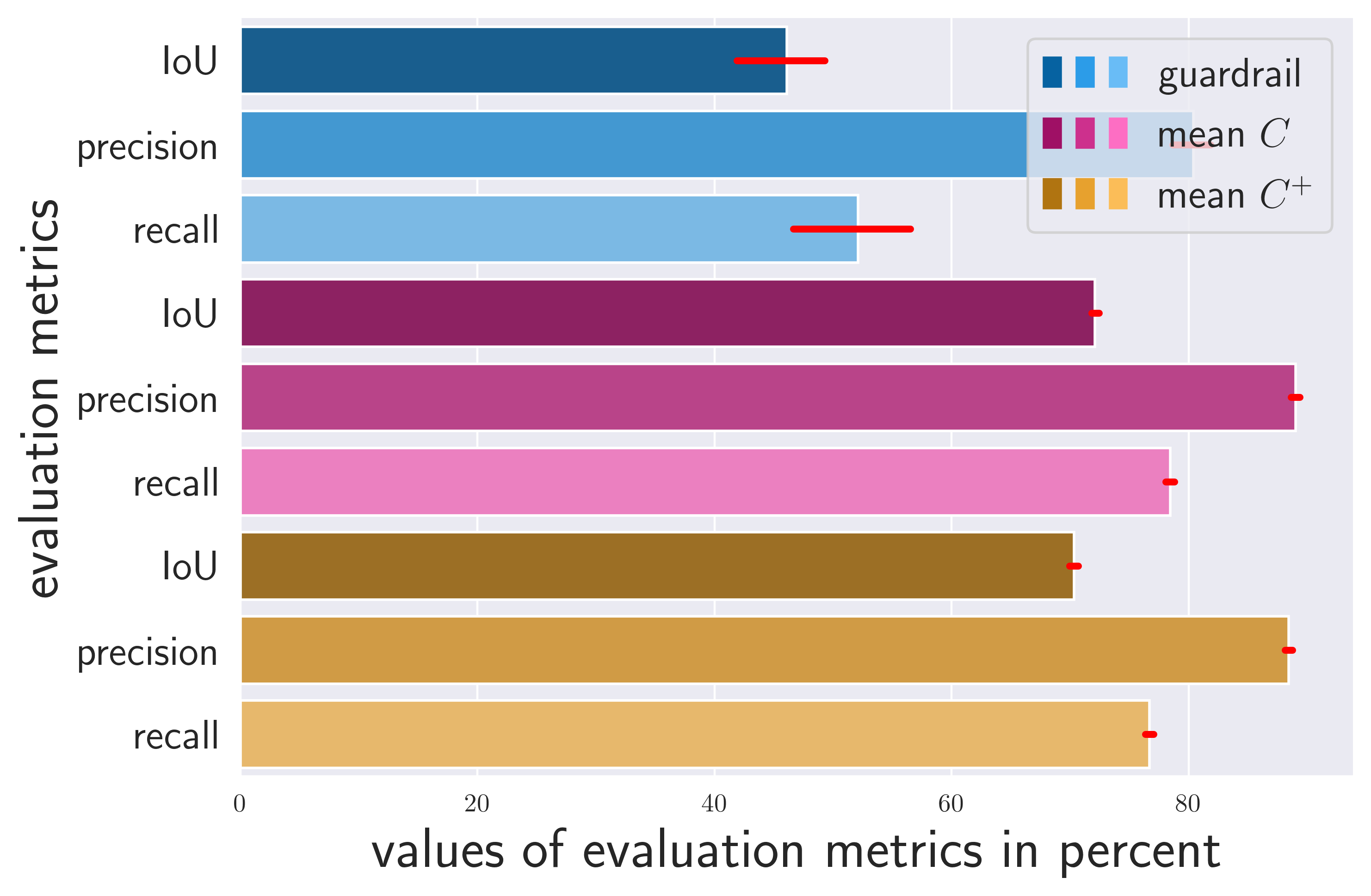}}\\
    \subfloat[experiment 4b]{\includegraphics[width=0.4\textwidth]{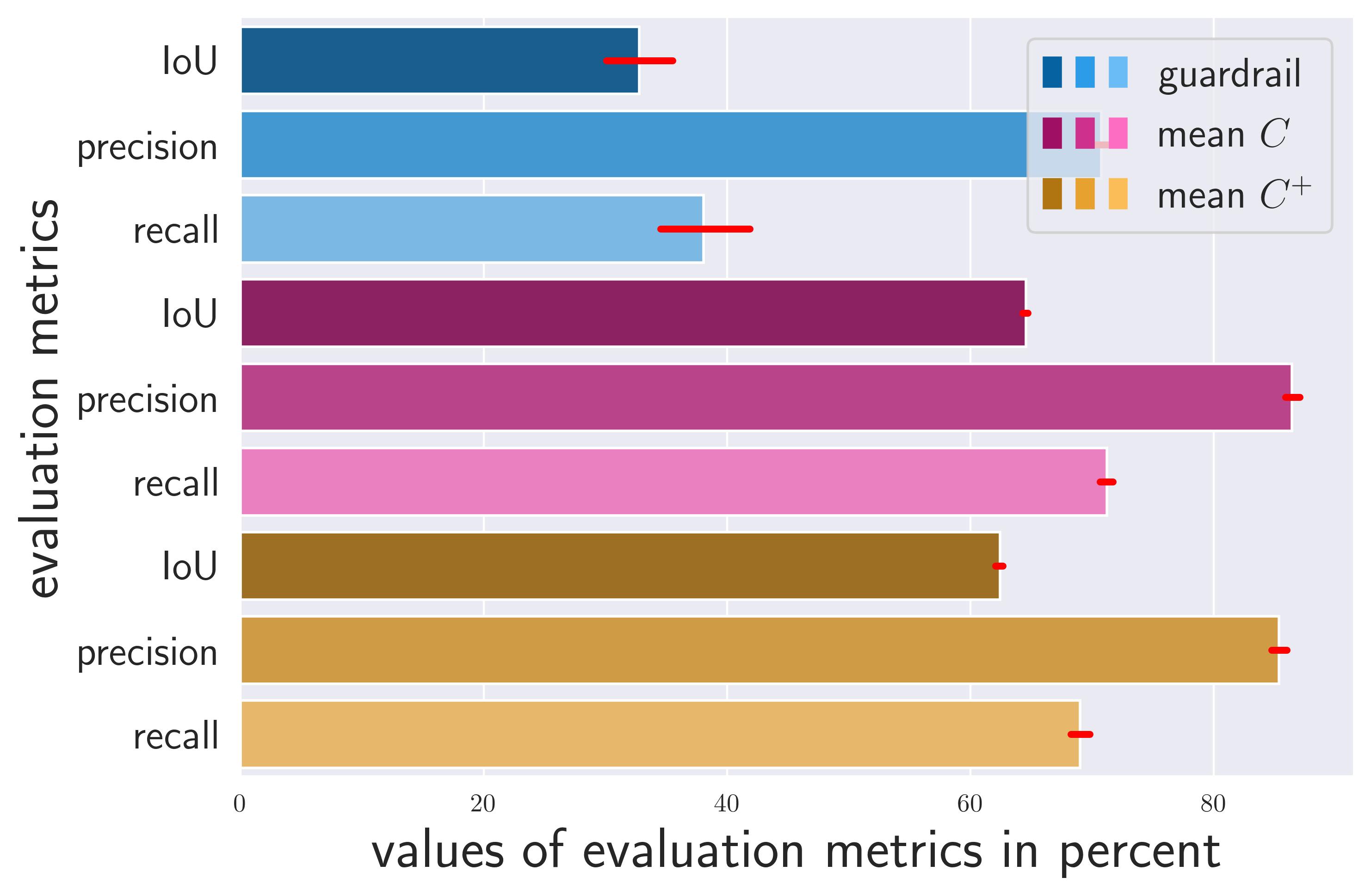}}~
    \subfloat[experiment 5]{\includegraphics[width=0.4\textwidth]{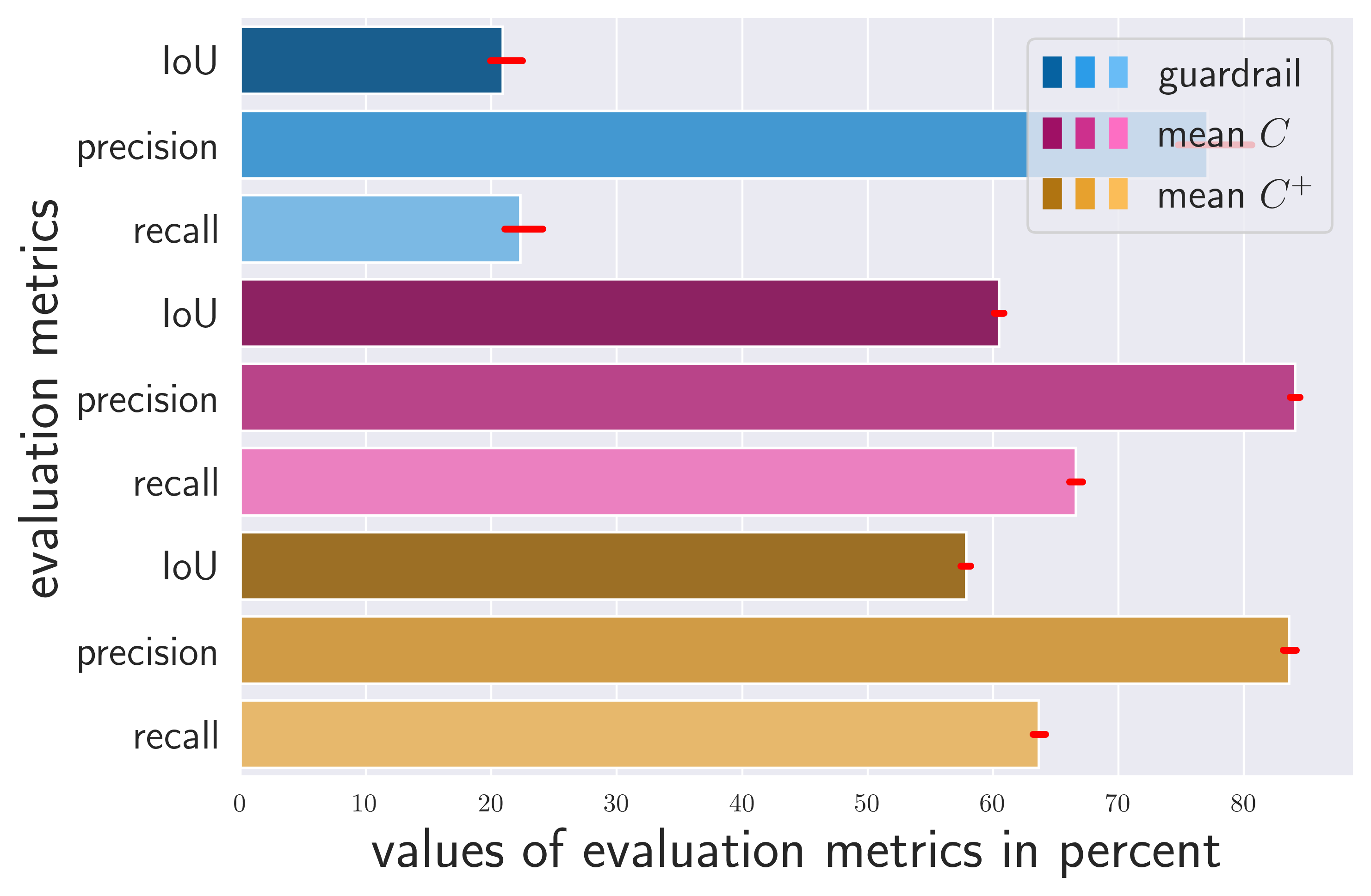}}
    \caption{Bar plots showing the evaluation metrics averaged over five runs per experiment. The standard deviation is indicated by the red lines.}
    \label{fig:mean_std}
\end{figure*}

\section{Results - Visualization}

In \cref{fig:results} we provide an overall visualization of all conducted experiments. Our approach predicts the novel objects with adequate accuracy while the predictions of the initial and the extended DNNs remain similar on previously-known objects. Note that in the fifth experiment, the A2D2 ground truth consists of coarser classes than the segmentation DNN, which is trained on Cityscapes. Further, \cref{fig:mean_std} illustrates the mean and standard deviation of the main evaluation metrics for each experiment, respectively. We observe, that the standard deviation values regarding the mean over $\mathcal{C}$ are at the maximum $1.20\%$, and besides that $\leq 1\%$. This is, our method is robust considering the initially known classes. In experiment 4 (a) and (b), we observe the highest standard deviation for the IoU values of the novel class with $4.80\%$ and $3.48\%$, respectively, which is $<2\%$ for all other experiments.

\end{document}